\documentclass{article}

     \PassOptionsToPackage{numbers}{natbib}

\usepackage[final]{neurips_2021}

\usepackage[utf8]{inputenc} %
\usepackage[T1]{fontenc}    %
\usepackage[]{hyperref}       %
\usepackage{url}            %
\usepackage{booktabs}       %
\usepackage{amsfonts}       %
\usepackage{amssymb,amsmath,amsthm}
\usepackage{nicefrac}       %
\usepackage{microtype}      %
\usepackage{algorithm}
\usepackage{algpseudocode}  %
\usepackage{graphicx}

\usepackage{lipsum}         %
\usepackage{xspace}         %
\usepackage{xargs}          %
\usepackage{mathtools}
\usepackage{bm}
\usepackage[dvipsnames]{xcolor} %
\usepackage{tabularx}       %
\usepackage{wrapfig}        %

\usepackage[export]{adjustbox}

\usepackage{tikz}           %
\usepackage{ifthen}
\usepackage{enumitem}
\usepackage{cleveref}
\usetikzlibrary{patterns}
\usetikzlibrary{positioning}

\makeatletter
\newcommand{\currentfontsize}{\f@size pt}
\makeatother

\usepackage[toc, page, header]{appendix}
\usepackage{url}

\definecolor{tab10_blue}{rgb}{0.121, 0.466, 0.705}
\definecolor{tab10_orange}{rgb}{1.0,   0.498, 0.054}
\definecolor{tab10_green}{rgb}{0.172, 0.627, 0.172}
\definecolor{tab10_red}{rgb}{0.839, 0.152, 0.156}
\definecolor{tab10_purple}{rgb}{0.580, 0.403, 0.741}
\definecolor{tab10_brown}{rgb}{0.549, 0.337, 0.294}
\definecolor{tab10_pink}{rgb}{0.890, 0.466, 0.760}
\definecolor{tab10_gray}{rgb}{0.498, 0.498, 0.498}
\definecolor{tab10_olive}{rgb}{0.737, 0.741, 0.133}
\definecolor{tab10_cyan}{rgb}{0.090, 0.745, 0.811}

\newcommandx{\customComment}[3]{\textcolor{#2}{\textsl{#1: #3}}}
\newcommandx{\customTodo}[3]{\textcolor{#2}{\textsl{#1: #3}}}

\newcommandx{\Thijs}[1]{\customComment{Thijs}{tab10_blue}{#1}}
\newcommandx{\Tao}[1]{\customComment{Tao}{tab10_blue}{#1}}
\newcommandx{\Lie}[1]{\customComment{Lie}{tab10_orange}{#1}}
\newcommandx{\Anastasia}[1]{\customComment{Anastasia}{tab10_green}{#1}}
\newcommandx{\Martin}[1]{\customComment{Martin}{tab10_red}{#1}}
\newcommandx{\NeedsRef}[1]{\textcolor{tab10_pink}{(ref)}}
\newcommandx{\Seb}[1]{\customComment{Seb}{tab10_pink}{#1}}

\newcommandx{\ThijsTodo}[1]{\customTodo{Thijs}{tab10_blue}{#1}}
\newcommandx{\LieTodo}[1]{\customTodo{Lie}{tab10_orange}{#1}}
\newcommandx{\AnastasiaTodo}[1]{\customTodo{Anastasia}{tab10_green}{#1}}
\newcommandx{\MartinTodo}[1]{\customTodo{Martin}{tab10_red}{#1}}

\newtheorem*{rep@theorem}{\rep@title}
\newcommand{\newreptheorem}[2]{%
\newenvironment{rep#1}[1]{%
 \def\rep@title{#2 \ref{##1}}%
 \begin{rep@theorem}}%
 {\end{rep@theorem}}}
\newreptheorem{lemma}{Lemma'}
\newreptheorem{definition}{Definition'}
\newreptheorem{proposition}{Proposition'}
\newreptheorem{theorem}{Theorem'}

\DeclarePairedDelimiterX{\lin}[2]{\langle}{\rangle}{#1, #2}

\DeclarePairedDelimiterX{\abs}[1]{\lvert}{\rvert}{#1}
\DeclarePairedDelimiterX{\norm}[1]{\lVert}{\rVert}{#1}
\DeclarePairedDelimiterX{\cbr}[1]{\{}{\}}{#1} %
\DeclarePairedDelimiterX{\rbr}[1]{(}{)}{#1} %
\DeclarePairedDelimiterX{\sbr}[1]{[}{]}{#1} %

\renewcommand{\epsilon}{\varepsilon}

\providecommand{\R}{\mathbb{R}} %
\providecommand{\C}{\mathbb{C}} %

\DeclareMathOperator{\expect}{\mathbb{E}}
\DeclareMathOperator{\E}{\mathbb{E}}

\DeclareMathOperator{\sgn}{sign}
\makeatletter
\def\sign{\@ifnextchar*{\@sgnargscaled}{\@ifnextchar[{\sgnargscaleas}{\@ifnextchar{\bgroup}{\@sgnarg}{\sgn} }}}
\def\@sgnarg#1{\sgn\rbr{#1}}
\def\@sgnargscaled#1{\sgn\rbr*{#1}}
\def\@sgnargscaleas[#1]#2{\sgn\rbr[#1]{#2}}
\makeatother
\DeclareMathOperator*{\argmin}{arg\,min}

\providecommand{\0}{\mathbf{0}}
\providecommand{\1}{\mathbf{1}}
\renewcommand{\aa}{\mathbf{a}}
\providecommand{\bb}{\mathbf{b}}
\providecommand{\cc}{\mathbf{c}}
\providecommand{\dd}{\mathbf{d}}
\providecommand{\ee}{\mathbf{e}}

\providecommand{\mm}{\mathbf{m}}

\providecommand{\pp}{\mathbf{p}}

\providecommand{\uu}{\mathbf{u}}

\providecommand{\xx}{\mathbf{x}}
\providecommand{\yy}{\mathbf{y}}

\providecommand{\xiv}{\boldsymbol{\xi}}
\providecommand{\mA}{\mathbf{A}}
\providecommand{\mB}{\mathbf{B}}

\providecommand{\mF}{\mathbf{F}}
\providecommand{\mG}{\mathbf{G}}

\providecommand{\mI}{\mathbf{I}}
\providecommand{\mJ}{\mathbf{J}}

\providecommand{\mP}{\mathbf{P}}

\providecommand{\mS}{\mathbf{S}}

\providecommand{\mU}{\mathbf{U}}
\providecommand{\mV}{\mathbf{V}}
\providecommand{\mW}{\mathbf{W}}
\providecommand{\mX}{\mathbf{X}}
\providecommand{\mY}{\mathbf{Y}}
\providecommand{\mZ}{\mathbf{Z}}

\providecommand{\mpi}{\bm{\pi}}

\providecommand{\cD}{\mathcal{D}}
\providecommand{\cE}{\mathcal{E}}

\providecommand{\cG}{\mathcal{G}}

\providecommand{\cN}{\mathcal{N}}
\providecommand{\cO}{\mathcal{O}}

\providecommand{\cV}{\mathcal{V}}

\newtheorem{theorem}{Theorem}

\newtheorem{proposition}[theorem]{Proposition}

\newtheorem{lemma}{Lemma}
\newtheorem{remark}[lemma]{Remark}

\newtheorem{definition}{Definition}

\newtheorem{assumption}[definition]{Assumption}

\makeatletter
\newcommand{\myitem}[1]{%
\item[\textbf{(#1)}]\protected@edef\@currentlabel{#1}%
}
\makeatother

\providecommand{\dpsgd}{DP-SGD\xspace}
\providecommand{\dsquare}{D$^2$\xspace}
\providecommand{\gradtrack}{Gradient Tracking\xspace}
\providecommand{\cifar}{Cifar-10\xspace}
\providecommand{\imagenet}{ImageNet\xspace}
\providecommand{\bert}{BERT\xspace}
\providecommand{\vgg}{VGG-11\xspace}

\providecommand{\half}{\nicefrac{1}{2}}
\newcommand\tsum{\textstyle\sum\nolimits}

\newcommandx*\prm[2][1=x]{\mX_{#2}^{\ifthenelse{ \equal{#1}{x} }{}{(#1)}}}

\newcommand{\tablefontsize}{\scriptsize}

\hypersetup{
    allbordercolors = [rgb]{0.090, 0.745, 0.811}, %
}

\providecommand{\RelaySum}{RelaySum\xspace}
\providecommand{\RelaySumGrad}{RelaySGD/Grad\xspace}
\providecommand{\RelaySumModel}{RelaySGD\xspace}
\providecommand{\taumax}{\tau_\text{max}}

\title{\RelaySum for Decentralized Deep Learning\\ on Heterogeneous Data}

\author{
Thijs Vogels\thanks{Equal contribution. Corresponding authors \texttt{thijs.vogels@epfl.ch} and \texttt{lie.he@epfl.ch}.} \\ EPFL
\And Lie He$^*$  \\ EPFL
\And Anastasia Koloskova  \\ EPFL
\And Tao Lin  \\ EPFL
\AND Sai Praneeth Karimireddy  \\ EPFL
\And Sebastian U.\ Stich  \\ EPFL
\And Martin Jaggi \\ EPFL
}

\begin{document}

\maketitle

\begin{abstract}
      In decentralized machine learning, workers compute model updates on their local data.
      Because the workers only communicate with few neighbors without central coordination, these updates propagate progressively over the network.
      This paradigm enables distributed training on networks without all-to-all connectivity, helping to protect data privacy as well as to reduce the communication cost of distributed training in data centers.
      A key challenge, primarily in decentralized deep learning, remains the handling of differences between the workers' local data distributions.
      To tackle this challenge, we study the \RelaySum mechanism for information propagation in decentralized learning.
      \RelaySum uses spanning trees to distribute information exactly uniformly across all workers with finite delays depending on the distance between nodes.
      In contrast, the typical gossip averaging mechanism only distributes data uniformly asymptotically while using the same communication volume per step as \RelaySum.
      We prove that \RelaySumModel, based on this mechanism, is independent of data heterogeneity and scales to many workers, enabling highly accurate decentralized deep learning on heterogeneous data.
      Our code is available at \url{http://github.com/epfml/relaysgd}.
\end{abstract}

\section{Introduction}

Ever-growing datasets lay at the foundation of the recent breakthroughs in machine learning.
Learning algorithms therefore must be able to leverage data distributed over multiple devices, in particular for reasons of efficiency and data privacy.
There are various paradigms for distributed learning, and they differ mainly in how the devices collaborate in communicating model updates with each other.
In the \emph{all-reduce} paradigm, workers average model updates with all other workers at every training step.
In \emph{federated learning}~\citep{mcmahan2017fedavg}, workers perform local updates before sending them to a central server that returns their global average to the workers.
Finally, \emph{decentralized learning} significantly generalizes the two previous scenarios.
Here, workers communicate their updates with only few directly-connected neighbors in a network, without the help of a server.

Decentralized learning offers strong promise for new applications, allowing any group of agents to collaboratively train a model while respecting the data locality and privacy of each contributor~\citep{nedic2020review}.
At the same time, it removes the single point of failure in centralized systems such as in federated learning~\citep{kairouz2019federated}, improving robustness, security, and privacy.
Even from a pure efficiency standpoint, decentralized communication patterns can speed up training in data centers~\citep{assran2019sgp}.

In decentralized learning, workers share their local stochastic gradient updates with the others through \emph{gossip} communication~\citep{xiao2004distavg}.
They send their updates to their neighbors, which iteratively propagate the updates further into the network.
The workers typically use iterative \emph{gossip averaging} of their models with their neighbors, using averaging weights chosen to ensure asymptotic uniform distribution of each update across the network.
It will take $\tau$ rounds of communication for an update from worker $i$ to reach a worker $j$ that is $\tau$ hops away, and when it first arrives, the update is exponentially weakened by repeated averaging with weights $< 1$.
In general networks, worker $j$ will never exactly, but only asymptotically receive its uniform share of the update.
The slow distribution of updates not only slows down training, but also makes decentralized learning sensitive to heterogeneity in workers' data distributions.

\begin{figure}[t]
    \includegraphics{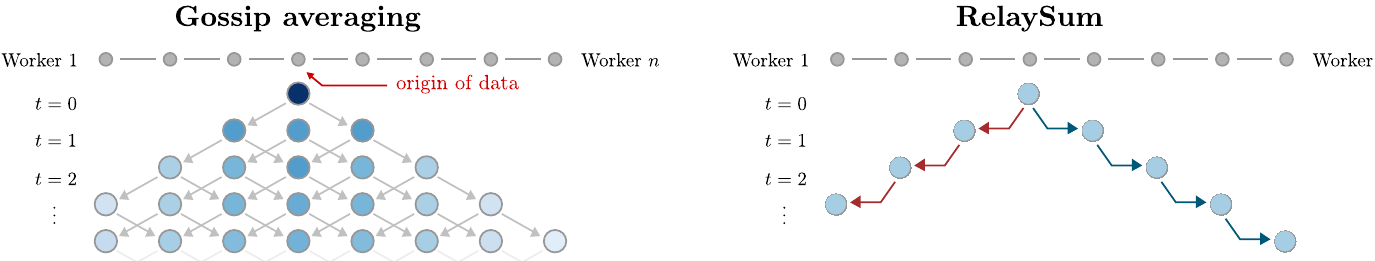}
    \vspace{-4mm}
    \caption{
        \label{fig:propagation}
        To spread information across a decentralized network, classical gossip averaging diffuses information slowly through the network.
        The left figure illustrates the spread of information originating from the fourth worker in a chain network.
        In \RelaySum, the messages are \emph{relayed} without reweighting, resulting in uniform delivery of the information to every worker.
        When multiple workers broadcast simultaneously (not pictured), \RelaySum can \emph{sum} their messages and use the same bandwidth as gossip averaging.
    }\vspace{-2mm}
\end{figure}

We study an alternative mechanism to gossip averaging, which we call \RelaySum.
\RelaySum operates on spanning trees of the network, and distributes information exactly uniformly within a finite number of gossip steps equal to the diameter of the network.
Rather than iteratively averaging models, each node acts as a `router' that \emph{relays} messages through the whole network without decaying their weight at every hop.
While naive all-to-all routing requires $n^2$ messages to be transmitted at each step, we show that on trees, only $n$ messages (one per edge) are sufficient.
This is enabled by the key observation that the routers can \emph{merge} messages by \emph{summation} to avoid any extra communication compared to gossip averaging.
\RelaySum achieves this using additional memory linear in the number of edges, and by tailoring the messages sent to different neighbors.
At each time step, \RelaySum workers receive a uniform average of exactly one message from each worker.
Those messages just originate from different time delays depending on how many hops they travelled.
The difference between gossip averaging and \RelaySum is illustrated in \autoref{fig:propagation}.

The \RelaySum mechanism is structurally similar to Belief Propagation algorithms for inference in graphical models. This link was made by \citet{zhang2019bp}, who used the same mechanism for decentralized weighted average consensus in control.

We use \RelaySum in the \RelaySumModel learning algorithm.
We theoretically show that this algorithm is not affected by differences in workers' data distributions.
Compared to other algorithms that have this property~\citep{tang2018d2,pu2018dsgt}, \RelaySumModel does not require the selection of averaging weights, and its convergence does not depend on the spectral gap of the averaging matrix, but instead on the network diameter.

While \RelaySum is formulated for trees, it can be used in any decentralized network.
We use the Spanning Tree Protocol~\citep{perlman85spanningtreeprotocol} to construct spanning trees of any network in a decentralized fashion.
\RelaySumModel often performs better on any such spanning tree than gossip-based methods on the original graph.
When the communication network can be chosen freely, the algorithm can use double binary trees~\citep{sanders2009trees}.
While these trees have logarithmic diameter and scale to many workers, \RelaySumModel in this setup uses only constant memory equivalent to two extra copies of the model parameters and sends and receives only two models per iteration.

Surprisingly, in deep learning with highly heterogeneous data, prior methods that are theoretically independent of data heterogeneity~\citep{tang2018d2,pu2018dsgt}, perform worse than heuristic methods that do not have this property, but use cleverly designed time-varying communication topologies~\citep{assran2019sgp}.
In extensive tests on image- and text classification, \RelaySumModel performs better than both kinds of baselines at equal communication budget.

\section{Related work}

Out of the multitude of decentralized optimization methods, first-order algorithms that interleave local gradient updates with a form of gossip averaging~\cite{nedic2017diging,johansson2009randomized} show most promise for deep learning.
Such algorithms are theoretically analyzed for convex and non-convex objectives in \cite{Nedic2009:distributedsubgrad,johansson2009randomized,nedic2017diging}, and
\citep{lian2017dpsgd,tang2018d2,assran2019sgp,lin2021quasiglobal} demonstrate that gossip-based methods can perform well in deep learning.

In a gossip averaging step, workers average their local models with the models of their direct neighbors.
The corresponding `mixing matrix' is a central object of study.
The matrix can be doubly-stochastic \cite{nedic2017diging,lian2017dpsgd,koloskova2021unified}, column-stochastic \cite{tsianos2012push,nedic2016sgp,xi2017dextra,assran2019sgp}, row-stochastic \cite{xi2017linear,xin2019frost}, or a combination \cite{xin2018linear,xin2019distributed,pu2021pushpull}.
Column-stochastic methods use the \textit{push-sum} consensus mechanism \citep{kempe2003pushsum} and can be used on directed graphs.
Our analysis borrows from the theory developed for those methods.

While gossip averages in general requires an infinite number of steps to reach exact consensus, another  line of work identifies mixing schemes that yield exact consensus in finite steps.
For some graphs, this is possible with time-independent averaging weights~\cite{Ko2010,Georgopoulos2011}.
One can also achieve finite-time consensus with time-varying mixing matrices.
On trees, for instance, exact consensus can be achieved by routing updates to a root node and back, in exactly diameter number of steps~\cite{Ko2010,Georgopoulos2011}.
On some graphs, tighter bounds can be established~\cite{Hendrickx2014:finitetime}.
For fully-connected networks with $n$ workers, \citet{assran2019sgp} design a sparse time-varying communication scheme that yields exact consensus in a cycle of $\log n$ averaging steps and performs well in deep learning.

The `relay' mechanism of \RelaySumModel was previously used by \citet{zhang2019bp} in the control community for the decentralized weighted average consensus problem, but they do not use it in the context of optimization. Zhang et al.\ also introduce a modified algorithm for loopy graphs, but this modification makes the achieved consensus inexact. The `relay' mechanism effectively turns a sparse graph into a fully-connected graph with communication delays. Work on delayed consensus~\cite{nedic2010convergence} and
optimization \cite{tsianos2011distributed,agarwal2012distributed} analyzes such schemes for centralized distributed algorithms. Those consensus schemes are, however, not directly applicable to decentralized optimization.

A fundamental challenge in decentralized learning is dealing with data that is not identically distributed among workers.
Because, in this case, workers pursue different optima, workers may drift~\cite{nedic2017diging} and this can harm convergence.
There is a large family of algorithms that introduce update corrections that provably mitigate such data heterogeneity.
Examples applicable to non-convex problems are exact diffusion~\citep{yuan2019exact-diff-1}, \gradtrack~\citep{Lorenzo2016GT-first-paper, pu2018dsgt, zhang2020GT-non-convex}, \dsquare~\citep{tang2018d2}, PushPull~\citep{pu2021pushpull}.
To tackle the same challenge, \citet{lin2021quasiglobal,yuan2021decentlam} propose modifications to local momentum to empirically improve performance in deep learning, but without provable guarantees.
\citet{pmlr-v139-lu21a} propose DeTAG which overlaps multiple consecutive gossip steps and gradient computations to accelerate information diffusion.
This technique could be applied to the \RelaySum mechanism, too.

\section{Method}\label{sec:method}

\paragraph{Setup}
We consider standard decentralized optimization with data distributed over $n\geq 1$ nodes:
\begin{equation*}\textstyle
    f^\star:= \min_{\xx\in\R^d} \left[f(\xx) = \frac{1}{n} \sum_{i=1}^n \left[ f_i(\xx) := \expect_{\xi\sim \cD_i} F_i(\xx, \xi_i)  \right] \right] \,.
\end{equation*}
Here $\cD_i$ denotes the distribution of the data on node $i$ and $f_i \colon \R^d \to \R$ the local optimization objectives.
Workers are connected by a network respecting a graph topology $\cG=(\cV, \cE)$, where $\cV=\{1,\ldots,n\}$ denotes the set of workers, and $\cE$ the set of undirected communication links between them (without self loops).
Each worker $i$ can only directly communicate with its neighbors $\cN_i \subset \cV$.

\paragraph{Decentralized learning with gossip}
We consider synchronous first-order algorithms that interleave local gradient-based updates\vspace{-2mm}
\begin{align*}
    \xx_i^{(t+\half)} = \xx_i^{(t)} + \uu_i^{(t)}
\end{align*}
with message exchange between connected workers.
For SGD with typical gossip averaging (\dpsgd~\citep{lian2017dpsgd}), the local updates can be written as $\uu_i^{(t)}=-\gamma \nabla f_i(\xx_i^{(t)}, \xi_i^{(t)})$, and the messages exchanged between pairs of connected workers $(i,j)$ are $\mm_{i\to j}^{(t)} = \xx_i^{(t+\half)} \in \R^d$.
Each timestep, the workers average their model with received messages,
\begin{equation}\textstyle
    \xx_i^{(t+1)} = \mW_{ii} \xx_i^{(t+\half)} + \sum_{j\in \cN_i} \mW_{ij} \mm_{j \to i}^{(t)}, \label{eq:dsgd}\tag{DP-SGD}
\end{equation}
using averaging weights defined by a \emph{gossip matrix} $\mW \in \R^{n \times n}$.

In this scheme, an update $\uu_i^{(t_1)}$ from any worker $i$ will be linearly incorporated into the model $\xx_j^{(t_2)}$ at a later timestep $t_2$ with weight $(\mW^{t_2 - t_1})_{ij}$.
The gossip matrix must be chosen such that these weights asymptotically converge to $\frac{1}{n}$, distributing all updates uniformly over the workers. This setup appears in, for example, \citep{lian2017dpsgd,koloskova2021unified}.

\paragraph{Uniform model averaging}
If the graph topology is fully-connected, any worker can communicate with any other worker, and it is ideal to use `all-reduce averaging',
\begin{equation*}\textstyle
    \xx_i^{(t+1)} = \frac{1}{n} \sum_{j = 1}^n \xx_j^{(t+\half)}.
\end{equation*}
Contrary to the decentralized scheme~\eqref{eq:dsgd}, this algorithm does not degrade in performance if data is distributed heterogeneously across workers.
In sparsely connected networks, however, all-reduce averaging requires routing messages through the network.
On arbitrary networks, such a routing protocol requires at least a number of communication steps equal to the network diameter $\taumax$---the minimum number of hops some messages have to travel.

\paragraph{\RelaySumModel} In this paper, we approximate the all-reduce averaging update as
\begin{equation}\label{eq:relaysum:model}\textstyle
    \xx_i^{(t+1)} = \frac{1}{n} \sum_{j = 1}^n \xx_j^{(t  - \tau_{ij} +\half)}, \tag{\RelaySumModel}
\end{equation}
where $\tau_{ij}$ is minimum number of network hops between workers $i$ and $j$ (and $\tau_{ii}=0$).
Since it takes $\tau_{ij}$ steps to route a message from worker $i$ to $j$, this scheme could be implemented using a peer-to-peer routing protocol like Ethernet.
Of course, this naive implementation drastically increases the bandwidth used compared to gossip averaging.
The key insight of this paper is that, on tree networks, the \RelaySumModel update rule can be implemented while using the same communication volume per step as gossip averaging, using additional memory linear in the number of a worker's direct neighbors.

\paragraph{\RelaySum}
To implement \RelaySumModel, we require a communication mechanism that delivers sums of delayed `parcels' $s_w^{(t)} = \sum_{j =1}^n p_j^{(t - \tau_{wj})}$ to each worker $w$ in a tree network, where the parcel $p_j^{(t)}$ is created by worker $j$ at time $t$.
To simplify the exposition, let us first consider the simplest type of tree network: a chain.
In a chain, a worker $w$ is connected to workers $w-1$ and $w+1$, if those exist, and the delays are $\tau_{ij} = |i-j|$. We can then decompose
\begin{align*}
    s_w^{(t)} = \sum_{j =1}^n p_j^{(t - \tau_{wj})} = p_w^{(t)} + \underbrace{\sum_{j=1}^{w-1} p_j^{(t-\tau_{wj})}}_\text{parcels from the `left'} + \underbrace{\sum_{j=w+1}^{n} p_j^{(t-\tau_{wj})}}_\text{parcels from the `right'}.
\end{align*}
The sum of parcels from the `left' will be sent as one message $m_{(w-1) \to w}$ from worker $w-1$ to $w$, and the sum of data from the `right' will be sent as one message $m_{(w+1) \to w}$ from $w+1$ to $w$.
Neighboring workers can compute these messages from the messages they received from their neighbors in the previous timestep.
Compared to typical gossip averaging, \RelaySum requires additional memory linear in the number of neighbors, but it uses the same volume of communication.

Algorithm~\ref{alg:relaysum_model} shows how this scheme is generalized to general tree networks and incorporated into \RelaySumModel. Along with the model parameters, we send scalar counters that are used in the first few iterations of the algorithm $t \leq \taumax$ to correct for messages that have not yet arrived.

\begin{algorithm}[t]
    \algrenewcommand\algorithmicrequire{\textbf{Input:}}
    \algblockdefx[NAME]{ParallelFor}{ParallelForEnd}[1]{\textbf{for} #1 \textbf{in parallel}}{\textbf{end for}}    \caption{\RelaySumModel}\label{alg:relaysum_model}
    \begin{algorithmic}[1] %
        \Require $\forall~i,~\xx_i^{(0)}=\xx^{(0)}$; $\forall~i,j, \mm^{(-1)}_{i\rightarrow j}=\0$, counts $c^{(-1)}_{i\rightarrow j}=0$, learning rate $\gamma$, tree network
        \For{$t=0,1,\ldots$}
        \ParallelFor{node $i$}
        \State $\xx^{(t+\half)}_i=\xx^{(t)}_i {\color{tab10_blue}- \gamma \nabla f_i(\xx_i^{(t)})}$\hfill{\color{tab10_blue}(or Adam/momentum)}
        \For{each neighbor $j\in \cN_i$}
        \State Send $\mm^{(t)}_{i\rightarrow j}=x_i^{(t+\half)} {\color{tab10_blue}+ \sum_{k\in\cN_i\backslash j}\mm^{(t-1)}_{k\rightarrow i}}$\hfill{\color{tab10_blue}(relay messages from other neighbors)}
        \State Send corresponding counters $c^{(t)}_{i\rightarrow j}=1 + \sum_{k\in\cN_i\backslash j} c^{(t-1)}_{k\rightarrow i}$
        \State Receive ($\mm^{(t)}_{j\rightarrow i}$, $c^{(t)}_{j\rightarrow i}$) from node $j$
        \EndFor
        \vspace{1.5mm}
        \State $\bar{n}_i^{(t+1)} = 1+\sum_{j\in\cN_i} c^{(t)}_{j\rightarrow i}$ \hfill {\color{gray}($\bar{n}$ converges to the total number of workers)}
        \State $\xx_i^{t+1}=\frac{1}{\bar{n}_i^{(t+1)}} \left( \xx_i^{(t+\half)}+\sum_{j\in\cN_i} \mm^{(t)}_{j\rightarrow i} \right)$ \hfill{\color{gray}$\left(=\frac{1}{n}\sum_{j=1}^n \xx_j^{(t -\tau_{ij} + \half)}\right)$}
        \ParallelForEnd
        \EndFor
    \end{algorithmic}
\end{algorithm}

\paragraph{Spanning trees}
\RelaySumModel is formulated on tree networks, but it can be used on any communication graph by constructing a spanning tree.
In a truly decentralized setting, we can use the Spanning Tree Protocol~\citep{perlman85spanningtreeprotocol} used in Ethernet to find such trees in a decentralized fashion.
The protocol elects a leader as the root of the tree, after which every other node finds the fastest path to this leader.

On the other hand, when the decentralized paradigm is used in a data center to reduce communication, \RelaySumModel can run on double binary trees~\citep{sanders2009trees} used in MPI and NCCL~\citep{nvidia2019speeding}.
The key idea of double binary trees is to use two different communication topologies for different parts of the model.
We communicate odd coordinates using a balanced binary tree $A$, and communicate the even coordinates with a complimentary tree $B$.
The trees $A$ and $B$ are chosen such that internal nodes (with 3 edges) in one tree are leaves (with only 1 edge) in the other.
Using the combination of two trees, \RelaySumModel requires only constant extra memory equivalent to at most 2 model copies (just like the Adam optimizer~\citep{kingma2015adam}), and it sends and receives the equivalent of 2 models (just like on a ring).

\section{Theoretical analysis}\label{sec:theoretical_analysis}

Since \RelaySumModel updates worker's models at time step $t+1$ using models from (at most) the past $\tau_{\max}$ steps, we conveniently reformulate \RelaySumModel in the following way:
Let $\mY^{(t)}, \mG^{(t)} \in\R^{n(\tau_{\max}+1)\times d}$ denote stacked worker models and gradients whose row vectors at index $n\!\cdot\!\tau+i$ represent
\begin{align*}
    \left[\mY^{(t)}\right]_{n\tau+i}^\top = \begin{cases}
        \xx_i^{(t-\tau)} & t\ge\tau         \\
        \xx^{(0)}        & \text{otherwise}
    \end{cases},
    \qquad
    \left[\mG^{(t)}\right]_{n\tau+i}^\top = \begin{cases}
        \nabla F_i(\xx_i^{(t-\tau)}; \xi_i^{(t-\tau)}) & t\ge\tau         \\
        \xx^{(0)}                                      & \text{otherwise}
    \end{cases}
\end{align*}
for all times $t\ge0$, delay $\tau\in[0, \tau_{\max}]$ and worker $i\in[n]$.
Then \eqref{eq:relaysum:model} can be written as
\begin{align*}%
    \mY^{(t+1)} = \mW \mY^{(t)} - \gamma \tilde{\mW}\mG^{(t)}
\end{align*}
where $\mW,\tilde{\mW}\in\R^{n(\tau_{\max}+1)\times n(\tau_{\max}+1)}$ are non-negative matrices whose elements are
\begin{align*}
    \left[\mW\right]_{n\tau+i, n\tau'+j} = \begin{cases}
        \tfrac{1}{n} & \tau=0 \text{ and } \tau'=\tau_{ij} \\
        1            & i=j \text{ and } \tau=\tau'+1       \\
        0            & \text{otherwise}
    \end{cases},
    \qquad
    \left[\tilde{\mW}\right]_{n\tau+i, n\tau'+j} = \begin{cases}
        \tfrac{1}{n} & \tau=0 \text{ and } \tau'=\tau_{ij} \\
        0            & \text{otherwise}
    \end{cases}
\end{align*}
for all $\tau,\tau'\in[0, \tau_{\max}]$ and $i,j\in[n]$.
The matrix $\mW$ can be interpreted as the mixing matrix of an `augmented graph'~\citep{nedic2010convergence} with additional virtual `forwarding nodes'.
$\mW$ is row stochastic and its largest eigenvalue is 1.
The vector of all ones $\1_{n(\tau_{\max}+1)}\in\R^{n(\tau_{\max}+1)}$ is a right eigenvector of $\mW$ and let $\mpi\in\R^{n(\tau_{\max}+1)}$ be the left eigenvector such that $\mpi^\top\1_{n(\tau_{\max}+1)}=1$.%

We characterize the convergence rate of the consensus distance in the following key lemma:
\begin{lemma}[Key lemma]\label{lemma:algo1:key}
    There exists an integer $m=m(\mW)>0$ such that for any $\mX\in\R^{n(\tau_{\max}+1)\times d}$ we have\vspace{-3mm}
    \begin{align*}
        \| \mW^m \mX - \1\mpi^\top \mX \|^2 \le (1-p)^{2m} \| \mX - \1\mpi^\top \mX \|^2,
    \end{align*}
    where $p=\frac12(1-|\lambda_2(\mW)|)$ is a constant.\vspace{-1mm}
\end{lemma}
All the following optimization convergence results will only depend on the \emph{effective spectral gap} $\rho := \frac{p}{m}$ of $\mW$.
We empirically observe that $\rho =\Theta(1/n)$ for a variety of network topologies (see \autoref{fig:empirical-rho} in Appendix~\ref{apx:theory}).

\begin{remark}
    The above key lemma is similar to \citep[Assumption 4]{koloskova2021unified} for gossip-type averaging with symmetric matrices.
    However, in our case $\mW$ is just a row stochastic matrix, and its spectral norm $\|\mW\|_2>1$.
    In general, the consensus distance can increase after just one single communication step (multiplication by $\mW$).
    That is why we need $m > 1$.
    The proof of the Lemma relies on a Perron-Frobenius type theorem, and holds over several steps $m$ instead of a single iteration.
    It means \RelaySum defines a consensus algorithm with linear convergence rate which pulls models closer.
\end{remark}

Our main convergence results hold under the following common assumptions, as e.g.~\cite{koloskova2021unified}.
\begin{assumption}[L-smoothness]\label{assumption:smoothness:stochastic}
    For each $i\in[n]$, $F_i(\xx, \xi): \R^D\times\Omega_i\rightarrow\R$ is differentiable for each $\xi\in\text{supp}(\cD_i)$ and there exists a constant $L\ge0$ such that for each $\xx,\yy\in\R^d$, $\xi\in\text{supp}(\cD_i)$:
    \begin{equation*}
        \| \nabla F_i(\xx, \xi) - \nabla F_i(\yy, \xi) \| \le L \| \xx - \yy \| \,.
    \end{equation*}
\end{assumption}
\begin{assumption}[Uniform bounded noise]\label{assumption:uniform_sigma_zeta}
    There exists constant $\bar{\sigma}$, such that for all $\xx \in\R^d$, $i\in[n]$,
    \begin{align*}
        \E_{\xi}\| \nabla F_i(\xx, \xi) - \nabla f_i(\xx) \|^2 \le \bar{\sigma}^2.
    \end{align*}
\end{assumption}

\begin{assumption}[$\mu$-convexity]\label{assumption:convexity}
    For $i\in[n]$, each function $f_i:\R^d\rightarrow \R$ is $\mu$-(strongly) convex for constant $\mu\ge0$. That is, $\forall~ \xx,\yy\in\R^{d}$\vspace{-1mm}
    \begin{align*}
        f_i(\xx) - f_j(\yy) + \tfrac{\mu}{2} \|\xx-\yy\|^2_2 \le  \nabla f_i(\xx)^\top (\xx-\yy) \,.
    \end{align*}
\end{assumption}

\begin{theorem}[\RelaySumModel]
    \label{thm:1}
    For any target accuracy $\epsilon>0$ and an optimal solution $\xx^\star$,
    \vspace{-2mm}

    \textbf{(Convex:)} under Assumptions~\ref{assumption:smoothness:stochastic}, \ref{assumption:uniform_sigma_zeta} and \ref{assumption:convexity} with $\mu\ge0$, it holds that $\tfrac{1}{T+1}\tsum_{t=0}^T \left(f(\overline\xx^{(t)}) \!-\! f(\xx^\star)\right)\le \epsilon$ after\vspace{-1mm}
    \[
        \textstyle
        \cO\left(\frac{\bar{\sigma}^2}{n \epsilon^2}
        + \frac{C\sqrt{L}\bar{\sigma}}{\epsilon^{\nicefrac{3}{2}}}
        + \frac{CL}{\epsilon}
        \right) R_0^2
    \]
    iterations.
    Here $\overline\xx^{(t)}\!:=\!\mpi^\top\mY^{(t)}$ averages past models,  $R_0^2\!=\!\| \xx^0 - \xx^\star \|^2$, and
    $C\!=\!\cO(\frac{1}{\rho}\tau_{\max}^{\nicefrac{3}{2}})$.
    \textbf{(Non-convex:)} under Assumptions~\ref{assumption:smoothness:stochastic} and \ref{assumption:uniform_sigma_zeta}, it holds that $\tfrac{1}{T+1}\tsum_{t=0}^T \| \nabla f(\overline\xx^{(t)}) \|^2  \le \epsilon$ after\vspace{-1mm}
    \[
        \textstyle
        \cO\left(\frac{\bar{\sigma}^2}{n \epsilon^2}
        + \frac{C \bar{\sigma}}{\epsilon^{\nicefrac{3}{2}}}
        + \frac{C}{\epsilon}
        \right) L F_0
    \]
    iterations, where $F_0 := f(\overline\xx^{(0)})-f(\xx^\star)$.
\end{theorem}
The dominant term in our convergence result, $\cO\bigl(\frac{\bar\sigma^2}{n\epsilon^2} \bigr)$ matches with the dominant term in the convergence rate of centralized (`all-reduce') mini-batch SGD, and thus can not be improved.

In contrast to other methods, the presented convergence result of \RelaySumModel is independent of the data heterogeneity $\zeta^2$ in
\citep[Assumption 3b]{koloskova2021unified}.
\begin{definition}[Data heterogeneity] \label{assumption:zeta2}
    There exists a constant $\zeta^2$ such that  $\forall~i\in[n], \xx\in\R^d$
    \begin{equation*}\textstyle
        \| \nabla f_i(\xx) - \nabla f(\xx) \|_2^2 \le \zeta^2 \,.
    \end{equation*}
\end{definition}\vspace{-1mm}
\begin{remark}
    For convex objectives, Assumptions~\ref{assumption:uniform_sigma_zeta} and \ref{assumption:zeta2} can be relaxed to only hold at the optimum~$\xx^\star$.
    A weaker variant of \Cref{assumption:smoothness:stochastic} only uses $L$-smoothness of $f_i$ \citep[Assumption 1b]{koloskova2021unified}.
\end{remark}\vspace{-1.3mm}
Comparing to gossip averaging for convex $f_i$ which has complexity
$\cO(\tfrac{\bar{\sigma}^2}{n\epsilon^2}
    + (\tfrac{\zeta}{\rho}
    + \tfrac{\bar{\sigma}}{\sqrt{\rho}})
    \frac{\sqrt{L}}{\epsilon^{\nicefrac{3}{2}}}
    +\tfrac{L}{\rho\epsilon}
    ) R_0^2\vspace{-1mm}$, our rate for \RelaySumModel does not depend on $\zeta^2$ and has same leading term $\cO(\frac{\bar{\sigma}^2}{n\epsilon^2})$ as $D^2$.

\section{Experimental analysis and practical properties}\label{sec:experiments}

\subsection{Effect of network topology}

\paragraph{Random quadratics}
To efficiently investigate the scalability of \RelaySumModel with respect to the number of workers,
and to study the benefits of binary tree topologies over chains, we introduce a family of synthetic functions.
We study \emph{random quadratics} with local cost functions $f_i(\xx) = \norm{\mA_i \xx - \bb_i^\top \xx}^2$ to precisely control all constants that appear in our theoretical analysis.
The Hessians~$\mA_i$ are initialized randomly, and their spectrum is scaled to achieve a desired smoothness~$L$ and strong convexity $\mu$.
The offsets $\bb_i$ ensure a desired level of heterogeneity $\zeta^2$ %
and distance between optimum and initialization $r_0$. Appendix~\ref{apx:setup:quadratics} describes the generation of these quadratics in detail.

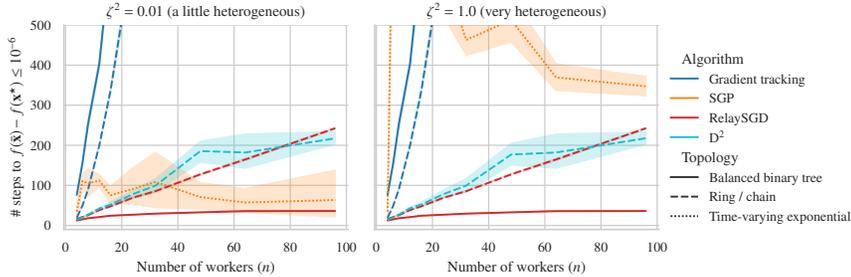
\begin{figure}[ht]
    \DeclareRobustCommand\dotGray{\tikz{\fill[fill=gray] (0,0) rectangle (5pt,5pt);}~}
    \DeclareRobustCommand\dotBlue{\tikz{\fill[fill=tab10_blue] (0,0) rectangle (5pt,5pt);}~}
    \DeclareRobustCommand\dotOrange{\tikz{\fill[fill=tab10_orange] (0,0) rectangle (5pt,5pt);}~}
    \DeclareRobustCommand\dotCyan{\tikz{\fill[fill=tab10_cyan] (0,0) rectangle (5pt,5pt);}~}
    \DeclareRobustCommand\dotRed{\tikz{\fill[fill=tab10_red] (0,0) rectangle (5pt,5pt);}~}
    \DeclareRobustCommand\dashedLines{\tikz{\draw[white] (0,-2pt) -- (0, 2pt); \draw[densely dashed,thick] (0, 0) -- (14pt, 0);}}
    \DeclareRobustCommand\dottedLines{\tikz{\draw[white] (0,-2pt) -- (0, 2pt); \draw[dotted,thick] (0, 0) -- (13pt, 0);}}
    \DeclareRobustCommand\solidLines{\tikz{\draw[white] (0,-2pt) -- (0, 2pt); \draw[thick] (0, 0) -- (14pt, 0);}}
    \centering
    \includegraphics[height=.28\textwidth]{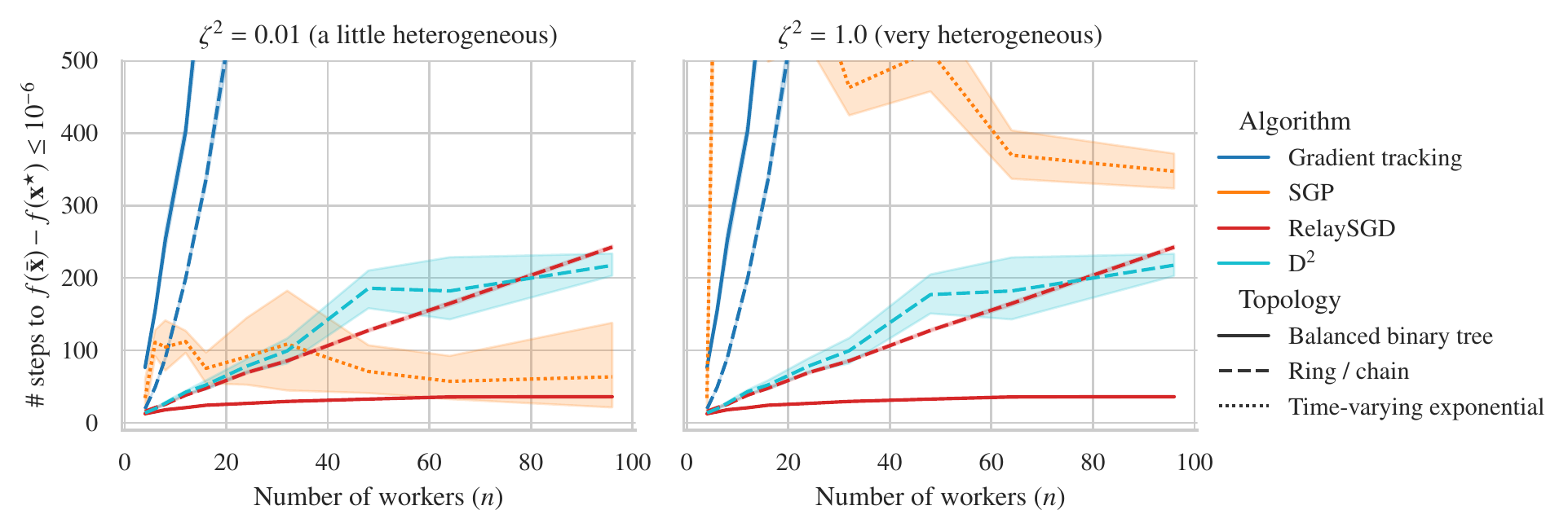}\hfill
    \vspace{-4mm}
    \caption{
        \label{fig:effect_of_network_topology_linear}
        Time required to optimize random quadratics ($\sigma^2=0, r_0=10, L=1, \mu=0.5$) to suboptimality $\leq 10^{-6}$ with varying numbers of workers with tuned constant learning rates.
        On a ring (\dashedLines), \dotCyan\dsquare and \dotRed \RelaySumModel require steps linear in the number of workers, and this number is \emph{independent of the data heterogeneity}.
        \RelaySumModel reduces this to $\log n$ on a balanced tree topology (\solidLines), but trees do not improve \dotCyan\dsquare or \dotBlue \gradtrack.
        For \dotOrange SGP with time-varying exponential topology (\dottedLines), the number of steps does not consistently grow with more workers, but this number becomes higher with more heterogeneity (left v.s.\ right plot).
    }
\end{figure}

\paragraph{Scalability on rings and trees}
Using these quadratics, \autoref{fig:effect_of_network_topology_linear} studies the number of steps required to reach a suboptimality $f(\bar{\xx}) - f(\xx^\star) \leq \epsilon$ with tuned constant learning rates.
On \emph{ring} topologies with uniform (1/3) gossip weights (and chains for \RelaySum), all compared methods require steps at least linear in the number of workers to reach the target quality.
\RelaySumModel and \dsquare empirically scale significantly better than \gradtrack, these methods are all independent of data heterogeneity.
On a \emph{balanced binary tree network} with Metropolis-Hastings weights~\citep{xiao2004distavg}, both \dsquare and \gradtrack notably do not scale better than on a ring, while \RelaySumModel on these trees requires only a number of steps logarithmic in the number of workers.
SGP with their time-varying exponential topology scales well, too, but it requires more steps on more heterogeneously distributed data.

\subsection{Spanning trees compared to other topologies}

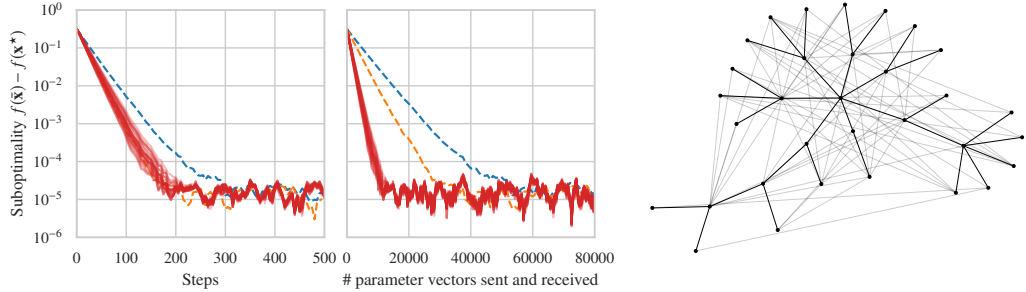
\begin{figure}[t]
    \parbox{.6\textwidth}{
        \includegraphics[width=.6\textwidth]{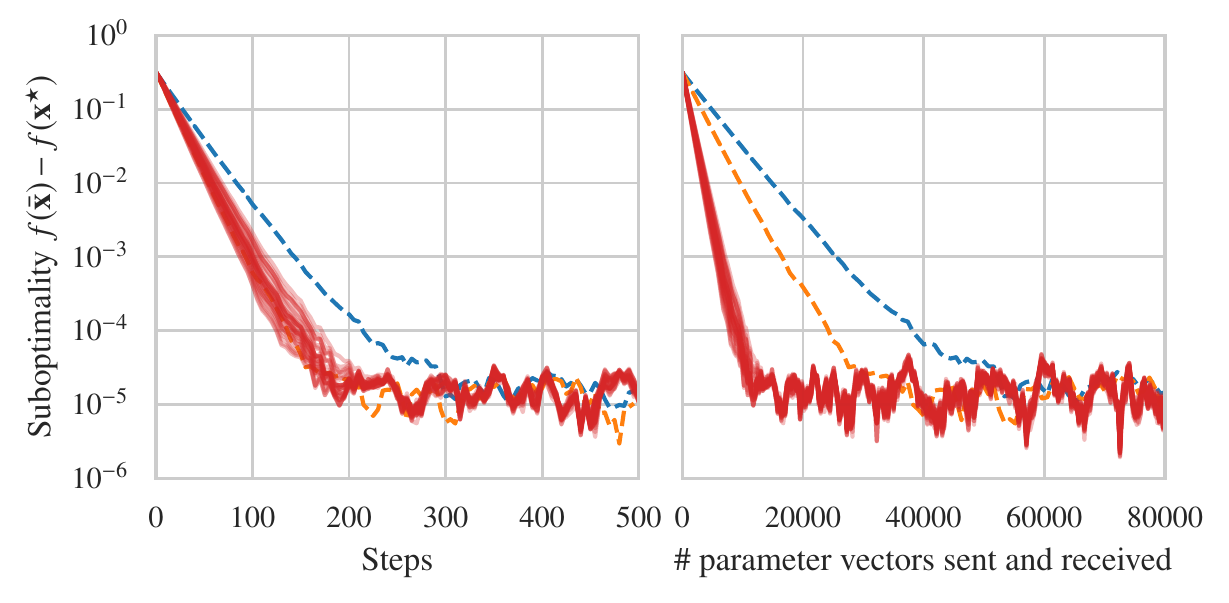}%
    }%
    \parbox{.4\textwidth}{
        \vspace{-.5cm}
        \includegraphics[width=.4\textwidth]{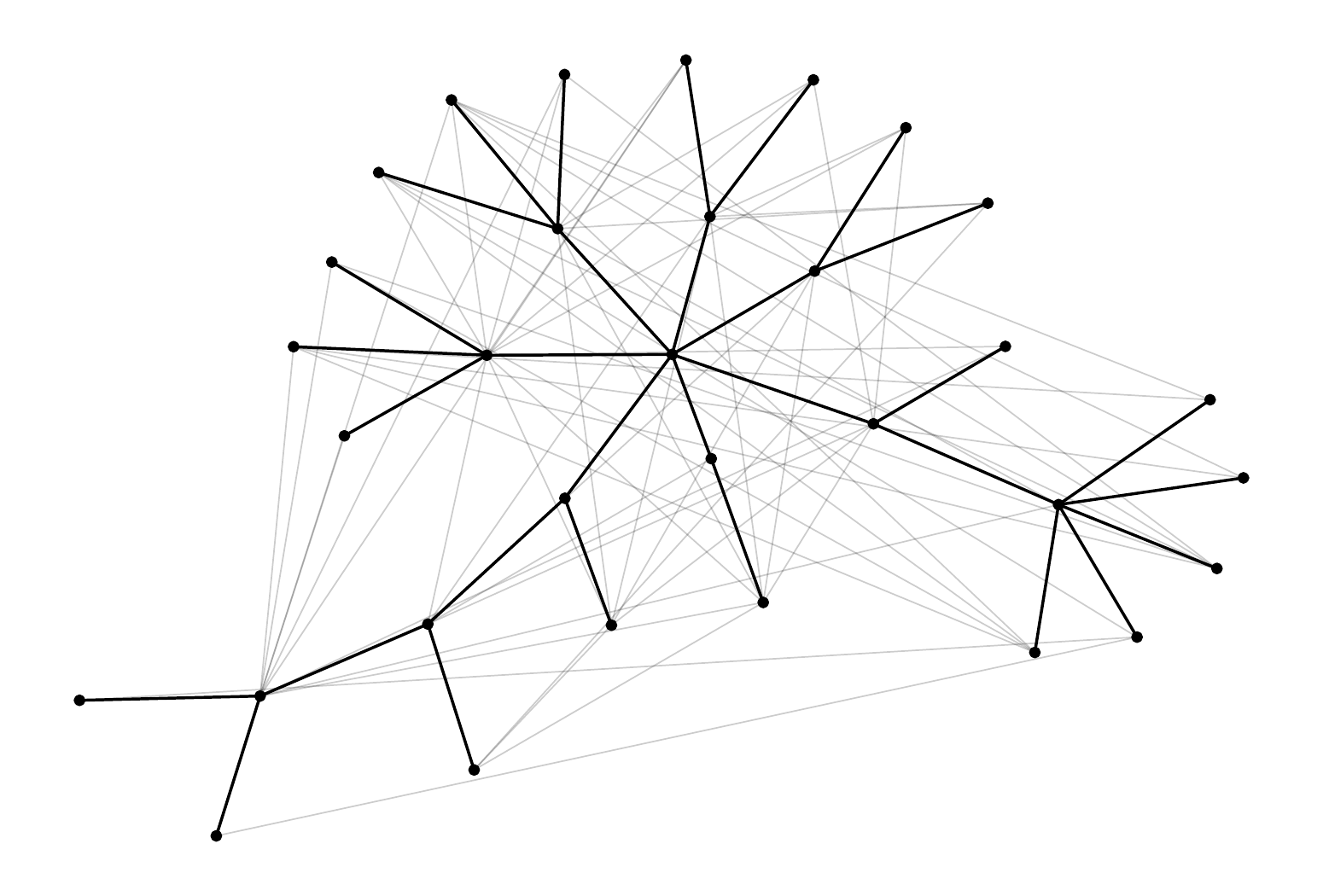}%
    }
    \vspace{-4mm}
    \DeclareRobustCommand\dotZero{\tikz{\fill[fill=gray] (0,0) rectangle (5pt,5pt);}~}
    \DeclareRobustCommand\dotOne{\tikz{\fill[fill=tab10_blue] (0,0) rectangle (5pt,5pt);}~}
    \DeclareRobustCommand\dotTwo{\tikz{\fill[fill=tab10_orange] (0,0) rectangle (5pt,5pt);}~}
    \DeclareRobustCommand\dotThree{\tikz{\fill[fill=tab10_green] (0,0) rectangle (5pt,5pt);}~}
    \DeclareRobustCommand\dotFour{\tikz{\fill[fill=tab10_red] (0,0) rectangle (5pt,5pt);}~}
    \DeclareRobustCommand\dashedLines{\tikz{\draw[white] (0,-2pt) -- (0, 2pt); \draw[densely dashed,thick] (0, 0) -- (14pt, 0);}}
    \DeclareRobustCommand\dottedLines{\tikz{\draw[white] (0,-2pt) -- (0, 2pt); \draw[densely dotted] (0, 0) -- (14pt, 0);}}
    \DeclareRobustCommand\solidLines{\tikz{\draw[white] (0,-2pt) -- (0, 2pt); \draw[thick] (0, 0) -- (14pt, 0);}}
    \caption{
        \label{fig:social_graph_spanning_tree}
        Performance of \dotFour \RelaySumModel on spanning trees of the Social Network graph (32 nodes) found using Spanning Tree Protocol, compared to \dotOne \dpsgd and \dotTwo \dsquare on the full network.
        Solid lines (\solidLines) indicate spanning trees while dashed lines (\dashedLines) indicate the full graph.
        The figure on the right shows one spanning tree on top of the original network.
        Learning rates are tuned to reach suboptimality $\leq 10^{-5}$ on random quadratics ($\zeta^2=0.1, \sigma^2=0.1, r_0=1, L=1, \mu=0.5$).
        \dotFour\RelaySumModel on spanning trees converges as fast as \dotTwo\dsquare on the full network,
        while the total communication on spanning trees is smaller than on the full graph.
        \vspace{-1mm}
    }
\end{figure}

\RelaySumModel cannot utilize all available edges in arbitrary networks to communicate, but is restricted to a spanning tree of the graph.
We empirically find that this restriction is not limiting.
In \autoref{fig:social_graph_spanning_tree}, we take an organic social network topology based on the Davis Southern Women graph~\citep{davis1930socialwomen} from NetworkX~\citep{hagberg2008networkx}, and construct random spanning trees found by the Spanning Tree Protocol~\citep{perlman85spanningtreeprotocol}.
On any such spanning tree, \RelaySumModel optimizes random heterogeneous quadratics as fast as \dsquare on the full graph with Metropolis-Hastings weights~\citep{xiao2004distavg}, significantly faster than \dpsgd.

For decentralized learning used in a fully-connected data center for communication efficiency, the deep learning experiments below show that \RelaySumModel on double binary trees outperforms the most popular non-tree-based communication scheme used in decentralized deep learning~\citep{assran2019sgp}.

\subsection{Effect of data heterogeneity in decentralized deep learning}

We study the performance of \RelaySumModel in deep-learning based image- and text classification.
While the algorithm is theoretically independent of dissimilarities in training data, other methods (\dsquare, \RelaySumGrad) that have the same property often lose accuracy in the presence of high data heterogeneity~\cite{lin2021quasiglobal}.
To study the dependence of \RelaySumModel in practical deep learning,
we partition training data strictly across 16 workers and distribute the classes using a Dirichlet process~\citep{yurochkin2019bayesian,lin2021quasiglobal}.
The Dirichlet parameter $\alpha$ controls the heterogeneity of the data across workers.

We compare \RelaySumModel against a variety of other algorithms.
\dpsgd~\citep{lian2017dpsgd} is the most natural combination of SGD with gossip averaging,
and we chose \dsquare~\citep{tang2018d2} to represent the class of previous work that is theoretically robust to heterogeneity.
We extend \dsquare to allow varying step sizes and local momentum, according to Appendix~\ref{apx:alg:d2-momentum}, and make it suitable for practical deep learning.
Although Stochastic Gradient Push~\citep{assran2019sgp} is not theoretically independent of data heterogeneity, it is a popular choice in the data center setting, where they use a time-varying exponential scheme on $2^d$ workers that mixes exactly uniformly in $d$ rounds (Appendix~\ref{apx:alg:sgp}).
We also compare to \dpsgd with quasi-global momentum~\citep{lin2021quasiglobal}, a practical method recently introduced to increase robustness to heterogeneous data.

\autoref{tab:cifar-results-trees} evaluates \RelaySumModel in the fully-connected data center setting where we limit the communication budget per iteration to two models.
We use 16-workers on \cifar, following the experimental details outlined in Appendix~\ref{apx:setup} and hyper-parameter tuning procedure from Appendix~\ref{apx:hyperparams}.
For this experiment, we consider three topologies: (1) double binary trees as described in \autoref{sec:method}, (2) rings, and (3) the time-varying exponential scheme of Stochastic Gradient Push (SGP)~\citep{assran2019sgp}. Because SGP normally sends/receives only one model per communication round, we execute two synchronous communication steps per gradient update, increasing its latency.
The various algorithms compared have different optimal topology choices.
In \autoref{tab:cifar-results-trees} we only include the optimal choice for each algorithm. \autoref{tab:best-topologies} qualitatively compares the possible combinations.
We opt for the \vgg architecture because it does not feature BatchNorm~\cite{ioffe2015batchnorm}.
BatchNorm poses particular challenges to data heterogeneity, and the search for alternatives is an active, and orthogonal, area of research~\citep{liu2020evonorm}.

\begin{figure}
    \includegraphics[width=.45\textwidth, valign=t]{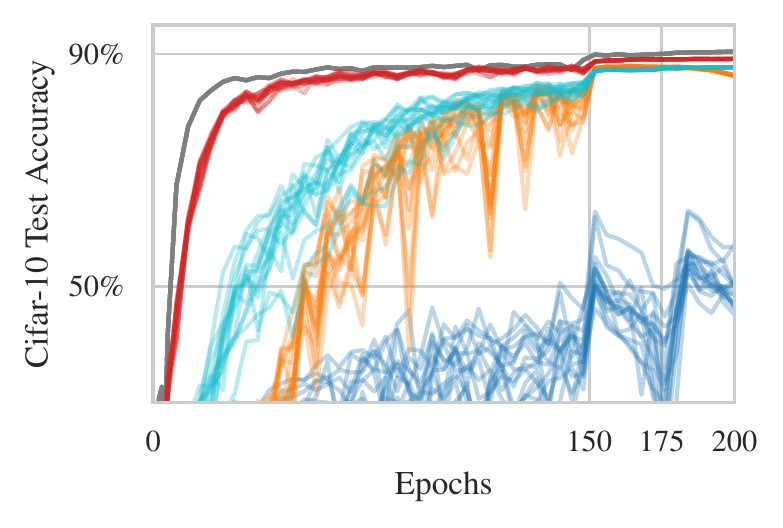}\hfill
    \begin{minipage}[t]{0.52\textwidth}%
        \DeclareRobustCommand\dotZero{\tikz{\fill[fill=darkgray] (0,0) rectangle (5pt,5pt);}~}%
        \DeclareRobustCommand\dotOne{\tikz{\fill[fill=tab10_blue] (0,0) rectangle (5pt,5pt);}~}%
        \DeclareRobustCommand\dotTwo{\tikz{\fill[fill=tab10_orange] (0,0) rectangle (5pt,5pt);}~}%
        \DeclareRobustCommand\dotThree{\tikz{\fill[fill=tab10_cyan] (0,0) rectangle (5pt,5pt);}~}%
        \DeclareRobustCommand\dotFour{\tikz{\fill[fill=tab10_red] (0,0) rectangle (5pt,5pt);}~}%
        \vspace{-2mm}
        \caption{
            \label{fig:cifar-curves}
            Test accuracy during training of 16 workers with heterogeneous data ($\alpha=0.01$) on \cifar.
            Like, with the \dotZero all-reduce baseline, all workers in \dotFour \RelaySumModel on double binary trees quickly reach good accuracy, while this takes longer for \dotThree SGP with time-varying exponential topology and \dotTwo \dsquare on a ring. \dotOne \dpsgd does not reach good accuracy with such heterogeneous data.
        }
    \end{minipage}
    \vspace{-4mm}
\end{figure}

Even though \RelaySumModel does not use a time-varying topology, it performs as well as or better than SGP, and \RelaySumModel with momentum suffers minimal accuracy loss up to heterogeneity $\alpha=0.01$, a level higher than considered in previous work~\citep{lin2021quasiglobal}.
While \dsquare is theoretically independent of data heterogeneity, and while some of its random repetitions yield good results, it is unstable in the very heterogeneous setting.
Moreover, \autoref{fig:cifar-curves} shows that workers with \RelaySumModel achieve high test accuracies quicker during training than with other algorithms.

These findings are confirmed on ImageNet \cite{deng2009imagenet} with the ResNet-20-EvoNorm architecture~\citep{liu2020evonorm} in \autoref{tab:imagenet-results}.
On the BERT fine-tuning task from \citep{lin2021quasiglobal}, \autoref{tab:bert-results} demonstrates that \RelaySumModel with the Adam optimizer, customary for such NLP tasks, outperforms all compared algorithms.

\begin{table}
    \caption{
        \cifar \cite{krizhevsky2009learning} test accuracy with the \vgg architecture.
        We vary the data heterogeneity~$\alpha$~\citep{lin2021quasiglobal} between 16 workers.
        Each method sends/receives 2 models per iteration.
        We use a ring topology for \dpsgd and \dsquare because they perform better on rings than on trees.
        \RelaySum with momentum achieves the best results across all levels of data heterogeneity.
        \label{tab:cifar-results-trees}
    }
    \vspace{-1mm}
    \input{figures/generated/cifar10-trees.tex}
\end{table}

\begin{table}
    \tablefontsize
    \caption{
        \label{tab:best-topologies}
        Motivation of topology choices.
        For each algorithm, we compare 4 topologies configured to send/receive 2 models at each SGD iteration. The algorithms have different optimal topologies.
    }
    \vspace{-1mm}
    \centering
    \newcommand{\best}{\textbf{Best result}}
    \newcommand{\equivalent}{\color{lightgray}Equivalent to \dpsgd}
    \newcommand{\unsupported}{\color{lightgray}Unsupported}
    \begin{tabularx}{\textwidth}{X l l l l}
        \toprule
        Algorithm & Ring & Chain ($=$ spanning tree of ring) & Double binary trees & Time-varying exponential~\cite{assran2019sgp}       \\
        \cmidrule(lr){1-1} \cmidrule(lr){2-5}
        \RelaySumModel & \unsupported & Worse than double b.\ trees (\ref{apx:exp:rings-vs-trees})  & \best & \unsupported \\
        \dpsgd & \best & Worse than ring & Worse than ring (\ref{apx:exp:rings-vs-trees}) & \unsupported \\
        \dsquare & \best & Worse than ring & Worse than ring (\ref{apx:exp:rings-vs-trees}) & \unsupported \\
        SGP & \equivalent & \equivalent & \equivalent & \best \\
        \bottomrule
    \end{tabularx}
\end{table}

\begin{table}
    \caption{
        Test accuracies on ImageNet, using 16 workers with heterogeneous data ($\alpha=0.1$).
        Even when communicating over a simple chain network, \RelaySumModel performs similarly to SGP with their time-varying exponential communicating scheme. Methods use default learning rates (Appendix~\ref{apx:hyper:imagenet}).
        \label{tab:imagenet-results}
    }
    \vspace{-4mm}
    \centering
    \begin{minipage}[c]{\textwidth}
        \tablefontsize
\centering
\begin{tabularx}{\textwidth}{l X l}
    \toprule
    Algorithm & Topology & Top-1 Accuracy \\
    \cmidrule(lr){1-2}     \cmidrule(lr){3-3}
    Centralized {\color{gray}(baseline)} & fully-connected & 69.7\% \\
    \textbf{\RelaySumModel w/ momentum} & double binary trees & 60.0\% \\
    \dpsgd \cite{lian2017dpsgd} w/ quasi-global momentum \cite{lin2021quasiglobal} & ring & 55.8\% \\
    \dsquare \cite{tang2018d2} w/ momentum & ring & {\color{gray}diverged at epoch 65, at 49.5\%} \\
    SGP \cite{assran2019sgp} w/ momentum & time-varying exponential \cite{assran2019sgp} & 58.5\% \\
    \bottomrule
\end{tabularx}
    \end{minipage}
\end{table}

\begin{table}
    \begin{minipage}[c]{0.51\textwidth}
        \vspace{2mm}
        \tablefontsize
\centering
\begin{tabularx}{\textwidth}{X l l}
    \toprule
    Algorithm & Topology & Top-1 Accuracy \\
    \cmidrule(lr){1-2}     \cmidrule(lr){3-3}
    Centralized Adam & fully-connected & 94.2\% $\pm$ 0.1\% \\
    \textbf{Relay-Adam} & double binary trees & 93.2\% $\pm$ 0.6\% \\
    \dpsgd Adam & ring & 87.3\% $\pm$ 0.6\% \\
    Quasi-global Adam~\citep{lin2021quasiglobal} & ring & 88.3\% $\pm$ 0.7\% \\
    SGP~\citep{assran2019sgp} Adam & time-varying exp.\ & 88.3\% $\pm$ 0.3\% \\
    \bottomrule
\end{tabularx}
    \end{minipage}\hfill
    \begin{minipage}[c]{0.47\textwidth}
        \caption{
            DistilBERT~\citep{sanh2019distilbert} fine-tuning on AG news data~\citep{zhang2015character} using 16 nodes with heterogeneous data ($\alpha=0.1$).
            Transformers are usually trained with Adam,
            and \RelaySumModel naturally supports Adam updates.
            (Appendix~\ref{apx:setup:bert}).
            \label{tab:bert-results}
        }
        \vspace{-1em}
    \end{minipage}
\end{table}

\begin{table}
    \caption{
        Robustness to unreliable networks.
        On \cifar/\vgg with 16 workers and heterogeneous data ($\alpha=0.01$), we compare momentum versions of the best-performing algorithms from \autoref{tab:cifar-results-trees}.
        Like gossip-based algorithms, \RelaySumModel with the robust update rule \ref{eq:robust-relaysum-model} can tolerate up to 10\% dropped messages and converge to full test accuracy.
        Without modification, \dsquare does not share this property.    
        \label{tab:robustness}
    }
    \vspace{-1mm}
    \centering
    \tablefontsize
\begin{tabularx}{\textwidth}{l X l l l}
    \toprule
      Algorithm & Topology & Reliable network & 1\% dropped messages & 10\% dropped messages \\
    \cmidrule(lr){1-2} \cmidrule(lr){3-5}
    \RelaySumModel w/ momentum & trees & 89.2\% & 89.3\% & 89.3\% \\
     \dpsgd~\citep{lian2017dpsgd} w/ quasi-global mom.~\citep{lin2021quasiglobal} & ring & 78.3\% & 76.2\% & 76.9\% \\
     \dsquare~\citep{tang2018d2} w/ momentum  & ring& 87.4\% & {\color{gray}diverges} & {\color{gray}diverges} \\
     SGP~\citep{assran2019sgp} w/ momentum & time-varying & 88.5\% & 88.6\% & 88.1\% \\
    \bottomrule
\end{tabularx}
\end{table}

\vspace{-1mm}

\subsection{Robustness to unreliable communication}

Peer-to-peer applications are a central use case for decentralized learning.
Decentralized learning algorithms must therefore be robust to workers joining and leaving, and to unreliable communication between workers.
Gossip averaging naturally features such robustness, but for methods like \dsquare, that correct for local data biases, achieving such robustness is non-trivial.
As a proxy for these challenges, in \autoref{tab:robustness}, we verify that \RelaySumModel can tolerate randomly dropped messages.
The algorithm achieves this by reliably counting the number of models summed up in each message.
For this experiment, we use an extended version of Algorithm~\ref{alg:relaysum_model}, where line 10 is replaced by
\begin{align}\textstyle
    \xx_i^{(t+1)} = \frac{1}{n}\Big( \xx_i^{(t+\half)} + \sum_{j \in \mathcal{N}_i} \mm_{j\to i}^{(t)} + (n - \bar{n}_i^{(t+1)} )\xx_i^{(t)}\Big). \label{eq:robust-relaysum-model}
\end{align}
We count the number of models received as $\bar{n}$, and substitute any missing models ($< n$) by the previous state $\xx_i^{(t)}$.
\RelaySumModel trains reliably to good test accuracy with up to 10\% deleted messages.
This behavior is on par with a similarly modified SGP~\citep{assran2019sgp} that corrects for missing energy.
In contrast, \dsquare becomes unstable with undelivered messages and diverges.

\vspace{-1mm}

\section{Conclusion}

\vspace{-1mm}

Decentralized learning has great promise as a building block in the democratization of deep learning.
Deep learning relies on large datasets, and while large companies can afford those, many individuals together can, too.
Of course, their data does not follow the exact same distribution, calling for robustness of decentralized learning algorithms to data heterogeneity.
Algorithms with this property have been proposed and analyzed theoretically, but they do not always perform well in deep learning.

In this paper, we propose \RelaySumModel for distributed optimization over decentralized networks with heterogeneous data.
Unlike algorithms based on gossip averaging, \RelaySumModel \emph{relays} models through spanning trees of a network without decaying their magnitude.
This yields an algorithm that is both theoretically independent of data heterogeneity, but also high performing in actual deep learning tasks.
With its demonstrated robustness to unreliable communication, \RelaySumModel makes an attractive choice for peer-to-peer deep learning and applications in large-scale data centers.

\begin{ack}
      This project was supported by SNSF grant 200020\_200342, as well as EU project DIGIPREDICT, and a Google PhD Fellowship.

      We thank Yatin Dandi and Lenka Zdeborová for pointing out the similarities between this algorithm and Belief Propagation during a poster session.
      This discussion helped us find the strongly related article by \citet{zhang2019bp} that we missed initially.

      We thank Renee Vogels for proofreading of the manuscript.
\end{ack}

{\small
\bibliography{bibliography_autogen}
}

\appendix

{\hypersetup{linkcolor=black}
\parskip=0em
\renewcommand{\contentsname}{Contents of the Appendix}
\tableofcontents
\addtocontents{toc}{\protect\setcounter{tocdepth}{3}}
}

\section{Convergence Analysis of \RelaySumModel}\label{apx:theory}
The structure of this section is as follows: \Cref{ssec:notations} describes the notations used in the proof; \Cref{ssec:technical} introduces the properties of mixing matrix $\mW$ and useful inequalities and lemmas; \Cref{ssec:results} elaborates the results of \Cref{thm:1} for non-convex, convex, and strongly convex objectives, all of the technical details are deferred to \Cref{ssec:convex}, \Cref{ssec:strongly-convex} and \Cref{ssec:non-convex}.

\subsection{Notation}\label{ssec:notations}
We use upper case, bold letters for matrices and lower case, bold letters for vectors. By default,let $\|\cdot\|$ and $\| \cdot \|_F$ be the spectral norm and Frobenius norm for matrices and 2-norm $\|\cdot\|_2$ be the Euclidean norm for vectors.

Let $\tau_{ij}$ be the delay between node $i$ and node $j$ and let $\tau_{\max}=\max_{ij}\tau_{ij}$. Let $$\mZ^{(t)}=[\xx_1^{(t)}, \ldots, \xx_n^{(t)}]^\top\in\R^{n\times d}$$
be the state at time $t$ and let
$$\nabla \mF^{(t)}=[\nabla F_1(\xx_1^{(t)}; \xi_1^{(t)}), \ldots, \nabla F_n(\xx_n^{(t)}; \xi_n^{(t)})]^\top\in\R^{n\times d}$$
be the worker gradients at time $t$. Denote $\mY^{(t)}$ and $\mG^{(t)}$ as the state (models) and gradients respectively, of all nodes, from time $t-\tau_{\max}$ to $t$.
\begin{align*}
    \mY^{(t)} = \begin{bmatrix}
        \mZ^{(t)} \\
        \mZ^{t-1} \\
        \vdots    \\
        \mZ^{t-\tau_{\max}}
    \end{bmatrix}\in\R^{n(\tau_{\max}+1)\times d},
    \qquad
    \mG^{(t)} = \begin{bmatrix}
        \nabla \mF^{(t)} \\
        \nabla \mF^{t-1} \\
        \vdots           \\
        \nabla \mF^{t-\tau_{\max}}
    \end{bmatrix}\in\R^{n(\tau_{\max}+1)\times d}.
\end{align*}
The mixing matrix $\mW$ can be alternatively defined as follows
\begin{definition}[Mixing matrix $\mW$]\label{def:W}
    Define $\mW, \tilde{\mW}\in\R^{n(\tau_{\max}+1)\times n(\tau_{\max}+1)}$ such that \RelaySumModel can be reformulated as
    \begin{align*}%
        \mY^{(t+1)}
        =\underbrace{\begin{bmatrix}
                \mW_0  & \mW_1  & \ldots & \mW_{\tau_{\max}-1} & \mW_{\tau_{\max}} \\
                \mI    & \0     & \ldots & \0                  & \0                \\
                \vdots &        & \ddots & \ddots              & \vdots            \\
                \0     & \ldots & \ldots & \mI                 & 0
            \end{bmatrix}}_{\mW}
        \mY^{(t)}
        -\gamma
        \underbrace{\begin{bmatrix}
                \mW_0  & \mW_1  & \ldots & \mW_{\tau_{\max}-1} & \mW_{\tau_{\max}} \\
                \0     & \0     & \ldots & \0                  & \0                \\
                \vdots &        & \ddots & \ddots              & \vdots            \\
                \0     & \ldots & \ldots & \0                  & \0
            \end{bmatrix}}_{\tilde{\mW}}
        \mG^{(t)}
    \end{align*}
    where $\sum_{i=1}^n \mW_i = \frac{1}{n}\1_n\1_n^\top$.
\end{definition}

\subsection{Technical Preliminaries}\label{ssec:technical}
\subsubsection{Properties of $\mW$.}
In this part, we show that $\mW$ enjoys similar properties as Perron-Frobenius Theorem in \Cref{theorem:Perron:W} and its left dominant eigenvector $\mpi$ has specific structure in \Cref{lemma:W:basic}. Then we use the established tools to prove the key Lemma~\ref{lemma:algo1:key}. Finally, we define constants $C$ and $C_1$ in \Cref{definition:C} which are used to simplify the convergence results in \Cref{ssec:results}.

\begin{definition}[Spectral radius.]
    Let $\lambda_1,\ldots,\lambda_n$ be the eigenvalues of a matrix $\mA\in\C^{n\times n}$. Then its spectral radius $\rho(\mA)$ is defined as:
    \begin{align*}
        \rho(\mA) = \max\{|\lambda_1|,\ldots,|\lambda_n|\}.
    \end{align*}
\end{definition}

\begin{lemma}\label{lemma:W:basic}
    The $\mW$ in \Cref{def:W} satisfies
    \begin{enumerate}[nosep]
        \item The spectral radius $\rho(\mW)=1$ and 1 is an eigenvalue of $\mW$ and $\1_{n(\tau_{\max}+1)}\in\R^{n(\tau_{\max}+1)}$ is its right eigenvector.
        \item The left eigenvector $\mpi\in\R^{n(\tau_{\max}+1)}$ of eigenvalue 1 is nonnegative and $[\mpi]_i = \pi_0>0, \forall~i\in[n]$
              and $\mpi^\top\1_{n(\tau_{\max}+1)}=1$.
    \end{enumerate}
\end{lemma}
\begin{proof}
    Since $\mW$ is a row stochastic matrix, the Gershgorin Circle Theorem asserts the spectral radius
    \[
        \rho(\mW)=|\lambda_1(\mW)|\le1.
    \]
    It is clear that 1 is an eigenvalue of $\mW$ and $\1_{n(\tau_{\max}+1)}$ is its right eigenvector, we have $\rho(\mW)=1$.

    Let $\mpi\in\R^{n(\tau_{\max}+1)}$ be the left eigenvector corresponding to 1 and denote it as
    \begin{align*}
        \bm{\pi}=
        \begin{bmatrix}
            \bm{\pi}_0 \\
            \bm{\pi}_1 \\
            \vdots     \\
            \bm{\pi}_{\tau_{\max}}
        \end{bmatrix}\in\R^{n(\tau_{\max}+1)}
    \end{align*}
    where $\bm{\pi}_i \in\R^n,\forall~i=0,1,\ldots,\tau_{\max}$. Since $\bm{\pi} = \mW^{\top}\bm{\pi}$, we have
    \begin{align*}
        \begin{bmatrix}
            \bm{\pi}_0 \\
            \bm{\pi}1  \\
            \vdots     \\
            \bm{\pi}_{\tau_{\max}}
        \end{bmatrix}
        =\bm{\pi}=
        \mW^{\top}\bm{\pi}=
        \begin{bmatrix}
            \mW_0^\top \bm{\pi_0} + \bm{\pi_1}                           \\
            \mW_1^\top \bm{\pi_0} + \bm{\pi_2}                           \\
            \vdots                                                       \\
            \mW_{\tau_{\max}-1}^\top \bm{\pi_0} + \bm{\pi_{\tau_{\max}}} \\
            \mW_{\tau_{\max}}^\top \bm{\pi_0}
        \end{bmatrix}
    \end{align*}
    which holds true in each block. Then summing up all blocks yields
    \begin{align*}
        \sum_{i=0}^{\tau_{\max}}\bm{\pi_i} = \left(\sum_{i=0}^{\tau_{\max}} \mW_i^\top\right) \bm{\pi_0} + \sum_{i=1}^{\tau_{\max}} \bm{\pi_i}
        =\frac{1}{n}\1_n\1_n^\top \bm{\pi}_0 + \sum_{i=1}^{\tau_{\max}} \bm{\pi_i}
    \end{align*}
    which means $\bm{\pi}_0 = \frac{1}{n}\1_n\1_n^\top \bm{\pi}_0$ and therefore $\bm{\pi}_0=\pi_0 \1_{n}$ is a vector of same value. %

    Other coordinate blocks of $\mpi$ can be derived as
    \begin{align*}
        \mpi_{i} = \left(\sum_{k=i}^{\tau_{\max}} \mW_k^\top \right)\mpi_0
        \qquad \forall~i=1,\ldots,\tau_{\max}.
    \end{align*}
    Since $\mW_i$ are nonnegative matrices, we can scale $\mpi$ such that $\pi_0>0$ and $\1^\top\mpi=1$. Therefore $\mpi$ is a nonnegative vector.
\end{proof}

\begin{lemma}\label{lemma:W:geometric_multiplicity}
    If $\lambda\in\C$ is an eigenvalue of $\mW$ and $|\lambda|=\rho(\mW)=1$, then $\lambda=1$ and its geometric multiplicity is 1.
\end{lemma}
\begin{proof}
    Let $\bm{v}\in\C^{n(\tau_{\max}+1)}$ be a right eigenvector corresponding to eigenvalue $\lambda\in\C$ which $|\lambda|=1$.

    Denote $\bm{v}$ as
    \begin{align*}
        \bm{v}=
        \begin{bmatrix}
            \bm{v}_0 \\
            \bm{v}_1 \\
            \vdots   \\
            \bm{v}_{\tau_{\max}}
        \end{bmatrix}\in\C^{n(\tau_{\max}+1)}.
    \end{align*}
    where $\bm{v}_i\in\C^n, \forall~i=0,\ldots,\tau_{\max}$.
    Then $\mW\bm{v}=\lambda\bm{v}$ implies
    \begin{align*}
        \mW\bm{v}=
        \begin{bmatrix}
            \sum_{i=0}^{\tau_{\max}}\mW_i \bm{v_i} \\
            \bm{v}_0                               \\
            \vdots                                 \\
            \bm{v}_{\tau_{\max}-2}                 \\
            \bm{v}_{\tau_{\max}-1}
        \end{bmatrix}
        =
        \lambda\bm{v}=
        \begin{bmatrix}
            \lambda\bm{v}_0 \\
            \lambda\bm{v}1  \\
            \vdots          \\
            \lambda\bm{v}_{\tau_{\max}}
        \end{bmatrix}.
    \end{align*}
    The last $\tau$ equations ensures $\bm{v}_i = \lambda^{-i}\bm{v}_0$ and thus the first equality becomes
    \begin{align*}
        \left(\sum_{i=0}^{\tau_{\max}}\mW_i \lambda^{-i} \right)\bm{v}_0=\lambda\bm{v}_0
    \end{align*}
    Denote $\bm{v}_0=[x_1, x_2, \ldots, x_n]^\top\in\C^{n}$, then $\forall~i=1,\ldots,n$
    \begin{align}\label{eq:donimance:1}
        \tsum_{j=1}^{n} \tfrac{1}{n} \lambda^{-\tau_{ij}} x_j=\lambda x_i.
    \end{align}
    Pick $i$ such that $|\lambda x_i| = \max_{j} |\lambda x_j|$, then
    \begin{align*}
        |\lambda x_i| = |\tsum_{j=1}^{n} \tfrac{1}{n} \lambda^{-\tau_{ij}} x_j|
        \le              \tfrac{1}{n} \tsum_{j=1}^{n} | \lambda^{-\tau_{ij}} x_j|
        =                \tfrac{1}{n} \tsum_{j=1}^{n} |\lambda^{-\tau_{ij}}| |x_j|
        =               \tfrac{1}{n} \tsum_{j=1}^{n} |x_j|
        \le             |x_i|
    \end{align*}
    where we use the triangular inequality $|a+b|\le|a|+|b|$ and $|ab|=|a||b|$ for all $a,b\in\C$.

    Note that as $|\lambda x_i|=|\lambda| |x_i|=|x_i|$, the triangular inequality is in fact an equality which means $\lambda^{-\tau_{ij}} x_j$ could be written as
    \begin{align*}
        \lambda^{-\tau_{ij}} x_j = a_{ij} \xi\qquad \forall~j\in[n].
    \end{align*}
    where $a_{ij}\ge 0$ and $\xi\in\C$. Here $\xi\neq0$, otherwise $\bm{v}=\0$ which contradicts to $\bm{v}$ is an eigenvector. Then \eqref{eq:donimance:1} becomes
    \begin{align*}
        \tfrac{1}{n} \tsum_{j=1}^{n} a_{ij} \xi=\lambda a_{ii} \xi.
    \end{align*}
    which implies $|\tfrac{1}{n} \tsum_{j=1}^{n} a_{ij}|=|a_{ii}|$. As $|\lambda x_i| = \max_{j} |\lambda x_j|$, we know $a_{ii}\ge a_{ij}$ for all $j$, thus
    \begin{align*}
        a_{i1}=\ldots=a_{in}=a \ge0,
    \end{align*}
    moreover, $a>0$ as $a=0$ again leads to $\bm{v}=\0$. Then \eqref{eq:donimance:1} becomes
    \begin{align*}
        \lambda a\xi=\lambda x_i =\tfrac{1}{n} \tsum_{j=1}^{n} \lambda^{-\tau_{ij}} x_j =
        \tfrac{1}{n} \tsum_{j=1}^{n} a \xi = a \xi
    \end{align*}
    which shows $\lambda=1$ as $a>0$ and $\xi\neq0$.

    Therefore, $\bm{v}_0=a\1_n\in{\R^n}$ and $\bm{v}=a\1_{n(\tau_{\max}+1)}\in{\R^{n(\tau_{\max}+1)}}$. It mean the eigenspace of 1 is one-dimensional and thus its geometric multiplicity is 1.
\end{proof}

\begin{lemma}\label{lemma:W:algebraic_multiplicity}
    The algebraic multiplicity of eigenvalue 1 of $\mW$ is 1.
\end{lemma}
\begin{proof}
    Proof by contradiction. Let $\mP\in\R^{n(\tau_{\max}+1)\times n(\tau_{\max}+1)}$ be the invertible matrix which transform $\mW$ to its Jordan normal form $\mJ$ by
    \begin{align*}
        \mP^{-1} \mW \mP = \mJ=\begin{bmatrix}
            \mJ_1 &        &       \\
                  & \ddots &       \\
                  &        & \mJ_p
        \end{bmatrix}
    \end{align*}
    where $\mJ_1$ is the block for eigenvalue 1. If we assume the algebraic multiplicity of 1 greater equal than 2, and use the \Cref{lemma:W:geometric_multiplicity} that its geometric multiplicity is 1, then $\mJ_1$ should look like
    \begin{align*}
        \mJ_1=\begin{bmatrix}
            1 & 1 &        &   \\
              & 1 & \ddots &   \\
              &   & \ddots & 1 \\
              &   &        & 1
        \end{bmatrix}
    \end{align*}
    which is a square matrix of at least 2 columns. Denote the first two columns of $\mP$ as $\pp_1$ and $\pp_2$. We can see that $\pp_1=\1_{n(\tau_{\max}+1)}$. Then inspecting $\mP^{-1} \mW \mP = \mJ$ for $\pp_2$ yields
    \begin{align*}
        \mW\pp_2 = \pp_1 + \pp_2 = \1_{n(\tau_{\max}+1)} + \pp_2.
    \end{align*}
    Multiply both sides by $\mpi^\top$ gives
    \begin{align*}
        \mpi^\top\mW\pp_2 = & \mpi^\top \1_{n(\tau_{\max}+1)} +\mpi^\top \pp_2 \\
        \mpi^\top\pp_2 =    & \mpi^\top \1_{n(\tau_{\max}+1)} +\mpi^\top \pp_2 \\
        0 =                 & \mpi^\top \1_{n(\tau_{\max}+1)}
    \end{align*}
    which contradicts \Cref{lemma:W:basic} that $\mpi^\top \1_{n(\tau_{\max}+1)}=1$. Thus the algebraic multiplicity of 1 is 1.
\end{proof}

\begin{theorem}[Perron-Frobenius Theorem for $\mW$]\label{theorem:Perron:W}%
    The mixing $\mW$ of \RelaySumModel satisfies
    \begin{enumerate}[nosep]
        \item (Positivity) $\rho(\mW) = 1$ is an eigenvalue of $\mW$.
        \item (Simplicity) The algebraic multiplicity of 1 is 1.
        \item (Dominance) $\rho(\mW)=|\lambda_1(\mW)|>|\lambda_2(\mW)|\ge\ldots\ge|\lambda_{n(\tau_{\max}+1)}(\mW)|$.
        \item (Nonnegativity) The $\mW$ has a nonnegative left eigenvector $\mpi$ and right eigenvector $\1_{n(\tau_{\max}+1)}$.
    \end{enumerate}
\end{theorem}
\begin{proof}
    Statements 1 and 4 follow from \Cref{lemma:W:basic}. Statement 2 follows from \Cref{lemma:W:algebraic_multiplicity}. Statement 3 follows from \Cref{lemma:W:geometric_multiplicity} and \Cref{lemma:W:algebraic_multiplicity}.
\end{proof}

\begin{lemma}[Gelfand's formula]\label{lemma:gelfand}
    For any matrix norm $\|\cdot\|$, we have
    \begin{align*}
        \rho(\mA) = \lim_{k\rightarrow\infty} \| \mA^k\|^{\frac{1}{k}}.
    \end{align*}
\end{lemma}

\begin{figure}
    \centering
    \includegraphics{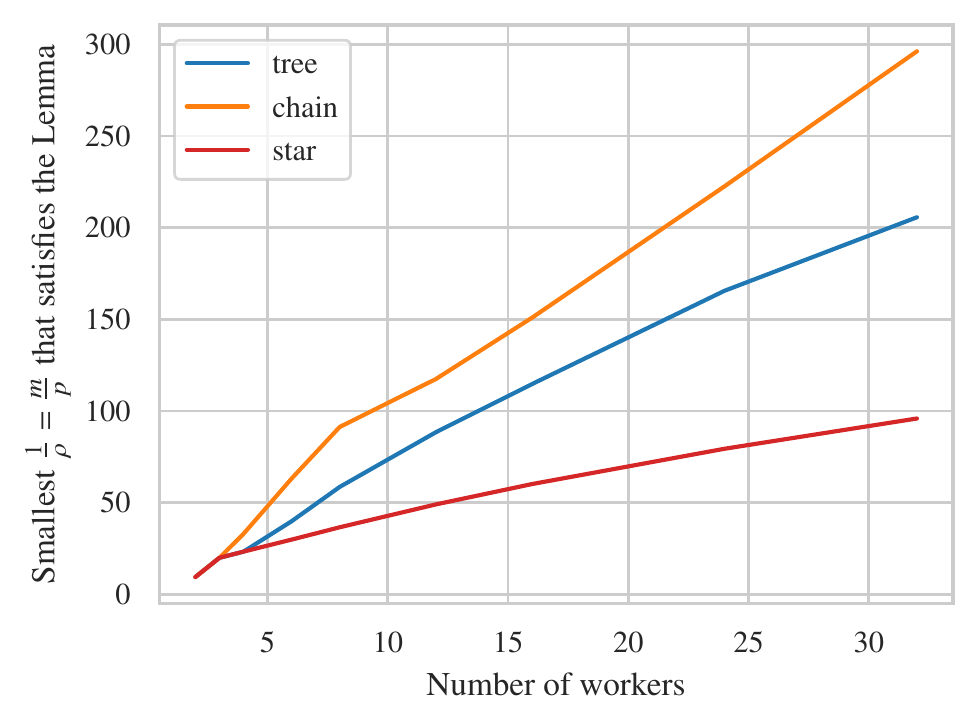}
    \caption{
        Optimal ratios for $\rho = p/m$ for Lemma~\autoref{lemma:algo1:key} computed empirically for three common types of graph topologies.        \label{fig:empirical-rho}
    }
\end{figure}

We characterize the convergence rate of the consensus distance in the following key lemma:
\begin{replemma}{lemma:algo1:key}[Key lemma]
    Given $\mW$ and $\mpi$ as before.
    There exists an integer $m=m(\mW)>0$ such that for any $\mX\in\R^{n(\tau_{\max}+1)\times d}$ we have\vspace{-1mm}
    \begin{align*}
        \| \mW^m \mX - \1\mpi^\top \mX \|^2 \le (1-p)^{2m} \| \mX - \1\mpi^\top \mX \|^2,
    \end{align*}
    where $p=\frac12(1-|\lambda_2(\mW)|)$ is a constant.\vspace{-1mm}
\end{replemma}
All the following optimization convergence results will only depend on the \emph{effective spectral gap} $\rho := \frac{p}{m}$ of $\mW$.
We empirically observe that $\rho =\Theta(1/n)$ for a variety of network topologies, as shown in Figure \ref{fig:empirical-rho}.

\begin{proof}[Proof of key lemma \ref{lemma:algo1:key}]
    First, let $\{\lambda_i\}$ and $\{\bm{v}_i\}$ be the eigenvalues and right eigenvectors of $\mW$ where $\lambda_1=1$ and $\bm{v}_1=\1_{n(\tau_{\max}+1)}$, then
    \begin{align*}
        (\mW-\1\mpi^\top) \bm{v}_1 = & (\mW-\1\mpi^\top) \1 = \0                                                            \\
        (\mW-\1\mpi^\top) \bm{v}_i = & \mW\bm{v}_i-\1\mpi^\top\bm{v}_i = \mW\bm{v}_i = \lambda_i\bm{v}_i \qquad \forall~i>1
    \end{align*}
    where $\mpi^\top\bm{v}_i=0$ because
    \[
        (1 - \lambda_i) \mpi^\top\bm{v}_i =\mpi^\top\bm{v}_i - \lambda_i \mpi^\top\bm{v}_i
        = (\mpi^\top\mW)\bm{v}_i - \mpi^\top(\mW\bm{v}_i)=0.
    \]
    The spectrum of $\mW-\1\mpi^\top$ are
    \[
        \{0, \lambda_2, \ldots, \lambda_{n(\tau_{\max}+1)}\},
    \]
    and thus the spectral radius of $\mW-\1\mpi^\top$ is $|\lambda_2|<1$. Since
    \[
        \mW^m-\1\mpi^\top = (\mW-\1\mpi^\top)^m,
    \]
    then $\mW^m-\1\mpi^\top$ has a spectral radius of $|\lambda_2|^m<1$.

    Then, we apply Gelfand's formula (\Cref{lemma:gelfand}) with $\mA= \mW-\1\mpi^\top$ and can conclude that for a given $\epsilon \in (0, 1-|\lambda_2|)$, there exists a large enough integer $m>0$ such that
    \begin{align*}
        \|\mW^m - \1\mpi^\top \|=\| (\mW-\1\mpi^\top)^m \| \leq (\rho(\mW-\1\mpi^\top)+\epsilon)^m =(|\lambda_2|+\epsilon)^m <1.
    \end{align*}
    Thus
    \begin{align*}
        \| \mW^m \mX - \1\mpi^\top \mX \|^2
        \le \|\mW^m - \1\mpi^\top \|^2 \| \mX - \1\mpi^\top \mX \|^2
        \le (1-p)^{2m} \| \mX - \1\mpi^\top \mX \|^2
    \end{align*}
    where $p\in(0, 1-|\lambda_2|)$.
\end{proof}

\begin{definition}\label{definition:C}
    Given $\mW$ and $m$, and $\tilde{\mI}\in\R^{n(\tau_{\max}+1)\times n(\tau_{\max}+1)}$ is a matrix which satisfies
    \begin{align*}
        [\tilde{\mI}]_{ij}=\begin{cases}
            1 & i=j \le n         \\
            0 & \text{Otherwise}.
        \end{cases}
    \end{align*}
    We define constants $C_1^2:=\max_{i=0,\ldots,m-1} \|\mW^i \tilde{\mI} \|^2$ and $C=C(\mW)$ such that
    \begin{align*}
        C^2:= \frac{C_1^2}{\|\mW^{\infty} \tilde{\mI} \|^2}.
    \end{align*}
    where $\mW^{\infty}:=\1\mpi^\top$.
\end{definition}
In addition, the $\|\1\mpi^\top \tilde{\mI} \|^2$ can be computed as follows.
\begin{lemma}\label{lemma:C}
    Given $\tilde{\mI}$ in \Cref{definition:C}, we have the following estimate
    \begin{align*}
        \|\1\mpi^\top \tilde{\mI} \|^2=n^2(\tau_{\max}+1) \pi_0^2 \le n^3 \pi_0^2.
    \end{align*}
\end{lemma}
\begin{proof}
    For rank $r$ matrix $\|A\|^2\le\|A\|_F^2\le r \|A\|^2$. Since $\1\mpi^\top \tilde{\mI}$ is a rank 1 matrix, we know that
    \begin{align*}
        \|\1\mpi^\top \tilde{\mI} \|^2 =\|\1\mpi^\top \tilde{\mI} \|_F^2.
    \end{align*}
    As the first n entries of $\mpi$ are $\pi_0$, we can compute that
    \begin{align*}
        \|\1\mpi^\top \tilde{\mI} \|_F^2 = n^2(\tau_{\max}+1) \pi_0^2.
    \end{align*}
\end{proof}

\subsubsection{Useful inequalities and lemmas}
For convex objective, the noise in \Cref{assumption:uniform_sigma_zeta} can be defined only at the minimizer $\xx^\star$ which leads to \Cref{assumption:zeta2:convex}. This assumption is used in the proof of \Cref{prop:algo1:convex:bounded_noise}.
\begin{assumption}[Bounded noise at the optimum] \label{assumption:zeta2:convex}
    Let $\xx^\star=\argmin f(\xx)$ and define
    \begin{equation}\label{eq:bounded_noise:convex}
        \zeta_i^2 := \| \nabla f_i(\xx^\star) \|^2,\qquad \bar{\zeta}^2 := \tfrac{1}{n}\tsum_{i=1}^n \zeta_i^2.
    \end{equation}
    Further, define
    \begin{align*}
        \sigma_i^2 :=\E_{\xi_i}\| \nabla F_i(\xx^\star, \xi_i)-\nabla f_i(\xx^\star) \|^2
    \end{align*}
    and similarly as above, $\bar{\sigma}^2:=\frac{1}{n}\sum_{i=1}^n\sigma_i^2$. We assume that $\bar{\sigma}^2$ and $\bar{\zeta}^2$ are bounded.
\end{assumption}

\begin{lemma}[Cauchy-Schwartz inequality]
    For arbitrary set of $n$ vectors $\{\aa_i\}_{i=1}^n$, $a_i\in\R^d$
    \begin{equation}\label{eq:cs}
        \left\|\sum_{i=1}^n \aa_i\right\|^2 \le n \sum_{i=1}^n\| \aa_i\|^2.
    \end{equation}
\end{lemma}

\begin{lemma}
    If function $g(\xx)$ is $L$-smooth, then
    \begin{equation}\label{eq:smooth:optimum} %
        \|\nabla g(\xx) - \nabla g(\yy) \|^2 \le 2 L (g(\xx) - g(\yy) - \langle \xx - \yy, \nabla g(\yy) \rangle), \qquad \forall~\xx,\yy\in\R^d.
    \end{equation}
\end{lemma}

\begin{lemma}
    Let $\mA$ be a matrix with $\{\aa_i\}_{i=1}^n$ as its columns and $\bar{\aa}=\frac{1}{n}\sum_{i=1}^n \aa_i$, $\bar{\mA}=\bar{\aa}\1^\top$ then
    \begin{equation}\label{eq:frobenius-avg}
        \|\mA-\bar{\mA}\|_F^2 = \sum_{i=1}^n \| \aa_i - \bar{\aa} \|^2 \le \sum_{i=1}^n \| \aa_i \|^2 = \| \mA \|^2_F.
    \end{equation}
\end{lemma}

\begin{lemma}
    Let $\mA$,$\mB$ be two matrices
    \begin{equation}\label{eq:frobenius-spectral}
        \| \mA\mB \|_F^2 \le \| \mA \|^2_F \| \mB \|^2.
    \end{equation}
\end{lemma}

\subsection{Results of Theorem~\ref{thm:1}}\label{ssec:results}
In this subsection, we summarize the precise results of Theorem~\ref{thm:1} for convex, strongly convex and non-convex cases. The complete proofs for each case are then given in the following \Cref{ssec:convex}, \Cref{ssec:strongly-convex} and \Cref{ssec:non-convex}.
\begin{reptheorem}{thm:1}
    Given mixing matrix $\mW$ and $\tilde{\mW}$, constant $m$, $p$ defined in \Cref{lemma:algo1:key}, $C$, $C_1$ defined in \Cref{definition:C}. Under Assumption~\ref{assumption:smoothness:stochastic} and~\ref{assumption:uniform_sigma_zeta}, then for any target accuracy $\epsilon>0$,

    \textbf{Non-convex:} if the objective is non-convex, then $\frac{1}{T+1}\sum_{t=0}^T \| \nabla f(\bar{\xx}^{(t)}) \|^2\le \epsilon$ after
    \begin{align*}
        \cO\left(
        \frac{\bar{\sigma}^2}{n\epsilon^2}
        + \frac{Cm\bar{\sigma}}{\sqrt{p}\epsilon^{3/2}}
        + \frac{C_1m}{p\epsilon}
        \right) Lr_0
    \end{align*}
    iterations, where $r_0=f(\xx^{(0)}) - f^\star$.

    \textbf{Convex:} if the objective is convex and $\xx^\star$ is the minimizer, then $\tfrac{1}{T+1}\tsum_{t=0}^T \left(f(\bar{\xx}^{(t)}) - f(\xx^\star)\right)\le \epsilon$ after
    \begin{align*}
        \cO\left(
        \frac{\bar{\sigma}^2}{n\epsilon^2}
        + \frac{Cm\sqrt{L}\bar{\sigma}}{\sqrt{p}\epsilon^{3/2}}
        + \frac{Lm\sqrt{n}C}{p\epsilon}
        \right) r_0
    \end{align*}
    iterations, where $r_0=\| \xx^0 - \xx^\star \|^2$.

    \textbf{Strongly-convex:} if the objective is $\mu$ strongly convex and $\xx^\star$ is the minimizer, then $\frac{1}{W_T} \sum_{t=0}^T w_t (\E f(\bar{\xx}^{(t)})-f^\star) + \mu \E\|\bar{\xx}^{(T+1)}-\xx^\star\|^2 \le \epsilon$ after
    \begin{align*}
        \tilde{\cO}\left(
        \frac{\bar{\sigma}^2}{\mu n\epsilon^2}
        + \frac{Lm^2C^2 \bar{\sigma}^2}{\mu n p^2\epsilon}
        + \frac{s}{a} \log{\frac{bsr_0}{\epsilon}}
        \right)
    \end{align*}
    iterations, where $r_0=\| \xx^0 - \xx^\star \|^2$, $w_t=(1-\frac{\mu\gamma n\pi_0}{2})^{-(t+1)}$ and $W_T=\sum_{t=0}^Tw_t$ and $a=\frac{\mu n\pi_0}{2}$, $b=\frac{2}{n\pi_0}$,  $s= \frac{aT}{\ln\max\{\frac{ba^2T^2r_0}{\pi_0\bar{\sigma}^2}, 2\}}$.
\end{reptheorem}
In all three cases, the convergence rate is independent of the heterogeneity $\zeta^2$.

\subsection{Proof of Theorem~\ref{thm:1} in the convex case}
\label{ssec:convex}

Let $\bar{\xx}^{(t)}:=\left(\mpi^\top \mY ^{(t)}\right)^\top$ and $\bar{\mY}^{(t)}:= \1\mpi^\top \mY ^{(t)}$.
Let $\xx^\star$ be the minimizer of $f$ and define the following iterates
\begin{itemize}[nosep]
    \item $r_t:=\|\bar{\xx}^{(t)} - \xx^\star \|^2$,
    \item $e_t:= f(\bar{\xx}^{(t)}) - f(\xx^\star)$,
    \item $\Xi_t:=\frac{1}{n} \|\bar{\mY}^{(t)}- \mY^{(t)}\|^2_F$.
\end{itemize}

The consensus distance $\Xi_t$ can be written as follows
\begin{equation}\label{eq:Xi:vector_form}
    \Xi_t=\frac{1}{n} \sum_{i=1}^n \sum_{\tau=0}^{\tau_{\max}} \| \bar{\xx}^{(t)} - \xx_i^{(t-\tau)} \|^2.
\end{equation}

There is a related term $\sum_{i=1}^n\sum_{j=1}^{n} \|\bar{\xx}^{(t)} - \xx_i^{(t-\tau_{ij})}\|^2$ which will be used frequently in the proof. The next lemma explains their relations.
\begin{lemma}\label{lemma:Xi:alternative}
    For all $t\ge 0$
    \begin{equation*}
        \sum_{i=1}^n\sum_{j=1}^{n} \|\bar{\xx}^{(t)} - \xx_i^{(t-\tau_{ij})}\|^2
        \le n^2 \Xi_t.
    \end{equation*}
    where $\xx^{(0)}=\xx^{(-1)}=\ldots=\xx^{(-\tau_{\max})}$.
\end{lemma}
\begin{proof}
    Rewrite the $\tau_{ij}$ as an indicator function
    \begin{align*}
        \sum_{i=1}^n\sum_{j=1}^{n} \|\bar{\xx}^{(t)} - \xx_i^{(t-\tau_{ij})}\|^2
        = & \sum_{i=1}^n\sum_{j=1}^{n} \sum_{\tau=0}^{\tau_{\max}} \1_{\{\tau=\tau_{ij}\}} \|\bar{\xx}^{(t)} - \xx_i^{(t-\tau)}\|^2.
    \end{align*}
    This term can be relaxed by removing the indicator function
    \begin{align*}
        \sum_{i=1}^n\sum_{j=1}^{n} \|\bar{\xx}^{(t)} - \xx_i^{(t-\tau_{ij})}\|^2
        \le & n \sum_{i=1}^n\sum_{\tau=0}^{\tau_{\max}}  \|\bar{\xx}^{(t)} - \xx_i^{(t-\tau)}\|^2.
    \end{align*}
    Then applying \eqref{eq:Xi:vector_form} for the consensus distance in vector form completes the proof.
\end{proof}

The next two propositions upper bound the difference between stochastic gradients and full gradients.
\begin{proposition}\label{prop:algo1:convex:bounded_noise}
    Under \Cref{assumption:smoothness:stochastic} and \ref{assumption:uniform_sigma_zeta}.
    Then for $t\ge0$,
    \begin{align*}
        \E \left\|
        \mpi^\top \tilde{\mW} (\E\mG^{(t)} - \mG^{(t)})
        \right\|^2
        \le 3n\pi_0^2(L^2 \Xi_t + 2L e_t + \bar{\sigma}^2).
    \end{align*}
\end{proposition}
\begin{proof}
    Use $T_0$ to denote the left hand side quantity
    \begin{align*}
        T_0 & \hspace{4mm}:=                                        &                                                         & \E \left\|
        \mpi^\top \tilde{\mW} (\E\mG^{(t)} - \mG^{(t)})
        \right\|^2                                                                                                                         \\
            & \hspace{5mm}=                                         &                                                         &
        \E \left\|\frac{\pi_0}{n}
        \sum_{i=1}^n \sum_{j=1}^n (\nabla f_j(\xx_j^{(t-\tau_{ij})}) - \nabla F_j(\xx_j^{(t-\tau_{ij})}; \xi_j^{(t-\tau_{ij})}))
        \right\|^2                                                                                                                         \\
            & \stackrel{\text{Cauchy-Schwartz } \eqref{eq:cs}}{\le}
            &                                                       & \frac{\pi_0^2}{n} \sum_{i=1}^n \E \left\| \sum_{j=1}^n(
        \nabla f_j(\xx_j^{(t-\tau_{ij})}) - \nabla F_j(\xx_j^{(t-\tau_{ij})}; \xi_j^{(t-\tau_{ij})}))
        \right\|^2.
    \end{align*}
    Since the randomness inside the norm are independent, we have
    \begin{align*}
        T_0 \le &
        \frac{\pi_0^2}{n} \sum_{i=1}^n \sum_{j=1}^n \E \left\|
        \nabla f_j(\xx_j^{(t-\tau_{ij})}) - \nabla F_j(\xx_j^{(t-\tau_{ij})}; \xi_j^{(t-\tau_{ij})})
        \right\|^2.
    \end{align*}

    Inside the vector norm, we can add and subtract terms the same terms and apply Cauchy-Schwartz \eqref{eq:cs}
    \begin{align*}
        T_0        \le & \frac{3\pi_0^2}{n} \sum_{i=1}^n \sum_{j=1}^n \E \big\|
        \nabla F_j(\xx_j^{(t-\tau_{ij})}; \xi_j^{(t-\tau_{ij})}) - \nabla F_j(\bar{\xx}^{(t)}; \xi_j^{(t-\tau_{ij})}) + \nabla f_j(\xx_j^{(t-\tau_{ij})}) - \nabla f_j(\bar{\xx}^{(t)}) \big\|^2 \\
                       & +
        \frac{3\pi_0^2}{n} \sum_{i=1}^n \sum_{j=1}^n \E \left\|
        \nabla F_j(\bar{\xx}^{(t)}; \xi_j^{(t-\tau_{ij})})
        - \nabla F_j(\xx^\star; \xi_j^{(t-\tau_{ij})})
        + \nabla f_j(\bar{\xx}^{(t)})
        - \nabla f_j(\xx^\star)
        \right\|^2                                                                                                                                                                               \\
                       & +
        \frac{3\pi_0^2}{n} \sum_{i=1}^n \sum_{j=1}^n \E \left\|
        \nabla F_j(\xx^\star; \xi_j^{(t-\tau_{ij})})) - \nabla f_j(\xx^\star)
        \right\|^2.
    \end{align*}
    Use the inequality that for $a=\E Y$, $\E\norm{Y-a}^2=\E\norm{Y}^2-\norm{a}^2\le\E\norm{Y}^2$, then we have
    \begin{align*}
        T_0 \le & \frac{3\pi_0^2}{n} \sum_{i=1}^n \sum_{j=1}^n \E \big\|
        \nabla F_j(\xx_j^{(t-\tau_{ij})}; \xi_j^{(t-\tau_{ij})}) - \nabla F_j(\bar{\xx}^{(t)}; \xi_j^{(t-\tau_{ij})})  \big\|^2 \\
                & +
        \frac{3\pi_0^2}{n} \sum_{i=1}^n \sum_{j=1}^n \E \left\|
        \nabla F_j(\bar{\xx}^{(t)}; \xi_j^{(t-\tau_{ij})})
        - \nabla F_j(\xx^\star; \xi_j^{(t-\tau_{ij})})
        \right\|^2                                                                                                              \\
                & +
        \frac{3\pi_0^2}{n} \sum_{i=1}^n \sum_{j=1}^n \E \left\|
        \nabla F_j(\xx^\star; \xi_j^{(t-\tau_{ij})}) - \nabla f_j(\xx^\star)
        \right\|^2
    \end{align*}
    Applying \Cref{assumption:smoothness:stochastic}, Smoothness \eqref{eq:smooth:optimum}, and \Cref{assumption:uniform_sigma_zeta} (or \Cref{assumption:zeta2:convex})
    to the three terms gives
    \begin{align*}
        T_0 \le & \frac{3L^2\pi_0^2}{n} \sum_{i=1}^n \sum_{j=1}^n \big\|
        \xx_j^{(t-\tau_{ij})} - \bar{\xx}^{(t)} \big\|^2
        + 6Ln\pi_0^2  (f(\bar{\xx}^{(t)}) - f(\xx^\star))
        + 3\pi_0^2 n \bar{\sigma}^2                                      \\
        \stackrel{\text{Lemma }\ref{lemma:Xi:alternative}}{\le}
                & 3n\pi_0^2(L^2 \Xi_t + 2L e_t + \bar{\sigma}^2).
    \end{align*}
    where in the last line we have used our previous Lemma \ref{lemma:Xi:alternative}.
\end{proof}
The next proposition is very similar to the \Cref{prop:algo1:convex:bounded_noise} except that it considers the matrix form instead of the projection onto $\mpi$.
\begin{proposition}\label{prop:algo1:convex:bounded_noise:matrix_form}
    Under \Cref{assumption:smoothness:stochastic} and \ref{assumption:uniform_sigma_zeta}.
    Then for $t\ge0$,
    \begin{align*}
        \E \left\|
        \tilde{\mW} (\E\mG^{(t)} - \mG^{(t)})
        \right\|^2_F
        \le 3(L^2 \Xi_t + 2L e_t + \bar{\sigma}^2).
    \end{align*}
\end{proposition}
\begin{proof}
    \begin{align*}
            & \E \left\|
        \tilde{\mW} (\E\mG^{(t)} - \mG^{(t)})
        \right\|_F^2                                                                                                                                                    \\
        =   & \sum_{i=1}^n \E \left\| \frac{1}{n} \sum_{j=1}^n (\nabla F(\xx_j^{(t-\tau_{ij})}; \xi_j^{(t-\tau_{ij})}) - \nabla f_j(\xx_j^{(t-\tau_{ij})}) ) \right\|^2 \\
        \le & \frac{1}{n^2} \sum_{i=1}^n \sum_{j=1}^n \E \left\| \nabla F(\xx_j^{(t-\tau_{ij})}; \xi_j^{(t-\tau_{ij})}) - \nabla f_j(\xx_j^{(t-\tau_{ij})}) \right\|^2
    \end{align*}
    The rest of the proof is identical to the one of \Cref{prop:algo1:convex:bounded_noise}.
\end{proof}

\begin{lemma}\label{lemma:alg1:convex:descent}(Descent lemma for convex objective.)
    If $\gamma \le \frac{1}{10L n\pi_0}$, then
    \begin{align*}
        r_{t+1} \le (1-\tfrac{\gamma\mu n \pi_0}{2}) r_t - \gamma n \pi_0e_t + 4\gamma L n \pi_0 \Xi_t
        + 3\gamma^2n\pi_0^2 \bar{\sigma}^2.
    \end{align*}
\end{lemma}
\begin{proof}
    Expand $r_{t+1}=\E\|\bar{\xx}^{(t+1)} - \xx^\star \|^2$ as follows
    \begin{align*}
        \E\|\bar{\xx}^{(t+1)} - \xx^\star \|^2
        = & \E\|\bar{\xx}^{(t)} - \gamma \mpi^\top \tilde{\mW} \mG^{(t)} - \xx^\star \|^2                                                            \\
        = & \E\|\bar{\xx}^{(t)} - \gamma \mpi^\top \tilde{\mW} \E\mG^{(t)} - \xx^\star +  \gamma \mpi^\top \tilde{\mW} (\E\mG^{(t)} - \mG^{(t)})\|^2
    \end{align*}
    Directly expand it into three terms
    \begin{align*}
        \E\|\bar{\xx}^{(t+1)} - \xx^\star \|^2
        = & \E\left(
        \|\bar{\xx}^{(t)} - \gamma \mpi^\top \tilde{\mW} \E\mG^{(t)} - \xx^\star\|^2
        + \gamma^2 \|\mpi^\top \tilde{\mW} (\E\mG^{(t)} - \mG^{(t)}))\|^2\right. \\
          & + \left.
        \left\langle \bar{\xx}^{(t)} - \gamma \mpi^\top \tilde{\mW} \E\mG^{(t)} - \xx^\star, \gamma \mpi^\top \tilde{\mW} (\E\mG^{(t)} - \mG^{(t)}))\right\rangle
        \right)
    \end{align*}
    where the 3rd term is 0 and the second term is bounded in \Cref{prop:algo1:convex:bounded_noise}. The first term is independent of the randomness
    \begin{align*}
          & \|\bar{\xx}^{(t)} - \gamma \mpi^\top \tilde{\mW} \E\mG^{(t)} - \xx^\star \|^2 \\
        = & \|\bar{\xx}^{(t)} - \xx^\star \|^2
        + \gamma^2 \underbrace{\| \mpi^\top \tilde{\mW} \E\mG^{(t)}\|^2}_{=:T_1}
        - 2 \gamma \underbrace{\langle\mpi^\top \tilde{\mW} \E\mG^{(t)}, \bar{\xx}^{(t)} - \xx^\star\rangle}_{=:T_2}.
    \end{align*}
    Since $\mpi^\top \tilde{\mW} \E\mG^{(t)}=\frac{\pi_0}{n} \sum_{i=1}^n \sum_{j=1}^{n} \nabla f_i(\xx_i^{(t-\tau_{ij})})$, first bound $T_1$
    \begin{align*}
        T_1 & =                                                           &  & \pi_0^2\left\| \frac{1}{n} \sum_{i=1}^n \sum_{j=1}^{n} \nabla f_i(\xx_i^{(t-\tau_{ij})}) \right\|^2                                                                                       \\
            & =                                                           &  & \pi_0^2 \left\| \frac{1}{n} \sum_{i=1}^n \sum_{j=1}^{n} (\nabla f_i(\xx_i^{(t-\tau_{ij})}) - \nabla f_i(\bar{\xx}^{(t)}) + \nabla f_i(\bar{\xx}^{(t)}) - \nabla f_i(\xx^\star))\right\|^2
        \\
            & \le                                                         &  & 2 \pi_0^2 \left(
        \left\| \frac{1}{n} \sum_{i=1}^n \sum_{j=1}^{n} (\nabla f_i(\xx_i^{(t-\tau_{ij})}) - \nabla f_i(\bar{\xx}^{(t)}))\right\|^2
        + \left\| \sum_{i=1}^n (\nabla f_i(\bar{\xx}^{(t)}) - \nabla f_i(\xx^\star))\right\|^2
        \right)                                                                                                                                                                                                                                                          \\
            & \le                                                         &  & 2\pi_0^2L^2 \sum_{i=1}^n \sum_{j=1}^{n}\left\|\xx_i^{(t-\tau_{ij})} - \bar{\xx}^{(t)}\right\|^2
        + 2n\pi_0^2\sum_{i=1}^n\left\| \nabla f_i(\bar{\xx}^{(t)})
        - \nabla f_i (\xx^\star) \right\|^2                                                                                                                                                                                                                              \\
            & \stackrel{\text{Smoothness }\eqref{eq:smooth:optimum}}{\le} &  & 2\pi_0^2L^2 \sum_{i=1}^n \sum_{j=1}^{n}\left\|\xx_i^{(t-\tau_{ij})} - \bar{\xx}^{(t)}\right\|^2
        + 4Ln^2\pi_0^2 (f(\bar{\xx}^{(t)}) - f(\xx^\star))                                                                                ,
    \end{align*}
    Using again Lemma \ref{lemma:Xi:alternative} we have
    \begin{align*}
        T_1 \le2L^2n^2\pi_0^2\Xi_t + 4Ln^2\pi^2_0e_t.
    \end{align*}
    Then bound $T_2$
    \begin{align*}
        T_2 & =                                                       &                                                          & \frac{\pi_0}{n} \sum_{i=1}^n \sum_{j=1}^{n}\langle \nabla f_i(\xx_i^{(t-\tau_{ij})}), \bar{\xx}^{(t)} - \xx^\star\rangle                                    \\
            & =                                                       &                                                          & \frac{\pi_0}{n} \sum_{i=1}^n \sum_{j=1}^{n} (\langle \nabla f_i(\xx_i^{(t-\tau_{ij})}), \bar{\xx}^{(t)} - \xx_i^{(t-\tau_{ij})}\rangle
        + \langle \nabla f_i(\xx_i^{(t-\tau_{ij})}), \xx_i^{(t-\tau_{ij})} - \xx^\star\rangle )                                                                                                                                                                                                \\
            & \ge                                                     &                                                          & \frac{\pi_0}{n} \sum_{i=1}^n \sum_{j=1}^{n} (f_i(\bar{\xx}^{(t)}) -f_i(\xx_i^{(t-\tau_{ij})}) - \tfrac{L}{2} \| \bar{\xx}^{(t)} - \xx_i^{(t-\tau_{ij})}\|^2 \\
            &                                                         &                                                          & + f_i(\xx_i^{(t-\tau_{ij})}) - f_i(\xx^\star) + \tfrac{\mu}{2} \| \xx_i^{(t-\tau_{ij})} - \xx^\star \|^2)                                                   \\
            & =                                                       &                                                          & n \pi_0 (f(\bar{\xx}^{(t)}) - f(\xx^\star))
        + \frac{\pi_0}{n}\sum_{i=1}^n \sum_{j=1}^{n} (\tfrac{\mu}{2} \| \xx_i^{(t-\tau_{ij})} - \xx^\star \|^2 -\tfrac{L}{2} \| \bar{\xx}^{(t)} - \xx_i^{(t-\tau_{ij})} \|^2)                                                                                                                  \\
            & \ge                                                     &                                                          & n \pi_0 (f(\bar{\xx}^{(t)}) - f(\xx^\star))
        + \frac{\pi_0}{n}\sum_{i=1}^n \sum_{j=1}^{n} (\tfrac{\mu}{4} \| \bar{\xx}^{(t)} - \xx^\star \|^2 -\tfrac{\mu+L}{2} \| \bar{\xx}^{(t)} - \xx_i^{(t-\tau_{ij})} \|^2)                                                                                                                    \\
            & \stackrel{\text{Lemma }\ref{lemma:Xi:alternative}}{\ge}
            &                                                         & n\pi_0e_t + \tfrac{n \mu \pi_0}{4} r_t - n L \pi_0 \Xi_t
    \end{align*}
    where the first inequality and the second inequality uses the $L$-smoothness and $\mu$-convexity of $f_i$.

    Combine both $T_1$, $T_2$ and \Cref{prop:algo1:convex:bounded_noise} we have
    \begin{align*}
        r_{t+1} \le & r_{t} + \gamma^2 n^2 \pi_0^2  (2L^2  \Xi_t + 4L e_t) - 2\gamma n \pi_0(e_t + \tfrac{\mu}{4} r_t - L  \Xi_t)  \\
                    & + \gamma^2 n(3L^2\pi_0^2  \Xi_t + 6L \pi_0^2 e_t + 3\pi_0^2 \bar{\sigma}^2)
        \\
        =           & (1-\tfrac{\gamma\mu n \pi_0}{2}) r_t - (2\gamma n \pi_0- 4 L\gamma^2 n^2 \pi_0^2 - 6L\gamma^2 n \pi_0^2) e_t \\
                    & +  (2\gamma^2 L^2 n^2 \pi_0^2 + 2 \gamma L n \pi_0 + 3L^2\gamma^2 n \pi_0^2)\Xi_t
        + 3\gamma^2n\pi_0^2 \bar{\sigma}^2
    \end{align*}
    In addition if $\gamma\le \frac{1}{10 L n \pi_0}$, then we can simplify the coefficient of $e_t$ and $\Xi_t$
    \begin{align*}
        4 L\gamma^2 n^2 \pi_0^2 + 6L\gamma^2 n \pi_0^2 \le                         & \gamma n\pi_0      \\
        2\gamma^2 L^2 n^2 \pi_0^2 + 2 \gamma L n \pi_0 + 3L^2\gamma^2 n\pi_0^2 \le & 4 \gamma L n \pi_0
    \end{align*}
    Then
    \begin{align*}
        r_{t+1} \le (1-\tfrac{\gamma\mu n \pi_0}{2}) r_t - \gamma n \pi_0e_t + 4\gamma L n \pi_0\Xi_t
        + 3\gamma^2n\pi_0^2 \bar{\sigma}^2.\qedhere
    \end{align*}
\end{proof}

\begin{lemma}\label{lemma:alg1:convex:consensus_distance}
    For $\gamma\le \tfrac{p}{10LmC_1}$ we have
    \begin{align*}
        \frac{1}{T+1}\sum_{t=0}^T \Xi_t
        \le & C_1^2\gamma^2 m^2 \frac{24}{p} \frac{\bar{\sigma^2}}{n}
        + \frac{80Lm^2}{p^2} C_1^2 \gamma^2 \frac{1}{T+1}\sum_{t=0}^T e_t
    \end{align*}
    where $C_1$ is defined in \Cref{definition:C}.
\end{lemma}
\begin{proof}
    First bound the consensus distance as follows:
    \begin{align*}
        n\Xi_{t}= & \E\| \mY^{(t)} - \bar{\mY}^{(t)} \|_F^2
        \le \E\| (\mY^{(t)} - \bar{\mY}^{(t-m)}) - (\bar{\mY}^{(t)}-\bar{\mY}^{(t-m)}) \|_F^2 \\
        \le       & \E\| \mY^{(t)} - \bar{\mY}^{(t-m)} \|_F^2
    \end{align*}
    where the last inequality we use the simple matrix inequality \eqref{eq:frobenius-avg}. For $t\ge m$ unroll to $t-m$.
    \begin{align*}
        n\Xi_{t}\le & \E \left\| \mW^m \mY^{(t-m)} -\gamma\sum_{k=t-m}^{t-1}\mW^{t-1-k}\tilde{\mW}\mG^{(k)} - \bar{\mY}^{(t-m)} \right\|_F^2
    \end{align*}
    Separate the stochastic part and deterministic part.
    \begin{align*}
        n\Xi_{t}\le & \left\| \mW^m \mY^{(t-m)} -\gamma\sum_{k=t-m}^{t-1}\mW^{t-1-k}\tilde{\mW}\E\mG^{(k)} - \bar{\mY}^{(t-m)} \right\|_F^2 \\
                    & + \E \left\|\gamma\sum_{k=t-m}^{t-1}\mW^{t-1-k}\tilde{\mW}(\E\mG^{(k)} - \mG^{(k)} )\right\|_F^2                      \\
        \le         & \left\| \mW^m \mY^{(t-m)} -\gamma\sum_{k=t-m}^{t-1}\mW^{t-1-k}\tilde{\mW}\E\mG^{(k)} - \bar{\mY}^{(t-m)} \right\|_F^2 \\
                    & +\gamma^2 m \sum_{k=t-m}^{t-1} \E \left\|\mW^{t-1-k}\tilde{\mW}(\E\mG^{(k)} - \mG^{(k)} )\right\|_F^2
    \end{align*}
    Given $\tilde{\mI}$ and $C_1$ in defined in \Cref{definition:C}, we know that $\tilde{\mW} =\tilde{\mI} \tilde{\mW}$. Then use \eqref{eq:frobenius-spectral} and \Cref{prop:algo1:convex:bounded_noise:matrix_form}
    \begin{align*}
        n\Xi_{t}\le & \left\| \mW^m \mY^{(t-m)} -\gamma\sum_{k=t-m}^{t-1}\mW^{t-1-k}\tilde{\mW}\E\mG^{(k)} - \bar{\mY}^{(t-m)} \right\|_F^2 \\
                    & + C_1^2\gamma^2 m  \sum_{k=t-m}^{t-1} \E \left\|\tilde{\mW}(\E\mG^{(k)} - \mG^{(k)} )\right\|_F^2                     \\
        \le         & \left\| \mW^m \mY^{(t-m)} -\gamma\sum_{k=t-m}^{t-1}\mW^{t-1-k}\tilde{\mW}\E\mG^{(k)} - \bar{\mY}^{(t-m)} \right\|_F^2 \\
                    & + C_1^2\gamma^2 m  \sum_{k=t-m}^{t-1}  3(L^2 \Xi_k + 2L e_k + \bar{\sigma}^2)
    \end{align*}
    Separate the first term as
    \begin{align*}
        n\Xi_{t}
        \le & (1+\alpha)\left\| \mW^m \mY^{(t-m)} - \bar{\mY}^{(t-m)} \right\|_F^2
        +(1+ \frac{1}{\alpha})\left\| \gamma\sum_{k=t-m}^{t-1}\mW^{t-1-k}\tilde{\mW}\E\mG^{(k)}\right\|_F^2 \\
            & + C_1^2\gamma^2 m  \sum_{k=t-m}^{t-1}  3(L^2 \Xi_k + 2L e_k + \bar{\sigma}^2)                 \\
        \le & (1+\alpha)(1-p)^{2m}\left\| \mY^{(t-m)} - \bar{\mY}^{(t-m)} \right\|_F^2
        +(1+ \frac{1}{\alpha})\left\| \gamma\sum_{k=t-m}^{t-1}\mW^{t-1-k}\tilde{\mW}\E\mG^{(k)}\right\|_F^2 \\
            & + C_1^2\gamma^2 m  \sum_{k=t-m}^{t-1}  3(L^2 \Xi_k + 2L e_k + \bar{\sigma}^2)
    \end{align*}
    where the first inequality uses $(a+b)^2\le (1+\epsilon)a^2 + (1+\frac{1}{\epsilon})b^2$ and take $\epsilon=(\frac{2-p}{2-2p})^{2m} - 1$.
    \begin{align*}
        1+\tfrac{1}{\epsilon} \le 1+ \tfrac{1-p}{mp} \le 1 + \tfrac{1}{mp} \le \tfrac{2}{p}.
    \end{align*}
    Then by applying our key lemma (\Cref{lemma:algo1:key}) we have
    \begin{align*}
        n\Xi_{t}
        \ \le\  & \Big(1-\frac{p}{2}\Big)^{2m}\left\| \mY^{(t-m)} - \bar{\mY}^{(t-m)} \right\|_F^2
        + \frac{2m}{p} C_1^2 \gamma^2 \sum_{k=t-m}^{t-1} \left\| \tilde{\mW}\E\mG^{(k)}\right\|_F^2 \\
                & + C_1^2\gamma^2 m  \sum_{k=t-m}^{t-1}  3(L^2 \Xi_k + 2L e_k + \bar{\sigma}^2)
    \end{align*}
    Next we bound $ \E\| \tilde{\mW} \mG^{(t')}\|^2_F$,
    \begin{align*}
        \E\| \tilde{\mW} \E\mG^{(k)}\|^2_F =                        & \tsum_{i=1}^n \E\| \tfrac{1}{n} \tsum_{j=1}^n \nabla f_j(\xx_j^{(k-\tau_{ij})}) \|^2                                                                                        \\
        =                                                           & \tsum_{i=1}^n \E\| \tfrac{1}{n} \tsum_{j=1}^n (\nabla f_j(\xx_j^{(k-\tau_{ij})}) - \nabla f_j(\bar{\xx}^{(k)}) + \nabla f_j(\bar{\xx}^{(k)}) - \nabla f_j(\xx^\star) ) \|^2 \\
        \le                                                         & \tfrac{2}{n}\tsum_{i=1}^n \tsum_{j=1}^n (\|\nabla f_j(\xx_j^{(k-\tau_{ij})}) - \nabla f_j(\bar{\xx}^{(k)}) \|^2
        + \|\nabla f_j(\bar{\xx}^{(k)}) - \nabla f_j(\xx^\star) ) \|^2)                                                                                                                                                                           \\
        \le                                                         & \tfrac{2}{n}\tsum_{i=1}^n \tsum_{j=1}^n (L^2\|\xx_j^{(k-\tau_{ij})} - \bar{\xx}^{(k)}\|^2
        + \|\nabla f_j(\bar{\xx}^{(k)}) - \nabla f_j(\xx^\star) ) \|^2)                                                                                                                                                                           \\
        \stackrel{\text{Lemma }\ref{lemma:Xi:alternative}}{\le}     &
        2L^2 n  \Xi_{k} + 2 \tsum_{j=1}^n \|\nabla f_j(\bar{\xx}^{(k)}) - \nabla f_j(\xx^\star) ) \|^2                                                                                                                                            \\
        \stackrel{\text{Smoothness }\eqref{eq:smooth:optimum}}{\le} &
        2L^2 n  \Xi_{k} + 4 nL e_{k}.
    \end{align*}
    Then
    \begin{align*}
        n\Xi_{t}
        \le & (1-\frac{p}{2})^{2m} n \Xi_{t-m}
        + \frac{2m}{p} C_1^2 \gamma^2 \sum_{k=t-m}^{t-1} (2L^2n\Xi_k+4nLe_k)
        + C_1^2\gamma^2 m  \sum_{k=t-m}^{t-1}  3(L^2 \Xi_k + 2L e_k + \bar{\sigma}^2)
    \end{align*}
    Then
    \begin{align*}
        \Xi_{t}
        \le & (1-\frac{p}{2})^{2m} \Xi_{t-m}
        + \frac{2m}{p} C_1^2 \gamma^2 \sum_{k=t-m}^{t-1} (5L^2\Xi_k+10Le_k)
        + 3 C_1^2\gamma^2 m^2 \frac{\bar{\sigma}^2}{n}.
    \end{align*}
    \paragraph{Unroll for $t<m$.} We can apply similar steps
    \begin{align*}
        n\Xi_t\le & \E \left\| \mW^{(t)}\mY^{(0)} - \gamma \sum_{k=0}^{t-1}\mW^{t-1-k}\tilde{\mW}\mG^{(k)} - \bar{\mY}^{(0)} \right\|_F^2
        = \E \left\| \gamma \sum_{k=0}^{t-1}\mW^{t-1-k}\tilde{\mW}\mG^{(k)}\right\|_F^2                                                   \\
        \le       & C_1^2\gamma^2m\sum_{k=0}^{t-1} \E \left\| \tilde{\mW}\mG^{(k)}\right\|_F^2
        \le 2C_1^2\gamma^2m\sum_{k=0}^{t-1} (5L^2n\Xi_k+10nLe_k+3\bar{\sigma}^2)
    \end{align*}
    \paragraph{Merge two parts together and sum over $t$.}
    \begin{align*}
        \frac{1}{T+1}\sum_{t=0}^T \Xi_t
        \le & \left(1-\frac{p}{2}\right)^{2m} \frac{1}{T+1}\sum_{t=m}^T  \Xi_{t-m}+ 6 C_1^2\gamma^2 m^2 \frac{\bar{\sigma}^2}{n}                                                                                         \\
            & + \frac{2m}{p} C_1^2 \gamma^2 \frac{1}{T+1}
        \left(
        \sum_{t=m}^T \sum_{k=t-m}^{t-1} (5L^2\Xi_k+10Le_k)
        +
        \sum_{t=0}^{m-1} \sum_{k=t-m}^{t-1} (5L^2\Xi_k+10Le_k)
        \right)                                                                                                                                                                                                          \\
        \le & \left(1-\frac{p}{2}\right)^{2m} \frac{1}{T+1}\sum_{t=0}^T  \Xi_{t}+ 6C_1^2\gamma^2 m^2 \frac{\bar{\sigma}^2}{n}    + \frac{2m^2}{p} C_1^2 \gamma^2 \frac{1}{T+1}\sum_{t=0}^T \left(5L^2\Xi_t+10Le_t\right)
    \end{align*}
    By taking $\gamma\le\frac{p}{10CLm}$, then $\frac{10L^2m^2}{p} C_1^2 \gamma^2\le\frac{p}{4}$.
    \begin{align*}
        \frac{1}{T+1}\sum_{t=0}^T \Xi_t
        \le & C_1^2\gamma^2 m^2 \frac{24}{p} \frac{\bar{\sigma}^2}{n}
        + \frac{80Lm^2}{p^2} C_1^2 \gamma^2 \frac{1}{T+1}\sum_{t=0}^T e_t. \qedhere
    \end{align*}
\end{proof}
\begin{lemma}[{Identical to \citep[Lemma 15]{koloskova2021unified}}]\label{lemma:final_recursion} %
    For any parameters $r_0\ge0,a\ge0,b\ge0,c\ge0$ there exists constant stepsizes $\gamma\le\frac{1}{c}$ such that
    \begin{align*}
        \Psi_T:=\frac{r_0}{\gamma(T+1)} + a\gamma + b\gamma^2\le 2\left(\frac{ar_0}{T+1}\right)^{\tfrac{1}{2}}
        +2b^{\tfrac{1}{3}}\left(\frac{r_0}{T+1}\right)^{\tfrac{2}{3}} + \frac{cr_0}{T+1}.
    \end{align*}
\end{lemma}
\begin{theorem}
    If $\gamma \le \frac{p}{30 Lm C_1}$, then
    \begin{align*}
        \tfrac{1}{T+1}\tsum_{t=0}^T \left(f(\bar{\xx}^{(t)}) - f(\xx^\star)\right)
        \le 8 \left( \frac{\bar{\sigma}^2r_0}{n(T+1)} \right)^{\tfrac{1}{2}}
        + 2 \left(
        \frac{16Cm\sqrt{L} \bar{\sigma} r_0 }{\sqrt{p}(T+1)}
        \right)^{\tfrac{2}{3}}
        + \frac{30Lm \sqrt{n} C r_0}{p(T+1)}.
    \end{align*}
    where $r_0=\| \xx^0 - \xx^\star \|^2$ and $C=C(\mW)$ is defined in \Cref{definition:C}.
\end{theorem}
\begin{proof}
    Reorganize \Cref{lemma:alg1:convex:descent} and average over time
    \begin{align*}
        \tfrac{1}{T+1}\tsum_{t=0}^T e_t
        \le \tfrac{1}{T+1}\tsum_{t=0}^T \left(
        \tfrac{r_{t}}{\gamma n \pi_0}
        - \tfrac{r_{t+1}}{\gamma n \pi_0}
        \right)
        + \tfrac{4L}{T+1} \tsum_{t=0}^T \Xi_t
        + 3\gamma\pi_0 \bar{\sigma}^2.
    \end{align*}
    Combining with \Cref{lemma:alg1:convex:consensus_distance} gives
    \begin{align*}
        \tfrac{1}{T+1}\tsum_{t=0}^T e_t
        \le \tfrac{1}{T+1}\tfrac{r_{0}}{\gamma n \pi_0}
        + 4L \left(
        C_1^2\gamma^2 m^2 \frac{24}{p} \frac{\bar{\sigma^2}}{n}
        + \frac{80Lm^2}{p^2} C_1^2 \gamma^2 \frac{1}{T+1}\sum_{t=0}^T e_k
        \right)
        + 3\gamma\pi_0\bar{\sigma}^2
    \end{align*}
    Select $\gamma \le \frac{p}{30 Lm C_1}$ such that $\tfrac{320L^2}{p^2}\gamma^2 m^2 C_1^2\le\frac{1}{2}$
    \begin{align*}
        \tfrac{1}{T+1}\tsum_{t=0}^T e_t
        \le \tfrac{2}{T+1}\tfrac{r_{0}}{\gamma n \pi_0}
        + 6\gamma\pi_0 \bar{\sigma}^2
        + \tfrac{96L}{p}\gamma^2 m^2 C_1^2 \frac{\bar{\sigma}^2}{n}.
    \end{align*}
    Applying \Cref{lemma:final_recursion} gives
    \begin{align*}
        \frac{1}{T+1}\sum_{t=0}^T e_t
        \le 40 \left( \frac{\bar{\sigma}^2r_0}{n(T+1)} \right)^{\tfrac{1}{2}}
        + 2 \left(
        \frac{\sqrt{mL} \bar{\sigma} r_0 }{\sqrt{p}(T+1)} \frac{16 C_1 \sqrt{m}}{n\pi_0 \sqrt{n}}
        \right)^{\tfrac{2}{3}}
        + \frac{dr_0}{n\pi_0(T+1)}
    \end{align*}
    where  $d=\max\{\frac{30 Lm C_1}{p}, 10L n\pi_0\}=\frac{30 Lm C_1}{p}$. As in \Cref{lemma:C},
    $$C_1=C \| \1\mpi^\top\tilde{\mI}\|= C n \sqrt{\tau_{\max}+1} \pi_0 \le C n \sqrt{n} \pi_0. $$
    We can further simplify it as
    \begin{align*}
        \frac{1}{T+1}\sum_{t=0}^T e_t
        \le 40 \left( \frac{\bar{\sigma}^2r_0}{n(T+1)} \right)^{\tfrac{1}{2}}
        + 2 \left(
        \frac{16Cm\sqrt{L} \bar{\sigma} r_0 }{\sqrt{p}(T+1)}
        \right)^{\tfrac{2}{3}}
        + \frac{30Lm \sqrt{n} C r_0}{p(T+1)}. \qedhere
    \end{align*}
\end{proof}

\subsection{Proof of Theorem~\ref{thm:1} in the strongly convex case}
\label{ssec:strongly-convex}
The proof for strongly convex objective follows similar lines as \cite{stich2019unified}:
\begin{theorem}
    Let $a=\frac{\mu n\pi_0}{2}$, $b=\frac{2}{n\pi_0}$, $c=6\pi_0\bar{\sigma}^2$, $A=\frac{400L}{p^2}m^2C_1^2\bar{\sigma}^2$, and
    let $\gamma=\frac{1}{s}\le \frac{1}{aT} \ln\max\{\frac{ba^2T^2r_0}{c}, 2\}$, then
    \begin{align*}
        \frac{1}{W_T} \sum_{t=0}^T w_t e_t + \mu r_{T+1}
        \le &
        \tilde{\cO} \left(
        bsr_0 \exp\left[-\frac{a(T+1)}{s}\right] + \frac{c}{a(T+1)} +  \frac{A}{a^2(T+1)^2}
        \right)
    \end{align*}
    where $w_t=(1-\frac{\mu\gamma n\pi_0}{2})^{-(t+1)}$.
\end{theorem}
\begin{proof}
    From \Cref{lemma:alg1:convex:descent} we know that if $\gamma \le \frac{1}{10L n\pi_0}$, then
    \begin{align*}
        r_{t+1} \le (1-\tfrac{\gamma\mu n \pi_0}{2}) r_t - \gamma n \pi_0e_t + 4\gamma L n \pi_0 \Xi_t
        + 3\gamma^2n\pi_0^2 \bar{\sigma}^2.
    \end{align*}
    Then
    \begin{align*}
        e_t \le \frac{1}{\gamma n\pi_0} (1-\frac{\mu\gamma n\pi_0}{2}) r_t - \frac{1}{\gamma n\pi_0} r_{t+1}
        + 4L \Xi_t
        + 3\gamma \pi_0 \bar{\sigma}^2.
    \end{align*}
    Multiply $w_t$ and sum over $t=0$ to $T$ and divided by $W_T$
    \begin{align*}
        \frac{1}{W_T} \sum_{t=0}^T w_t e_t \le &
        \frac{1}{W_T} \sum_{t=0}^T
        \left(
        \frac{1-\frac{\mu\gamma n\pi_0}{2}}{\gamma n\pi_0} w_t r_t
        -  \frac{w_t}{\gamma n\pi_0}  r_{t+1}
        \right)
        + \frac{4L}{W_T} \sum_{t=0}^T w_t \Xi_t
        + 3\gamma \pi_0 \bar{\sigma}^2.
    \end{align*}
    Set $(1-\frac{\mu\gamma n\pi_0}{2})w_{t+1}=w_t$, then
    \begin{align*}
        \frac{1}{W_T} \sum_{t=0}^T w_t e_t\le &
        \frac{1}{W_T}
        \left(
        \frac{1-\frac{\mu\gamma n\pi_0}{2}}{\gamma n\pi_0} w_0 r_0
        -  \frac{1-\frac{\mu\gamma n\pi_0}{2}}{\gamma n\pi_0} w_{T+1} r_{T+1}
        \right)
        + \frac{4L}{W_T} \sum_{t=0}^T w_t \Xi_t
        + 3\gamma \pi_0 \bar{\sigma}^2.
    \end{align*}
    Then using \Cref{lemma:alg1:convex:consensus_distance} we have
    \begin{align*}
            & \frac{1}{W_T} \sum_{t=0}^T w_t e_t +
        \frac{1-\frac{\mu\gamma n\pi_0}{2}}{\gamma n\pi_0 W_T} w_{T+1} r_{T+1} \\
        \le &
        \frac{1}{W_T} \frac{1-\frac{\mu\gamma n\pi_0}{2}}{\gamma n\pi_0} w_0 r_0
        + 4L \left(\tfrac{80C_1^2Lm^2}{p^2}\gamma^2 \tfrac{1}{W_T}\tsum_{t'=0}^{T} w_t \ee_{t'}
        + \tfrac{24}{p}\gamma^2 m^2 C_1^2 \frac{\bar{\sigma}^2}{n} \right)
        + 3\gamma \pi_0 \bar{\sigma}^2
    \end{align*}
    By taking $\gamma\le\frac{p}{30LmC_1}$ we have $\frac{320L^2m^2C_1^2 \gamma^2}{p^2}\le \frac{1}{2}$, then
    \begin{align*}
        \frac{1}{W_T} \sum_{t=0}^T w_t e_t +
        \frac{1-\frac{\mu\gamma n\pi_0}{2}}{\gamma n\pi_0 W_T} 2w_{T+1} r_{T+1}
        \le &
        \frac{1}{W_T} \frac{1-\frac{\mu\gamma n\pi_0}{2}}{\gamma n\pi_0} 2w_0 r_0
        + 6\gamma \pi_0 \bar{\sigma}^2
        + \tfrac{400L}{p^2}\gamma^2 m^2 C_1^2 \bar{\sigma}^2
    \end{align*}
    Since $W_T \ge w_T=(1-\frac{\mu\gamma n\pi_0}{2})^{-(T+1)}$ and $W_T\le \frac{2w_T}{\mu\gamma n\pi_0}$
    \begin{align*}
        \frac{1}{W_T} \sum_{t=0}^T w_t e_t +
        \mu r_{T+1}
        \le &
        \frac{(1-\frac{\mu\gamma n\pi_0}{2})^{T+1}}{\gamma n\pi_0} 2w_0 r_0
        + 6\gamma \pi_0 \bar{\sigma}^2
        + \tfrac{400L}{p^2}\gamma^2 m^2 C_1^2 \bar{\sigma}^2                       \\
        \le & \frac{e^{-\frac{\mu \gamma n\pi_0}{2}(T+1)}}{\gamma n\pi_0} 2w_0 r_0
        + 6\gamma \pi_0 \bar{\sigma}^2
        + \tfrac{400L}{p^2}\gamma^2 m^2 C_1^2 \bar{\sigma}^2
    \end{align*}
    Let $a=\frac{\mu n\pi_0}{2}$, $b=\frac{2}{n\pi_0}$, $c=6\pi_0\bar{\sigma}^2$, $A=\frac{400L}{p^2}m^2C_1^2\bar{\sigma}^2$, then
    \begin{align*}
        \frac{1}{W_T} \sum_{t=0}^T w_t e_t + \mu r_{T+1}
        \le &
        \frac{br_0}{\gamma} \exp[-a\gamma(T+1)] + c \gamma + A \gamma^2
    \end{align*}
    \paragraph{Tuning stepsize.}
    Let $\gamma=\frac{1}{d}\le \frac{1}{aT} \ln\max\{\frac{ba^2T^2r_0}{c}, 2\}$, then
    \begin{align*}
        \frac{1}{W_T} \sum_{t=0}^T w_t e_t + \mu r_{T+1}
        \le &
        \tilde{\cO} \left(
        bsr_0 \exp[-\frac{a(T+1)}{s}] + \frac{c}{a(T+1)} +  \frac{A}{a^2(T+1)^2}
        \right). \qedhere
    \end{align*}
\end{proof}

\subsection{Proof of Theorem~\ref{thm:1} in the non-convex case}\label{ssec:non-convex}
Let $\bar{\xx}^{(t)}:=\left(\mpi^\top \mY ^{(t)}\right)^\top$ and $\bar{\mY}^{(t)}:= \1\mpi^\top \mY ^{(t)}$. Let $f^\star$ be the optimal objective value at critical points.
We can define the following iterates
\begin{enumerate}[nosep]
    \item $r_t:= \E f(\bar{\xx}^{(t)}) - f^\star$ is the \textit{expected function suboptimality}.
    \item $e_t:=\| \nabla f(\bar{\xx}^{(t)}) \|^2 $
    \item $\Xi_t:= \frac{1}{n} \| \bar{\mY}^{(t)} - \mY^{(t)} \|^2_F$ is the \textit{consensus distance}.
\end{enumerate}
where the expectation is taken with respect to $\xiv^{(t)}\in\R^n$ the randomness across all workers at time $t$. Note that \Cref{lemma:Xi:alternative} still holds.

\Cref{prop:algo1:nonconvex:stochasticity} and \Cref{prop:algo1:nonconvex:stochasticity:matrix_form} bound the stochastic noise of the gradient.
\begin{proposition}\label{prop:algo1:nonconvex:stochasticity}
    Under \Cref{assumption:uniform_sigma_zeta},  we have
    \begin{align}\label{eq:prop:algo1:nonconvex:stochasticity:uniform}
        \E\norm{ \mpi^\top \tilde{\mW} (\mG^{(t)} - \E\mG^{(t)})}^2
        \le
        n \pi_0^2 \bar{\sigma}^2.
    \end{align}
\end{proposition}
\begin{proof} Denote $\E=\E_{\xiv}$. Use Cauchy-Schwartz inequality \Cref{eq:cs}
    \begin{align*}
        \E\norm{ \mpi^\top \tilde{\mW} (\mG^{(t)} - \E\mG^{(t)})}^2=
            & \E \left\| \frac{\pi_0}{n} \sum_{i=1}^n\sum_{j=1}^n (\nabla F_j(\xx_j^{(t-\tau_{ij})}; \xi_j^{(t-\tau_{ij})}) - \nabla f_j(\xx_j^{(t-\tau_{ij})}) ) \right\|^2   \\
        \le & \frac{\pi_0^2}{n}  \sum_{i=1}^n  \E \left\| \sum_{j=1}^n \nabla F_j(\xx_j^{(t-\tau_{ij})}; \xi_j^{(t-\tau_{ij})}) - \nabla f_j(\xx_j^{(t-\tau_{ij})}) \right\|^2
    \end{align*}
    Now the randomness inside the norm are independent
    \begin{align*}
         & \E\norm{ \mpi^\top \tilde{\mW} (\mG^{(t)} - \E\mG^{(t)})}^2
        \E\norm{ \mpi^\top \tilde{\mW} (\mG^{(t)} - \E\mG^{(t)})}^2
        \le
        n \pi_0^2 \bar{\sigma}^2. \qedhere
    \end{align*}
\end{proof}

\begin{proposition}\label{prop:algo1:nonconvex:stochasticity:matrix_form}
    Under \Cref{assumption:uniform_sigma_zeta},  we have
    \begin{align}\label{eq:prop:algo1:nonconvex:stochasticity:uniform:matrix_form}
        \E\norm{\tilde{\mW} (\mG^{(t)} - \E\mG^{(t)})}^2_F
        \le
        \bar{\sigma}^2.
    \end{align}
\end{proposition}

Next we establish the recursion of $r_t$
\begin{lemma}[Descent lemma for non-convex case]\label{lemma:nonconvex:recursion}
    Under \Cref{assumption:smoothness:stochastic} and \ref{assumption:uniform_sigma_zeta}.
    Let $\gamma \le \frac{1}{8 L n \pi_0}$, then
    \begin{align*}
        r_{t+1} \le & r_t - \frac{\gamma n \pi_0}{4}  e_t
        + 2\gamma  L^2 n \pi_0 \Xi_t
        + 2 \gamma^2 L n \pi^2_0 \bar{\sigma}^2.
    \end{align*}
\end{lemma}

\begin{proof}
    Since $f$ is $L$-smooth,
    \begin{align*}
        \E f(\bar{\xx}^{(t+1)})
        =   & \E f(\bar{\xx}^{(t)} - \gamma \mpi^\top \tilde{\mW} \mG^{(t)} ) \\
        \le & f(\bar{\xx}^{(t)})
        - \gamma  \underbrace{\langle \nabla f(\bar{\xx}^{(t)}),
            \mpi^\top \tilde{\mW} \E\mG^{(t)} \rangle}_{:=T_1}
        + \tfrac{\gamma^2L}{2} \underbrace{\E\norm{ \mpi^\top \tilde{\mW} \mG^{(t)} }^2}_{:=T_2}
    \end{align*}
    The first-order term $T_1$ has a lower bound
    \begin{align*}
        T_1 = &
        n\pi_0\langle \nabla f(\bar{\xx}^{(t)}),
        \tfrac{1}{n\pi_0} \mpi^\top \tilde{\mW} \E\mG^{(t)} \rangle \\
        =     & n\pi_0 \left(\norm{\nabla f(\bar{\xx}^{(t)})}^2
        +\langle
        \nabla f(\bar{\xx}^{(t)}),
        \tfrac{1}{n\pi_0}\mpi^\top \tilde{\mW} \E\mG^{(t)}- \nabla f(\bar{\xx}^{(t)})
        \rangle                                      \right)        \\
        \ge   & n\pi_0 \left(
        \tfrac{1}{2} \norm{\nabla f(\bar{\xx}^{(t)})}^2
        - \tfrac{1}{2} \norm{\tfrac{1}{n\pi_0}\mpi^\top \tilde{\mW} \E\mG^{(t)}
                - \nabla f(\bar{\xx}^{(t)})
            }^2                           \right)                   \\
        =     & n\pi_0 \left(
        \tfrac{1}{2} e_t
        - \tfrac{1}{2n^4} \norm{\tsum_{i=1}^n \tsum_{j=1}^n (\nabla f_j(\xx_j^{(t-\tau_{ij})})
                - \nabla f_j(\bar{\xx}^{(t)}))
            }^2        \right)                                      \\
        \ge   & n\pi_0 \left(
        \tfrac{1}{2} e_t
        - \tfrac{L^2}{2n^2} \tsum_{i=1}^n \tsum_{j=1}^n \norm{ \xx_j^{(t-\tau_{ij})}
            - \bar{\xx}^{(t)}
        }^2        \right)                                          \\
        \ge   & n\pi_0 \left(
        \tfrac{1}{2} e_t
        - \tfrac{L^2}{2} \Xi_t   \right)
    \end{align*}
    as $a^2-\langle a, b \rangle \ge \frac{a^2}{2} - \frac{b^2}{2}$ for $a,b\ge0$.

    On the other hand, separate the stochastic part and deterministic part of $T_2$ we have
    \begin{align*}
        T_2 \le & 2\E\norm{ \mpi^\top \tilde{\mW} (\mG^{(t)} - \E\mG^{(t)})}^2 + 2 \norm{\mpi^\top \tilde{\mW} \E \mG^{(t)} }^2.
    \end{align*}
    Under \Cref{assumption:uniform_sigma_zeta} and \Cref{prop:algo1:nonconvex:stochasticity}, we know the first term
    \begin{align*}
        \E\norm{ \mpi^\top \tilde{\mW} (\mG^{(t)} - \E\mG^{(t)})}^2
        \le n \pi^2_0 \bar{\sigma}^2.
    \end{align*}
    Consider the second term
    \begin{align*}
        \norm{\mpi^\top \tilde{\mW} \E \mG^{(t)} }^2 = &
        \left\| \frac{\pi_0}{n} \sum_{i=1}^n \sum_{j=1}^{n} \nabla f_j(\xx_j^{(t-\tau_{ij})}) \right\|^2                                                                                              \\
        =                                              &
        n^2\pi_0^2 \left\| \frac{1}{n^2}  \sum_{i=1}^n \sum_{j=1}^{n} \nabla f_j(\xx_j^{(t-\tau_{ij})}) - \nabla f(\bar{\xx}^{(t)}) + \nabla f(\bar{\xx}^{(t)})\right\|^2                             \\
        \le                                            & 2 n^2\pi_0^2 \left\| \frac{1}{n^2}  \sum_{i=1}^n \sum_{j=1}^{n} (\nabla f_j(\xx_j^{(t-\tau_{ij})}) - \nabla f_j(\bar{\xx}^{(t)})) \right\|^2
        + 2 n^2\pi_0^2 \left\| \nabla f(\bar{\xx}^{(t)})\right\|^2                                                                                                                                    \\
        \le                                            & 2 \pi_0^2 \sum_{i=1}^n \sum_{j=1}^{n}\left\| \nabla f_j(\xx_j^{(t-\tau_{ij})}) - \nabla f_j(\bar{\xx}^{(t)}) \right\|^2
        +  2 n^2\pi_0^2 \left\| \nabla f(\bar{\xx}^{(t)})\right\|^2
    \end{align*}
    Combine \Cref{assumption:uniform_sigma_zeta} we have
    \begin{align*}
        \norm{\mpi^\top \tilde{\mW} \E \mG^{(t)} }^2 \le 2n^2\pi_0^2 (L^2 \Xi_t + e_t).
    \end{align*}
    Therefore, the $T_2$ can be bounded as follows
    \begin{equation}
        T_2 \le 4 n^2 \pi^2_0 (\tfrac{\bar{\sigma}^2}{n} + L^2 \Xi_t + e_t).
    \end{equation}
    Gathering everything together
    \begin{align*}
        r_{t+1} \le & r_t - \tfrac{\gamma n \pi_0}{2} (e_t -L^2\Xi_t) + 2 \gamma^2 L n^2 \pi^2_0 (\tfrac{\bar{\sigma}^2}{n} + L^2 \Xi_t + e_t) \\
        \le         & r_t - \tfrac{\gamma n \pi_0}{2} (1 - 4 \gamma L n\pi_0) e_t
        + \gamma  L^2 n \pi_0 (1 + 2 \gamma L  n\pi_0 ) \Xi_t
        + 2 \gamma^2 L n \pi^2_0 \bar{\sigma}^2
    \end{align*}
    Let $\gamma \le \frac{1}{8 L n \pi_0}$, then
    \begin{align*}
        r_{t+1} \le & r_t - \frac{\gamma n \pi_0}{4}  e_t
        + 2\gamma  L^2 n \pi_0 \Xi_t
        + 2 \gamma^2 L n \pi^2_0 \bar{\sigma}^2 . \qedhere
    \end{align*}
\end{proof}
Next we bound the consensus distance
\begin{lemma}[Bounded consensus distance]\label{lemma:nonconvex:bounded_consensus_distance}
    Under \Cref{assumption:uniform_sigma_zeta},
    \begin{align*}
        \frac{1}{T+1}\sum_{t=0}^T \Xi_t
        \le &
        \frac{16C^2m^2}{p^2} \gamma^2 \bar{\sigma}^2
        + \frac{16C^2m^2}{p^2} \gamma^2 \frac{1}{T+1} \sum_{t=0}^Te_k \,.
    \end{align*}
\end{lemma}
\begin{proof}
    First bound the consensus distance by inserting $\bar{\mY}^{(t-m)}$
    \begin{align*}
        n \Xi_t = &
        \E \| \bar{\mY}^{(t)} - \mY^{(t)} \|^2_F
        \le \E \| (\bar{\mY}^{(t)} - \bar{\mY}^{(t-m)}) - (\mY^{(t)} - \bar{\mY}^{(t-m)}) \|^2_F \\
        \le       & \E \| \mY^{(t)} - \bar{\mY}^{(t-m)} \|^2_F
    \end{align*}
    where we used $\|A - \bar{A}\|_F^2=\sum_{i=1}^n\|\aa_i - \bar{\aa} \|^2
        \le \sum_{i=1}^n \| \aa_i \|^2=\|A\|^2_F$.

    For $t\ge m$ unroll $\mY^{(t)}$ until $t-m$.
    \begin{align*}
        n \Xi_t \le & \E \left\| \mW^{m} \mY^{(t-m)} - \gamma \sum_{k=t-m}^{t-1} \mW^{t-1-k} \tilde{\mW} \mG^{(k)}  - \bar{\mY}^{(t-m)} \right\|^2_F
    \end{align*}
    Separate stochastic part and deterministic part
    \begin{align*}
        n \Xi_t \le & \left\| \mW^{m} \mY^{(t-m)} - \gamma \sum_{k=t-m}^{t-1} \mW^{t-1-k} \tilde{\mW} \E \mG^{(k)}  - \bar{\mY}^{(t-m)} \right\|^2_F \\
                    & + \E \left\| \gamma \sum_{k=t-m}^{t-1} \mW^{t-1-k} \tilde{\mW} (\E\mG^{(k)} - \mG^{(k)})\right\|^2_F
    \end{align*}
    then let $C_1^2$  defined in \Cref{definition:C} and use $\|AB\|_F^2\le\|A\|^2_F\|B\|^2$ and \eqref{eq:prop:algo1:nonconvex:stochasticity:uniform:matrix_form}
    \begin{align*}
        n \Xi_t \le & \left\| \mW^{m} \mY^{(t-m)} - \gamma \sum_{k=t-m}^{t-1} \mW^{t-1-k} \tilde{\mW} \E \mG^{(k)}  - \bar{\mY}^{(t-m)} \right\|^2_F                                     \\
                    & + C_1^2 \gamma^2 m \sum_{k=t-m}^{t-1} \E \left\|  \tilde{\mW} (\E\mG^{(k)} - \mG^{(k)})\right\|^2_F                                                                \\
        \le         & \left\| \mW^{m} \mY^{(t-m)} - \gamma \sum_{k=t-m}^{t-1} \mW^{t-1-k} \tilde{\mW} \E \mG^{(k)}  - \bar{\mY}^{(t-m)} \right\|^2_F + C_1^2 \gamma^2 m^2 \bar{\sigma}^2
    \end{align*}

    Apply Cauchy-Schwartz inequality with $\alpha>0$
    \begin{align*}
        n \Xi_t \le
         & (1+\alpha) \left\|\mW^{m} \mY^{(t-m)} - \bar{\mY}^{(t-m)} \right\|^2_F
        + (1+\tfrac{1}{\alpha})  \left\| \gamma \sum_{k=t-m}^{t-1} \mW^{t-1-k} \tilde{\mW} \E \mG^{(k)} \right\|^2_F + C_1^2 \gamma^2 m^2 \bar{\sigma}^2
    \end{align*}
    Applying \Cref{lemma:algo1:key} to the first term
    \begin{align*}
        n \Xi_t \le
         & (1+\alpha) (1-p)^{2m} \| \mY^{(t-m)} - \bar{\mY}^{(t-m)} \|_F^2
        + (1+\tfrac{1}{\alpha})  \left\| \gamma \sum_{k=t-m}^{t-1} \mW^{t-1-k} \tilde{\mW} \E \mG^{(k)} \right\|^2_F + C_1^2 \gamma^2 m^2 \bar{\sigma}^2
    \end{align*}
    Take $\alpha=(\frac{2-p}{2-2p})^{2m}-1=(1+\frac{p}{2-2p})^{2m}-1\ge \frac{mp}{1-p}$ and use
    \begin{align*}
        1+\tfrac{1}{\alpha} \le 1+ \tfrac{1-p}{mp} \le 1 + \tfrac{1}{mp} \le \tfrac{2}{p},
    \end{align*}
    then use $\|AB\|_F^2\le\|A\|^2_F\|B\|^2$
    \begin{align*}
        n \Xi_t \le
         & \left(1-\frac{p}{2}\right)^{2m} \| \mY^{(t-m)} - \bar{\mY}^{(t-m)} \|_F^2
        + \frac{2}{p} \left\| \gamma \sum_{k=t-m}^{t-1} \mW^{t-1-k} \tilde{\mW} \E\mG^{(k)} \right\|^2_F
        + C_1^2 \gamma^2 m^2 \bar{\sigma}^2                                          \\
        \le
         & \left(1-\frac{p}{2}\right)^{2m} \| \mY^{(t-m)} - \bar{\mY}^{(t-m)} \|_F^2
        + \frac{2C_1^2m}{p} \gamma^2 \sum_{k=t-m}^{t-1} \left\| \tilde{\mW} \E\mG^{(k)} \right\|^2_F
        + C_1^2 \gamma^2 m^2 \bar{\sigma}^2.
    \end{align*}
    where the second term can be expanded by
    \begin{align*}
        \norm{ \tilde{\mW} \E \mG^{(k)} }^2_F
        =   &
        \sum_{i=1}^n \left\|  \frac{1}{n} \sum_{j=1}^{n} \nabla f_j(\xx_j^{(k-\tau_{ij})}) \right\|^2                                                        \\
        =   &
        \sum_{i=1}^n \left\| \frac{1}{n}  \sum_{j=1}^{n} \nabla f_j(\xx_j^{(k-\tau_{ij})}) - \nabla f(\bar{\xx}^{(k)}) + \nabla f(\bar{\xx}^{(k)})\right\|^2 \\
        \le & 2 \sum_{i=1}^n \left\| \frac{1}{n}  \sum_{j=1}^{n} (\nabla f_j(\xx_j^{(k-\tau_{ij})}) - \nabla f_j(\bar{\xx}^{(k)})) \right\|^2
        + 2 n \left\| \nabla f(\bar{\xx}^{(k)})\right\|^2                                                                                                    \\
        \le & \frac{2}{n} \sum_{i=1}^n \sum_{j=1}^{n}\left\| \nabla f_j(\xx_j^{(k-\tau_{ij})}) - \nabla f_j(\bar{\xx}^{(k)}) \right\|^2
        +  2 n \left\| \nabla f(\bar{\xx}^{(k)})\right\|^2                                                                                                   \\
        \le & 2nL^2 \Xi_k + 2n e_k
    \end{align*}
    Combine and reduce the $n$ on both sides
    \begin{align*}
        \Xi_t \le & \left(1-\frac{p}{2}\right)^{2m} \Xi_{t-m}
        + 2C_1^2m^2 \gamma^2 \frac{\bar{\sigma}^2}{n}
        + \frac{4C_1^2m}{p} \gamma^2 \sum_{k=t-m}^{t-1} (L^2\Xi_k + e_k ).
    \end{align*}
    \paragraph{Unroll for $t<m$.} For $t<m$, we can apply similar steps
    \begin{align*}
        n \Xi_t
        \le & \E \left\| \mW^{(t)} \mY^{(0)} - \gamma \sum_{k=0}^{t-1} \mW^{t-1-k} \tilde{\mW} \mG^{(k)}  - \bar{\mY}^{(0)} \right\|^2_F
        = \E \left\| \gamma \sum_{k=0}^{t-1} \mW^{t-1-k} \tilde{\mW} \mG^{(k)}\right\|^2_F                                               \\
        \le &
        C_1^2 \gamma^2 m\sum_{k=0}^{t-1} \E\left\| \tilde{\mW} \mG^{(k)}\right\|^2_F
        \le 2C_1^2 m \gamma^2 \sum_{k=0}^{t-1} ( \bar{\sigma}^2 + nL^2 \Xi_k + n e_k).
    \end{align*}
    \paragraph{Finally, sum over $t$}
    \begin{align*}
        \frac{1}{T+1}\sum_{t=0}^T \Xi_t
        \le & \left(1-\frac{p}{2}\right)^{2m} \frac{1}{T+1}\sum_{t=m}^T \Xi_{t-m}
        + 2C_1^2m^2 \gamma^2 \frac{\bar{\sigma}^2}{n}                             \\
            &
        + \frac{4C_1^2m}{p} \gamma^2  \frac{1}{T+1}
        \left(
        \sum_{t=m}^T  \sum_{k=t-m}^{t-1} (L^2\Xi_k + e_k )
        + \sum_{t=0}^{m-1}  \sum_{k=0}^{t-1} (L^2\Xi_k + e_k)
        \right)                                                                   \\
        \le &
        \left(1-\frac{p}{2}\right)^{2m} \frac{1}{T+1}\sum_{t=0}^T \Xi_{t}
        + 2C_1^2m^2 \gamma^2 \frac{\bar{\sigma}^2}{n}                                      + \frac{4C_1^2m^2}{p}   \frac{\gamma^2}{T+1} \sum_{t=0}^T(L^2\Xi_k + e_k ).
    \end{align*}
    by taking $\gamma\le \frac{p}{4 CLm}$ we have $\frac{4C_1^2m^2}{p} \gamma^2L^2\le \frac{p}{4}$, then rearrange the all of the $\Xi$ terms
    \begin{align*}
        \frac{1}{T+1}\sum_{t=0}^T \Xi_t
        \le &
        \frac{16C_1^2m^2}{p} \frac{\bar{\sigma}^2}{n} \gamma^2
        + \frac{16C_1^2m^2}{p^2} \gamma^2 \frac{1}{T+1} \sum_{t=0}^Te_k \qedhere
    \end{align*}
\end{proof}

We can use the lemmas for recursion and the descent in the consensus distance to conclude the following theorem.
\begin{theorem}
    Under \Cref{assumption:smoothness:stochastic} and \Cref{assumption:uniform_sigma_zeta}.
    For $\gamma\le \frac{p}{16C_1Lm}$
    \begin{align*}
        \frac{1}{T+1}\sum_{t=0}^T \| \nabla f(\bar{\xx}^{(t)}) \|^2
        \le & 16 \left(
        \frac{2L\bar{\sigma}^2r_0}{n(T+1)}
        \right)^{\frac{1}{2}}
        + 2 \left(
        \frac{16CLm\bar{\sigma}}{\sqrt{p}} \frac{8r_0}{T+1}
        \right)^{\frac{2}{3}}
        + \frac{16C_1Lm}{p} \frac{r_0}{T+1}
    \end{align*}
    where $C=C(\mW)$ is defined in \Cref{definition:C} and $r_0=f(\xx^{(0)}) - f^\star$. Alternatively, for any target accuracy $\epsilon$,  $\frac{1}{T+1}\sum_{t=0}^T \| \nabla f(\bar{\xx}^{(t)}) \|^2\le \epsilon$ after
    \begin{align*}
        \cO\left(
        \frac{\bar{\sigma}^2}{n\epsilon^2}
        + \frac{Cm\bar{\sigma}}{\sqrt{p}\epsilon^{3/2}}
        + \frac{C_1m}{p\epsilon}
        \right) Lr_0
    \end{align*}
    iterations.
\end{theorem}
\begin{remark}
    For gossip averaging \cite{koloskova2021unified}, the rate with $\zeta^2=0$ is
    \begin{align*}
        \cO\left(
        \frac{\bar{\sigma}^2}{n\epsilon^2}
        +  \frac{\sqrt{m}\bar{\sigma}}{\sqrt{p}\epsilon^{3/2}}
        + \frac{m}{p\epsilon}
        \right) Lr_0.
    \end{align*}
\end{remark}
\begin{proof}
    From \Cref{lemma:nonconvex:recursion} we know that for $\gamma \le \frac{1}{8 L n \pi_0}$
    \begin{align*}
        r_{t+1} \le & r_t - \frac{\gamma n \pi_0}{4}  e_t
        + 2\gamma  L^2 n \pi_0 \Xi_t
        + 2 \gamma^2 L n \pi^2_0 \bar{\sigma}^2.
    \end{align*}
    Rearrange the terms and average over $t$
    \begin{align*}
        \frac{1}{T+1}\sum_{t=0}^T e_t
        \le & \frac{1}{T+1}\sum_{t=0}^T (\frac{4r_t}{\gamma n\pi_0} - \frac{4r_{t+1}}{\gamma n\pi_0})
        +  \frac{8L^2}{T+1}\sum_{t=0}^T \Xi_t + 8 L \pi_0 \gamma \bar{\sigma}^2                       \\
        \le & \frac{1}{T+1} \frac{4r_0}{\gamma n\pi_0}
        + \frac{8L^2}{T+1}\sum_{t=0}^T \Xi_t + 8 L \pi_0 \gamma \bar{\sigma}^2
    \end{align*}
    On the other hand, from \Cref{lemma:nonconvex:bounded_consensus_distance} for $\gamma\le \frac{p}{4 C_1Lm}$ we have
    \begin{align*}
        \frac{1}{T+1}\sum_{t=0}^T \Xi_t
        \le &
        \frac{16C_1^2m^2}{p} \frac{\bar{\sigma}^2}{n} \gamma^2
        + \frac{16C_1^2m^2}{p^2} \gamma^2 \frac{1}{T+1} \sum_{t=0}^Te_k.
    \end{align*}
    Then
    \begin{align*}
        \frac{1}{T+1}\sum_{t=0}^T e_t
        \le & \frac{1}{T+1} \frac{4r_0}{\gamma n\pi_0}
        + 8L^2 \frac{16C_1^2m^2}{p^2} \gamma^2 \left( \frac{p \bar{\sigma}^2}{n}
        + \frac{1}{T+1} \sum_{t=0}^Te_k\right) + 8 L \pi_0 \gamma \bar{\sigma}^2
    \end{align*}
    By taking $\gamma\le\frac{p}{16C_1Lm}$ such that $8L^2 \frac{16C_1^2m^2}{p^2} \gamma^2\le \frac{1}{2}$,
    then
    \begin{align*}
        \frac{1}{T+1}\sum_{t=0}^T e_t
        \le & \frac{1}{T+1} \frac{8r_0}{\gamma n\pi_0}
        + 16 L \pi_0 \gamma \bar{\sigma}^2
        +  \frac{16^2L^2C_1^2m^2}{np} \gamma^2\bar{\sigma}^2
    \end{align*}
    Then applying \Cref{lemma:final_recursion} we have
    \begin{align*}
        \frac{1}{T+1}\sum_{t=0}^T e_t
        \le & 32 \left(
        \frac{L\bar{\sigma}^2r_0}{n(T+1)}
        \right)^{\frac{1}{2}}
        + 2 \left(
        \frac{16C_1Lm\bar{\sigma}}{\sqrt{np}} \frac{8r_0}{n\pi_0 (T+1)}
        \right)^{\frac{2}{3}}
        + \frac{dr_0}{T+1}
    \end{align*}
    where $d=\max \{\frac{16C_1Lm}{p}, 8 L n \pi_0 \} = \frac{16C_1Lm}{p}$. As in \Cref{lemma:C},
    $$C_1=C \| \1\mpi^\top\tilde{\mI}\|= C n \sqrt{\tau_{\max}+1} \pi_0 \le C n \sqrt{n} \pi_0. $$
    We can further simplify it as
    \begin{align*}
        \frac{1}{T+1}\sum_{t=0}^T e_t
        \le & 32 \left(
        \frac{L\bar{\sigma}^2r_0}{n(T+1)}
        \right)^{\frac{1}{2}}
        + 2 \left(
        \frac{16CLm\bar{\sigma}}{\sqrt{p}} \frac{8r_0}{T+1}
        \right)^{\frac{2}{3}}
        + \frac{dr_0}{T+1}. \qedhere
    \end{align*}
\end{proof}
\section{Detailed experimental setup}\label{apx:setup}

\subsection{\cifar}\label{apx:setup:cifar}

\autoref{tab:cifar-experimental-settings}

\begin{table}[h]
	\caption{Default experimental settings for Cifar-10/VGG-11}
	\scriptsize%
	\label{tab:cifar-experimental-settings}%
	\begin{tabularx}{\linewidth}{lX}
		\toprule
		Dataset              & Cifar-10~\citep{cifar10}                                                                                           \\
		Data augmentation    & random horizontal flip and random $32\times 32$ cropping                                                           \\
		Architecture         & VGG-11~\citep{krizhevsky2009learning}                                                                              \\
		Training objective   & cross entropy                                                                                                      \\
		Evaluation objective & top-1 accuracy                                                                                                     \\
		\midrule
		Number of workers    & 16                                                                                                                 \\
		Topology             & SGP: time-varying exponential, \RelaySumModel: double binary trees, baselines: best of ring or double binary trees \\
		Gossip weights       & Metropolis-Hastings (1/3 for ring)                                                                                 \\
		Data distribution    & Heterogeneous, not shuffled, according to Dirichlet sampling procedure from \cite{lin2021quasiglobal}              \\
		\midrule
		Batch size           & 32 patches per worker                                                                                              \\
		Momentum             & 0.9 (Nesterov)                                                                                                     \\
		Learning rate        & Tuned c.f.~\autoref{apx:hyper:cifar}                                                                               \\
		LR decay             & $/10$ at epoch 150 and 180                                                                                         \\
		LR warmup            & Step-wise linearly within 5 epochs, starting from 0                                                                \\
		\# Epochs            & 200                                                                                                                \\
		Weight decay         & $10^{-4}$                                                                                                          \\
		Normalization scheme & no normalization layer                                                                                             \\
		\midrule
		Repetitions          & 3, with varying seeds                                                                                              \\
		Reported metric      & Worst result of any worker of the worker's mean test accuracy over the last 5 epochs                               \\
		\bottomrule
	\end{tabularx}
\end{table}

\subsection{\imagenet}\label{apx:setup:imagenet}

\autoref{tab:imagenet-experimental-settings}

\begin{table}[h]
	\caption{Default experimental settings for ImageNet}
	\scriptsize%
	\label{tab:imagenet-experimental-settings}%
	\begin{tabularx}{\linewidth}{lX}
		\toprule
		Dataset              & ImageNet~\citep{deng2009imagenet}                                                                                  \\
		Data augmentation    & random resized crop ($224 \times 224$), random horizontal flip                                                     \\
		Architecture         & ResNet-20-EvoNorm~\citep{liu2020evonorm,lin2021quasiglobal}                                                        \\
		Training objective   & cross entropy                                                                                                      \\
		Evaluation objective & top-1 accuracy                                                                                                     \\
		\midrule
		Number of workers    & 16                                                                                                                 \\
		Topology             & SGP: time-varying exponential, \RelaySumModel: double binary trees, baselines: best of ring or double binary trees \\
		Gossip weights       & Metropolis-Hastings (1/3 for ring)                                                                                 \\
		Data distribution    & Heterogeneous, not shuffled, according to Dirichlet sampling procedure from \cite{lin2021quasiglobal}              \\
		\midrule
		Batch size           & 32 patches per worker                                                                                              \\
		Momentum             & 0.9 (Nesterov)                                                                                                     \\
		Learning rate        & based on centralized training (scaled to $0.1 \times \frac{32 * 16}{256}$)                                         \\
		LR decay             & $/10$ at epoch $30, 60, 80$                                                                                        \\
		LR warmup            & Step-wise linearly within 5 epochs, starting from 0.1                                                              \\
		\# Epochs            & 90                                                                                                                 \\
		Weight decay         & $10^{-4}$                                                                                                          \\
		Normalization layer  & EvoNorm~\citep{liu2020evonorm}                                                                                     \\
		\midrule
		Repetitions          & Just one                                                                                                           \\
		Reported metric      & Mean of all worker's test accuracies over the last 5 epochs                                                        \\
		\bottomrule
	\end{tabularx}
\end{table}

\subsection{\bert finetuning}\label{apx:setup:bert}

\autoref{tab:bert-experimental-settings}

\begin{table}[h]
	\caption{Default experimental settings for \bert finetuning}
	\scriptsize%
	\label{tab:bert-experimental-settings}%
	\begin{tabularx}{\linewidth}{lX}
		\toprule
		Dataset              & AG News~\citep{zhang2015character}                                                                    \\
		Data augmentation    & none                                                                                                  \\
		Architecture         & DistilBERT~\citep{sanh2019distilbert}                                                                 \\
		Training objective   & cross entropy                                                                                         \\
		Evaluation objective & top-1 accuracy                                                                                        \\
		\midrule
		Number of workers    & 16                                                                                                    \\
		Topology             & restricted to a ring (chain for \RelaySumModel)                                                       \\
		Gossip weights       & Metropolis-Hastings (1/3 for ring)                                                                    \\
		Data distribution    & Heterogeneous, not shuffled, according to Dirichlet sampling procedure from \cite{lin2021quasiglobal} \\
		\midrule
		Batch size           & 32 patches per worker                                                                                 \\
		Adam $\beta_1$       & 0.9                                                                                                   \\
		Adam $\beta_2$       & 0.999                                                                                                 \\
		Adam $\epsilon$      & $10^{-8}$                                                                                             \\
		Learning rate        & Tuned c.f.~\autoref{apx:hyper:bert}                                                                   \\
		LR decay             & constant learning rate                                                                                \\
		LR warmup            & no warmup                                                                                             \\
		\# Epochs            & 5                                                                                                     \\
		Weight decay         & $0$                                                                                                   \\
		Normalization layer  & LayerNorm~\citep{ba2016layer}                                                                         \\
		\midrule
		Repetitions          & 3, with varying seeds                                                                                 \\
		Reported metric      & Mean of all worker's test accuracies over the last 5 epochs                                           \\
		\bottomrule
	\end{tabularx}
\end{table}

\subsection{Random quadratics}\label{apx:setup:quadratics}

We generate quadratics $\frac{1}{n}\sum_{i=1}^n f_i(\xx)$ of $\xx \in \R^{d}$ where
\begin{align*}
	f_i(\xx) = \norm{ \mA_i \xx + \bb_i }^2_2.
\end{align*}
Here the local Hessian $\mA_i \in \R^{d \times d}$ control the shape of worker $i$'s local objective functions and the offset $\bb_i \in \R^{d}$ allows for shifting the worker's optimum.
The generation procedure is as follows:
\begin{enumerate}
	\item Sample $\mA_i \in \R^{d \times d}$ from an i.i.d.\ element-wise standard normal distribution, independently for each worker.
	\item Control the smoothness $L$ and strong-convexity constant $\mu$.
	      Decompose $\mA_i=\mU_i \mS_i \mV_i^\top$ using Singular Value Decomposition, and replace $\mA_i$ with $\mA_i \gets \mU_i \tilde \mS_i \mV_i^\top$, where $\tilde \mS_i \in \R^{d \times d}$ is a diagonal matrix with diagonal entries $[\mu, \frac{d-2}{d-1}\mu + \frac{1}{d-1}L, \ldots, L]$.
	\item Control the heterogeneity $\zeta_2$ by shifting worker's optima into random directions.
	      \begin{enumerate}
		      \item Sample random directions $\dd_i \in \R^{d}$ from an i.i.d.\ element-wise standard normal distributions, independently for each worker.
		      \item Instantiate a scalar $s \gets 1$ and optimize it using binary search:
		      \item Move local optima by $s\dd_i$ by setting $\bb_i \gets \mA_i s \dd_i$.
		      \item Move all optima $\bb_i \gets \bb_i - \mA_i \xx^\star$ such that the global optimum $\xx^\star$ remains at zero.
		      \item Evaluate $\zeta^2 = \frac{1}{n}\sum_{i=1}^n \norm{\nabla f_i(\xx^\star)}^2_2$ and adjust the scale factor $s$ until $\zeta^2$ is as desired. Repeat from step (c).
	      \end{enumerate}
	\item Control the initial distance to the optimum $r_0$.
	      Sample a random vector for the optimum $\xx^\star$ from an i.i.d.\ element-wise normal distribution and scale it to have norm $r_0$.
	      Shift all worker's optima in this direction by updating $\bb_i \gets \bb_i + \mA_i \xx^\star$.
\end{enumerate}

\section{Hyper-parameters and tuning details}\label{apx:hyperparams}

\subsection{\cifar}\label{apx:hyper:cifar}

For our image classification experiments on \cifar, we have independently tuned learning rates for each algorithm, at each data heterogeneity level $\alpha$, and separately for SGD with and without momentum.
We followed the following procedure:
\begin{enumerate}
	\item We found an appropriate learning rate for centralized (all-reduce) training (by using the procedure below)
	\item Start the search from this learning rate. For \RelaySumModel, we apply a correction computed as in \autoref{apx:alg:learning-rate}.
	\item Grid-search the learning rate by multiplying and dividing by powers of two.
	      Try larger and smaller learning rates, until the best result found so far is sandwiched between two learning rates that gave worse results.
	\item Repeat the experiment with 3 random seeds.
	\item If any of those replicas diverged, reduce the learning rate by a factor two until it does.
\end{enumerate}

For the experiments in \autoref{tab:cifar-results-trees}, we used the learning rates listed in \autoref{tab:apx:cifar-lr}.

\begin{table}[h]
	\caption{
		Learning rates used for \cifar / \vgg.
		Numbers between parentheses indicate the number of converged replications with this learning rate.
		\label{tab:apx:cifar-lr}
	}
	\tablefontsize
\begin{tabularx}{\textwidth}{l X l l l}
    \toprule
    Algorithm & Topology & $\alpha=1.00$ & $\alpha=0.1$ & $\alpha=.01$ \\
    && (most homogeneous) & & (most heterogeneous) \\
    \cmidrule(lr){1-2} \cmidrule(lr){3-5}
All-reduce & fully connected & 0.100 (3) &
0.100 (3) &
0.100 (3) \\
$\quad+$momentum &  & 0.100 (3) &
0.100 (3) &
0.100 (3) \\[1mm]
\textbf{\RelaySumModel}  & binary trees & 1.200 (3) & 
0.600 (3) & 
0.300 (3) \\
\textbf{$\quad+$local momentum}  &  & 0.600 (3) & 
0.300 (3) & 
0.150 (3) \\[1mm]
\dpsgd~\citep{lian2017dpsgd}  & ring & 0.400 (3) & 
0.100 (3) & 
0.200 (3) \\
$\quad+$quasi-global mom.~\citep{lin2021quasiglobal}   & & 0.100 (3) & 
0.025 (3) & 
0.050 (3) \\[1mm]
\dsquare~\citep{tang2018d2}  & ring & 0.200 (3) & 
0.200 (3) & 
0.100 (3) \\
$\quad+$local momentum  & & 0.050 (3) & 
0.050 (3) & 
0.013 (3) \\[1mm]Stochastic gradient push~\citep{assran2019sgp}  & time-varying exponential~\citep{assran2019sgp}     & 0.400 (3) & 
0.200 (3) & 
0.200 (3) \\
$\quad+$local momentum  &     & 0.100 (3) & 
0.100 (3) & 
0.025 (3) \\[1mm]    \bottomrule
\end{tabularx}
\end{table}

\subsection{\imagenet}\label{apx:hyper:imagenet}

Due to the high resource requirements, we did not tune the learning rate for our \imagenet experiments.
We identified a suitable learning rate based on prior work, and used this for all experiments.
For \RelaySumModel, we used the analytically computed learning rate correction from \autoref{apx:alg:learning-rate}.

\subsection{\bert finetuning}\label{apx:hyper:bert}
For DistilBERT fine-tuning experiments on AG News, we have independently tuned learning rate for each algorithm.
We search the learning rate in the grid of $\{ 1e-5, 3e-5, 5e-5, 7e-5, 9e-5 \}$ and we extend the grid to ensure that the best hyper-parameter lies in the middle of our search grids, otherwise we extend our search grid.

For the experiments in \autoref{tab:bert-results}, we used the learning rates listed in \autoref{tab:apx:bert-lr}.
\begin{table}[h]
	\caption{
		Tuned learning rates used for AG News / DistilBERT (\autoref{tab:bert-results})
		\label{tab:apx:bert-lr}
	}
    \centering
    \begin{minipage}{.5\textwidth}
        \tablefontsize
\centering
\begin{tabularx}{\textwidth}{l X l}
	\toprule
	Algorithm                                    & Topology        & Learning rate \\
	\cmidrule(lr){1-2}     \cmidrule(lr){3-3}
	Centralized Adam                             & fully-connected & 3e-5          \\
	\textbf{Relay-Adam}                          & chain           & 9e-4          \\
	\dpsgd Adam                                  & ring            & 1e-6          \\
	Quasi-global Adam~\citep{lin2021quasiglobal} & ring            & 1e-6          \\
	\bottomrule
\end{tabularx}

    \end{minipage}
\end{table}

\subsection{Random quadratics}\label{apx:hyper:quadratics}

For Figures~\ref{fig:effect_of_network_topology_linear} and \ref{fig:social_graph_spanning_tree}, we tuned the learning rate for each compared method to reach a desired quality level as quickly as possible, using binary search.
We made a distinction between methods that are expected to converge linearly, and methods that are expected to reach a plateau.
For experiments with stochastic noise, we tuned a learning rate without noise first, and then lowered the learning rate if needed to reach a desirable plateau.
Please see the supplied code for implementation details.
\section{Algorithmic details}\label{apx:algorithms}

\subsection{Learning-rate correction for \RelaySumModel}\label{apx:alg:learning-rate}

In \dpsgd as well as all other algorithms we compared to, a gradient-based update $\uu^{(t)}_i$ from worker $i$ at time $t$ will eventually, as $t \to \infty$ distribute uniformly with weights $\frac{1}{n}$ over all workers.
In \RelaySumModel, the update also distributes uniformly (typically much quicker), but it will converge to a weight $\alpha \leq \frac{1}{n}$.
The constant $\alpha$ is fixed throughout training and depends only on the network topology used.
To correct for this loss in energy, you can scale the learning rate by a factor $\frac{1}{\alpha n}$.

\begin{figure}[ht]
    \centering
    \includegraphics[width=0.5\textwidth]{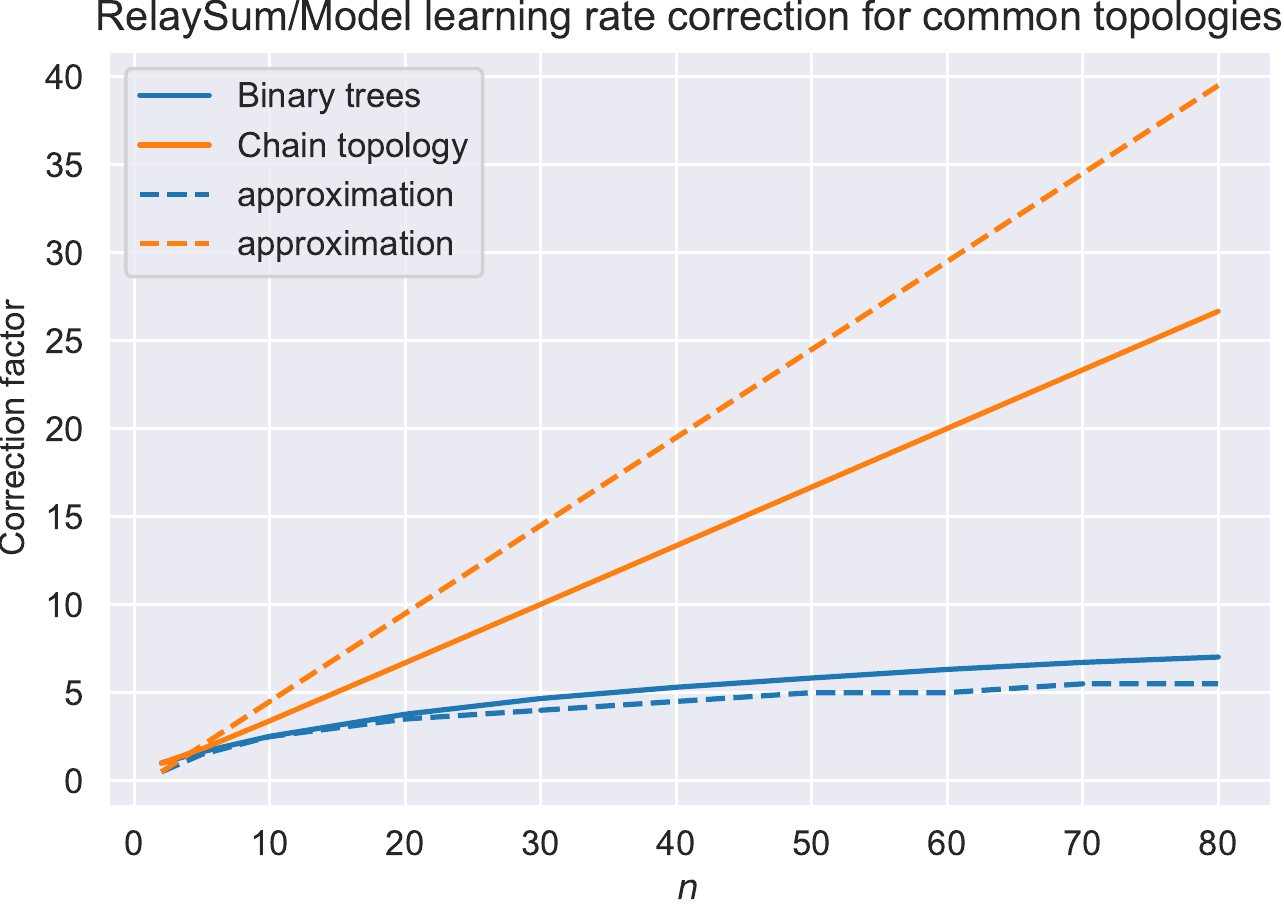}
    \vspace{-4mm}
    \caption{
        \label{fig:relaysum_model_correction_factor}
        This network-topology-dependent correction factor is computed as follows:
        Each worker initializes a scalar model to 0 and sends a single fixed value 1 as gradient update through the \RelaySumModel algorithm.
        For \dpsgd and all-reduce, workers would converge to 1, but for \RelaySumModel, we lose some of this energy.
        If the workers converge to a value $\alpha$, we will scale the learning rate with $1/\alpha$ for \RelaySumModel compared to all-reduce.
    }
\end{figure}

Experimentally, we pre-compute $\alpha$ for each architecture by initialing a \emph{scalar} model for each worker to zero, updating the models to $1$, and running \RelaySumModel until convergence with no further model updates.
The worker will converge to the value $\alpha$.
The correction factors that result from this procedure are illustrated in \autoref{fig:relaysum_model_correction_factor}.

In our deep learning experiments, we find that for each learning rate were centralized SGD converges, \RelaySumModel with the corrected learning rate converges too.
Note that this learning rate correction is only useful if you already have a tuned learning rate from centralized experiments, or experiments with algorithms such as \dpsgd.
If you start from scratch, tuning the learning rate for \RelaySumModel is no different form tuning the learning rate for any of the other algorithms.

\subsection{\RelaySumModel with momentum}\label{apx:alg:relaysum-model-momentum}

\RelaySumModel follows Algorithm~\ref{alg:relaysum_model}, but replaces the local update in line 3 with a local momentum.
For Nesterov momentum with momentum-parameter $\alpha$, this is:
\begin{align*}
    \mm_i^{(t)}       & = \alpha \, \mm_i^{(t-1)} + \nabla f_i(\xx_i^{(t)}) \quad \color{gray}(\text{initialize } \mm_i^{0} = 0) \\
    \xx_i^{(t+\half)} & = \xx_i^{(t)} - \gamma \left(\nabla f_i(\xx_i^{(t)}) + \alpha \, \mm_i^{(t)}\right).
\end{align*}

\subsection{\RelaySumModel with Adam}\label{apx:alg:relaysum-model-adam}

Modifiying \RelaySumModel (Algorithm~\ref{alg:relaysum_model}) to use Adam is analogous to \RelaySumModel with momentum (\autoref{apx:alg:relaysum-model-momentum}).
All Adam state is updated locally.
We use the standard Adam implementation of PyTorch 1.18.

\subsection{\dsquare with momentum}\label{apx:alg:d2-momentum}

We made slight modifications to the \dsquare algorithm from \citet{tang2018d2} to allow time-varying learning rates and local momentum. 
The version we use is listed as Algorithm~\ref{alg:d2_momentum}.
Note that \dsquare requires the smallest eigenvalue of the gossip matrix $\mW$ to be $\geq -1/3$.
This property is satisfied for Metropolis-Hasting matrices used on rings and double binary trees, but it was not in our Social Network Graph experiment (\autoref{fig:social_graph_spanning_tree}).
For this reason, we used the gossip matrix $(\mW + \mI)/2$, from the otherwise-equivalent Exact Diffusion algorithm~\citep{yuan2019exact-diff-1} on the social network graph.

\begin{algorithm}[h]
    \algrenewcommand\algorithmicrequire{\textbf{Input:}}
    \algblockdefx[NAME]{ParallelFor}{ParallelForEnd}[1]{\textbf{for} #1 \textbf{in parallel}}{\textbf{end for}}
    \caption{\dsquare~\citep{tang2018d2} with momentum}\label{alg:d2_momentum}
    \begin{algorithmic}[1] %
        \Require $\forall~i,~\xx_i^{(0)}=\xx^{(0)}$, learning rate $\gamma$, momentum $\alpha$, gossip matrix $\mW \in \R^{n \times n}$, $\cc_i^{(0)} = \mathbf{0} \in \R^{d}$.
        \For{$t=0,1,\ldots$}
        \ParallelFor{node $i$}
        \State Update the local momentum buffer $\mm_i^{(t)} = \alpha \, \mm_i^{(t-1)} + \nabla f_i(\xx_i^{(t)})$.
        \State Compute a local update $\uu_i^{(t)} = - \gamma (\nabla f_i(\xx_i^{(t)}) + \alpha\, \mm_i^{(t)})$.
        \State Update the local model $\xx^{(t+\half)}_i=\xx^{(t)}_i + \uu_i^{(t)} + \cc_i^{(t)}$.
        \State Average with neighbors: $\xx^{(t+1)}_i = \sum_{j \in \mathcal{N}_i} \mW_{ij} \xx_j^{(t+\half)}$.
        \State Update the local correction $\cc^{(t+1)}_i = \xx^{(t+1)}_i - \xx^{(t)}_i - \uu_i^{(t)}$.
        \ParallelForEnd
        \EndFor
    \end{algorithmic}
\end{algorithm}

\subsection{\gradtrack}\label{apx:alg:gradient-tracking}

Algorithm~\ref{alg:gradtrack} lists our implementation of Gradient Tracking from \citet{Lorenzo2016GT-first-paper}.

\begin{algorithm}[h]
    \algrenewcommand\algorithmicrequire{\textbf{Input:}}
    \algblockdefx[NAME]{ParallelFor}{ParallelForEnd}[1]{\textbf{for} #1 \textbf{in parallel}}{\textbf{end for}}
    \caption{\gradtrack~\citep{Lorenzo2016GT-first-paper}}\label{alg:gradtrack}
    \begin{algorithmic}[1] %
        \Require $\forall~i,~\xx_i^{(0)}=\xx^{(0)}$, learning rate $\gamma$, gossip matrix $\mW \in \R^{n \times n}$, $\cc_i^{(0)} = \mathbf{0} \in \R^{d}$.
        \For{$t=0,1,\ldots$}
        \ParallelFor{node $i$}
        \State Compute a local update $\uu_i^{(t)} = - \gamma \nabla f_i(\xx_i^{(t)})$.
        \State Update the local model $\xx^{(t+\half)}_i=\xx^{(t)}_i + \uu_i^{(t)} + \cc_i^{(t)}$.
        \State Average with neighbors: $\xx^{(t+1)}_i = \sum_{j \in \mathcal{N}_i} \mW_{ij} \xx_j^{(t+\half)}$.
        \State Update the correction and average: $\cc^{(t+1)}_i = \sum_{j \in \mathcal{N}_i} \mW_{ij} \left( \cc_i^{(t)} - \uu_i^{(t)} \right)$.
        \ParallelForEnd
        \EndFor
    \end{algorithmic}
\end{algorithm}

\subsection{Stochastic Gradient Push with the time-varying exponential topology}\label{apx:alg:sgp}

Stochastic Gradient Push with the time-varying exponential topology from \cite{assran2019sgp}
demonstrates that decentralized learning algorithms can reduce communication in a data center setting where each node could talk to each other node.
Algorithm~\ref{alg:sgp} lists our implementation of this algorithm.

\begin{algorithm}[h]
    \algrenewcommand\algorithmicrequire{\textbf{Input:}}
    \algblockdefx[NAME]{ParallelFor}{ParallelForEnd}[1]{\textbf{for} #1 \textbf{in parallel}}{\textbf{end for}}
    \caption{Stochastic Gradient Push with time-varying exponential topology~\citep{assran2019sgp}}\label{alg:sgp}
    \begin{algorithmic}[1] %
        \Require $\forall~i,~\xx_i^{(0)}=\xx^{(0)}$, learning rate $\gamma$, $n=2^k$ workers, $t'=0$.
        \For{$t=0,1,\ldots$}
        \ParallelFor{node $i$}
        \State $\xx^{(t+\half)}_i=\xx^{(t)}_i + \uu_i^{(t)} {\color{tab10_blue}- \gamma \nabla f_i(\xx_i^{(t)})}$.\hfill{\color{tab10_blue}(or momentum/Adam, like 
        \RelaySumModel)}
        \For{2 communication steps to equalize bandwidth with \RelaySumModel}
        \State Compute an offset $o$ = $2^{t' \text{ mod } k}$.
        \State Send $\xx^{(t+\half)}_i$ to worker $i - o$.
        \State Receive and overwrite $\xx_i^{(t+\half)} \gets \frac{1}{2}\left(\xx^{(t+\half)}_i + \xx^{(t+\half)}_{i+o} \right)$.
        \State $t' \gets t' + 1$.
        \EndFor
        \State Set $\xx_i^{(t+1)} = \xx_i^{(t+\half)}$.
        \ParallelForEnd
        \EndFor
    \end{algorithmic}
\end{algorithm}
\section{Additional experiments on \RelaySumModel}\label{apx:experiments}

\subsection{Rings vs double binary trees on \cifar}\label{apx:exp:rings-vs-trees}

In our experiments that target data-center inspired scenarios where the network topology is arbitrarily selected by the user to save bandwidth,
 \RelaySumModel uses double binary trees to communicate. 
They use the same memory and bandwidth as rings (2 models sent/received per iteration) but they delays only scale with $\log n$, enabling \RelaySumModel, in theory, to run with very large numbers of workers $n$.
\autoref{tab:ring-vs-tree} shows that in our \cifar experiments with 16 there are minor improvements from using double binary trees over rings.
Our baselines \dpsgd and \dsquare, however, perform significantly better on rings than on trees, so we use those results in the main paper.

\begin{table}[h]
    \caption{
        Comparing the performance of the algorithms in \autoref{tab:cifar-results-trees} on rings and double binary trees
        in the high-heterogeneity setting $\alpha=0.01$.
        In both topologies, workers send and receive two full models per update step.
        With 16 workers, \RelaySumModel with momentum seems to benefit from double binary trees, \RelaySumModel has more consistently good results on a chain.
        We still opt for double binary trees based on their promise to scale to many workers.
        Other methods do not benefit from double binary trees over rings.
        \label{tab:ring-vs-tree}
    }
    \centering
    \tablefontsize
\begin{tabularx}{.7\textwidth}{X l l}
    \toprule
    Algorithm                                              & Ring (Chain for \RelaySumModel) & Double binary trees \\
    \cmidrule(lr){1-1} \cmidrule(lr){2-3}
    \textbf{\RelaySumModel}                                & 86.5\% 
\tikz{
    \draw[white,line width=5pt] (0,0) -- (1.2,0);
    \draw[gray,line width=.3pt,->] (0,0) -- (1.2,0);
    \draw[line width=.6pt] (0.6867277459664789, 0pt) -- (0.7178186611695717, 0pt);
    \draw[line width=.6pt] (0.7178186611695717, -2pt) -- (0.7178186611695717, 2pt);
    \draw[line width=.6pt] (0.6867277459664789, -2pt) -- (0.6867277459664789, 2pt);
    \draw[line width=.6pt] (0.7140004716136235, -2pt) -- (0.7140004716136235, 2pt);
} & 
84.6\% 
\tikz{
    \draw[white,line width=5pt] (0,0) -- (1.2,0);
    \draw[gray,line width=.3pt,->] (0,0) -- (1.2,0);
    \draw[line width=.6pt] (0.2192731553857972, 0pt) -- (0.6840004920959468, 0pt);
    \draw[line width=.6pt] (0.6163640807975418, -2pt) -- (0.6163640807975418, 2pt);
    \draw[line width=.6pt] (0.2192731553857972, -2pt) -- (0.2192731553857972, 2pt);
    \draw[line width=.6pt] (0.6840004920959468, -2pt) -- (0.6840004920959468, 2pt);
} \\
    \textbf{$\quad+$local momentum}                        & 88.4\% 
\tikz{
    \draw[white,line width=5pt] (0,0) -- (1.2,0);
    \draw[gray,line width=.3pt,->] (0,0) -- (1.2,0);
    \draw[line width=.6pt] (0.9021822728893965, 0pt) -- (0.9523641277443294, 0pt);
    \draw[line width=.6pt] (0.9523641277443294, -2pt) -- (0.9523641277443294, 2pt);
    \draw[line width=.6pt] (0.9021822728893965, -2pt) -- (0.9021822728893965, 2pt);
    \draw[line width=.6pt] (0.9032732329585319, -2pt) -- (0.9032732329585319, 2pt);
} & 
89.1\% 
\tikz{
    \draw[white,line width=5pt] (0,0) -- (1.2,0);
    \draw[gray,line width=.3pt,->] (0,0) -- (1.2,0);
    \draw[line width=.6pt] (0.9720005230470139, 0pt) -- (1.0085459094155904, 0pt);
    \draw[line width=.6pt] (1.0085459094155904, -2pt) -- (1.0085459094155904, 2pt);
    \draw[line width=.6pt] (0.9976368411020795, -2pt) -- (0.9976368411020795, 2pt);
    \draw[line width=.6pt] (0.9720005230470139, -2pt) -- (0.9720005230470139, 2pt);
} \\[1mm]
    \dpsgd~\citep{lian2017dpsgd}                           & 53.9\% 
\tikz{
    \draw[white,line width=5pt] (0,0) -- (1.2,0);
    \draw[gray,line width=.3pt,->] (0,0) -- (1.2,0);
    \draw[line width=.6pt] (0.0, 0pt) -- (0.0, 0pt);
} & 
36.0\% 
\tikz{
    \draw[white,line width=5pt] (0,0) -- (1.2,0);
    \draw[gray,line width=.3pt,->] (0,0) -- (1.2,0);
    \draw[line width=.6pt] (0.0, 0pt) -- (0.0, 0pt);
} \\
    $\quad+$quasi-global mom.~\citep{lin2021quasiglobal}   & 63.3\% 
\tikz{
    \draw[white,line width=5pt] (0,0) -- (1.2,0);
    \draw[gray,line width=.3pt,->] (0,0) -- (1.2,0);
    \draw[line width=.6pt] (0.0, 0pt) -- (0.0, 0pt);
} & 
57.5\% 
\tikz{
    \draw[white,line width=5pt] (0,0) -- (1.2,0);
    \draw[gray,line width=.3pt,->] (0,0) -- (1.2,0);
    \draw[line width=.6pt] (0.0, 0pt) -- (0.0, 0pt);
} \\[1mm]
    \dsquare~\citep{tang2018d2}                            & 38.2\% 
\tikz{
    \draw[white,line width=5pt] (0,0) -- (1.2,0);
    \draw[gray,line width=.3pt,->] (0,0) -- (1.2,0);
    \draw[line width=.6pt] (0.0, 0pt) -- (0.0, 0pt);
} & 
{\color{gray}did not converge} \\
    $\quad+$local momentum                                 & 61.0\% 
\tikz{
    \draw[white,line width=5pt] (0,0) -- (1.2,0);
    \draw[gray,line width=.3pt,->] (0,0) -- (1.2,0);
    \draw[line width=.6pt] (0.0, 0pt) -- (0.7047277174212707, 0pt);
    \draw[line width=.6pt] (0.7025459192015889, -2pt) -- (0.7025459192015889, 2pt);
    \draw[line width=.6pt] (0.7047277174212707, -2pt) -- (0.7047277174212707, 2pt);
} & 
{\color{gray}did not converge} \\[1mm]
    \bottomrule
\end{tabularx}
\end{table}

\subsection{Scaling the number of workers on Cifar-10}\label{apx:exp:scaling}

In this experiment (\autoref{tab:scaling}), use momentum-SGD on 16, 32 and 64 workers compare the scaling of \RelaySumModel to SGP~\citep{assran2019sgp}. We fix the parameter $\alpha$ that determines the level of data heterogeneity to $\alpha=0.01$. Note that this level of $\alpha$ could lead to more challenging heterogeneity when there are many workers (and hence many smaller local subsets of the data), compared to when there are few workers.

\begin{table}[h]
    \caption{
        Scaling the number of workers in heterogeneous Cifar-10. The heterogeneity level $\alpha=0.01$ is kept constant, although it does change its meaning when the number of workers changes. \RelaySumModel scales at least well as Stochastic Gradient Push~\citep{assran2019sgp} (with equal communication budget). It is surprising that \RelaySumModel with 64 workers performs significantly better on a chain topology than on the double binary trees. This behavior does not match what our observations on quadratic toy-problems.
        \label{tab:scaling}
    }
    \centering
    \tablefontsize
\begin{tabularx}{\textwidth}{l X l l l}
    \toprule
    Algorithm & Topology & 16 workers & 32 workers & 64 workers \\
    \cmidrule(lr){1-2} \cmidrule(lr){3-5}
All-reduce {\color{gray}(baseline)} & fully connected & 89.5\% 
\tikz{
    \draw[white,line width=5pt] (0,0) -- (1.2,0);
    \draw[gray,line width=.3pt,->] (0,0) -- (1.2,0);
    \draw[line width=.6pt] (1.010182304815812, 0pt) -- (1.0620004859837617, 0pt);
    \draw[line width=.6pt] (1.0620004859837617, -2pt) -- (1.0620004859837617, 2pt);
    \draw[line width=.6pt] (1.010182304815812, -2pt) -- (1.010182304815812, 2pt);
    \draw[line width=.6pt] (1.051636812361804, -2pt) -- (1.051636812361804, 2pt);
} &
88.9\% 
\tikz{
    \draw[white,line width=5pt] (0,0) -- (1.2,0);
    \draw[gray,line width=.3pt,->] (0,0) -- (1.2,0);
    \draw[line width=.6pt] (1.0614288364137925, 0pt) -- (1.1034288065774094, 0pt);
    \draw[line width=.6pt] (1.1034288065774094, -2pt) -- (1.1034288065774094, 2pt);
    \draw[line width=.6pt] (1.0788573878152037, -2pt) -- (1.0788573878152037, 2pt);
    \draw[line width=.6pt] (1.0614288364137925, -2pt) -- (1.0614288364137925, 2pt);
} &
87.2\% 
\tikz{
    \draw[white,line width=5pt] (0,0) -- (1.2,0);
    \draw[gray,line width=.3pt,->] (0,0) -- (1.2,0);
    \draw[line width=.6pt] (1.0759025652234144, 0pt) -- (1.1013659747635447, 0pt);
    \draw[line width=.6pt] (1.0759025652234144, -2pt) -- (1.0759025652234144, 2pt);
    \draw[line width=.6pt] (1.1013659747635447, -2pt) -- (1.1013659747635447, 2pt);
} \\[1mm]
\RelaySumModel  & binary trees & 89.3\% 
\tikz{
    \draw[white,line width=5pt] (0,0) -- (1.2,0);
    \draw[gray,line width=.3pt,->] (0,0) -- (1.2,0);
    \draw[line width=.6pt] (0.9970913610675124, 0pt) -- (1.0390914217992269, 0pt);
    \draw[line width=.6pt] (1.0390914217992269, -2pt) -- (1.0390914217992269, 2pt);
    \draw[line width=.6pt] (0.9970913610675124, -2pt) -- (0.9970913610675124, 2pt);
} & 
86.1\% 
\tikz{
    \draw[white,line width=5pt] (0,0) -- (1.2,0);
    \draw[gray,line width=.3pt,->] (0,0) -- (1.2,0);
    \draw[line width=.6pt] (0.9154288130147115, 0pt) -- (0.9240002632141114, 0pt);
    \draw[line width=.6pt] (0.9240002632141114, -2pt) -- (0.9240002632141114, 2pt);
    \draw[line width=.6pt] (0.9217145272663665, -2pt) -- (0.9217145272663665, 2pt);
    \draw[line width=.6pt] (0.9154288130147115, -2pt) -- (0.9154288130147115, 2pt);
} & 
63.7\% 
\tikz{
    \draw[white,line width=5pt] (0,0) -- (1.2,0);
    \draw[gray,line width=.3pt,->] (0,0) -- (1.2,0);
    \draw[line width=.6pt] (0.23268299713367385, 0pt) -- (0.5439513100356591, 0pt);
    \draw[line width=.6pt] (0.5439513100356591, -2pt) -- (0.5439513100356591, 2pt);
    \draw[line width=.6pt] (0.23268299713367385, -2pt) -- (0.23268299713367385, 2pt);
    \draw[line width=.6pt] (0.4264391035568425, -2pt) -- (0.4264391035568425, 2pt);
} \\
  & chain & 88.4\% 
\tikz{
    \draw[white,line width=5pt] (0,0) -- (1.2,0);
    \draw[gray,line width=.3pt,->] (0,0) -- (1.2,0);
    \draw[line width=.6pt] (0.8514549623836181, 0pt) -- (0.9758186638355246, 0pt);
    \draw[line width=.6pt] (0.9758186638355246, -2pt) -- (0.9758186638355246, 2pt);
    \draw[line width=.6pt] (0.8514549623836181, -2pt) -- (0.8514549623836181, 2pt);
    \draw[line width=.6pt] (0.9065459262241006, -2pt) -- (0.9065459262241006, 2pt);
} & 
86.6\% 
\tikz{
    \draw[white,line width=5pt] (0,0) -- (1.2,0);
    \draw[gray,line width=.3pt,->] (0,0) -- (1.2,0);
    \draw[line width=.6pt] (0.8788573827062333, 0pt) -- (0.9920002647808615, 0pt);
    \draw[line width=.6pt] (0.8788573827062333, -2pt) -- (0.8788573827062333, 2pt);
    \draw[line width=.6pt] (0.9731430922235755, -2pt) -- (0.9731430922235755, 2pt);
    \draw[line width=.6pt] (0.9920002647808615, -2pt) -- (0.9920002647808615, 2pt);
} & 
83.1\% 
\tikz{
    \draw[white,line width=5pt] (0,0) -- (1.2,0);
    \draw[gray,line width=.3pt,->] (0,0) -- (1.2,0);
    \draw[line width=.6pt] (0.9108293783373945, 0pt) -- (1.0085854951928295, 0pt);
    \draw[line width=.6pt] (0.9878049870816672, -2pt) -- (0.9878049870816672, 2pt);
    \draw[line width=.6pt] (1.0085854951928295, -2pt) -- (1.0085854951928295, 2pt);
    \draw[line width=.6pt] (0.9108293783373945, -2pt) -- (0.9108293783373945, 2pt);
} \\[1mm]
Stochastic gradient push~\citep{assran2019sgp}  & time-varying exponential~\citep{assran2019sgp}     & 87.0\% 
\tikz{
    \draw[white,line width=5pt] (0,0) -- (1.2,0);
    \draw[gray,line width=.3pt,->] (0,0) -- (1.2,0);
    \draw[line width=.6pt] (0.5522731921889557, 0pt) -- (0.8858186765150587, 0pt);
    \draw[line width=.6pt] (0.5522731921889557, -2pt) -- (0.5522731921889557, 2pt);
    \draw[line width=.6pt] (0.8667277287353164, -2pt) -- (0.8667277287353164, 2pt);
    \draw[line width=.6pt] (0.8858186765150587, -2pt) -- (0.8858186765150587, 2pt);
} & 
68.9\% 
\tikz{
    \draw[white,line width=5pt] (0,0) -- (1.2,0);
    \draw[gray,line width=.3pt,->] (0,0) -- (1.2,0);
    \draw[line width=.6pt] (0.0, 0pt) -- (0.611428788730078, 0pt);
    \draw[line width=.6pt] (0.611428788730078, -2pt) -- (0.611428788730078, 2pt);
} & 
62.4\% 
\tikz{
    \draw[white,line width=5pt] (0,0) -- (1.2,0);
    \draw[gray,line width=.3pt,->] (0,0) -- (1.2,0);
    \draw[line width=.6pt] (0.2982439834897108, 0pt) -- (0.4281952264832288, 0pt);
    \draw[line width=.6pt] (0.2982439834897108, -2pt) -- (0.2982439834897108, 2pt);
    \draw[line width=.6pt] (0.4281952264832288, -2pt) -- (0.4281952264832288, 2pt);
} \\
    \bottomrule
\end{tabularx}
\end{table}

\begin{table}[h]
    \caption{
        Tuned learning rates for \autoref{tab:scaling}.
        We tuned the learning rate for each setting on a multiplicative grid with spacing $\sqrt{2}$, and then repeated each experiment 3 times. If both repetitions diverged, we would change to a smaller learning rate in the grid.
        Numbers in parentheses are the `effective' learning rates corrected according to \autoref{apx:alg:learning-rate}.
        \label{tab:scaling-learning-rates}
    }
    \centering
    \tablefontsize
\begin{tabularx}{\textwidth}{l X l l l l l l}
    \toprule
    Algorithm & Topology & \multicolumn{2}{X}{16 workers} & \multicolumn{2}{X}{32 workers} & \multicolumn{2}{X}{64 workers} \\
    \cmidrule(lr){1-2} \cmidrule(lr){3-4} \cmidrule(lr){5-6} \cmidrule(lr){7-8}
All-reduce {\color{gray}(baseline)} & fully connected & 0.1 & {\color{gray}(0.100)} &
0.05 & {\color{gray}(0.050)} &
0.05 & {\color{gray}(0.050)} \\[1mm]
\RelaySumModel  & binary trees & 0.282 & {\color{gray}(0.066)} & 
0.2 & {\color{gray}(0.035)} & 
0.2 & {\color{gray}(0.027)} \\
  & chain & 0.2 & {\color{gray}(0.047)} & 
0.4 & {\color{gray}(0.070)} & 
0.8 & {\color{gray}(0.108)} \\[1mm]
Stochastic gradient push~\citep{assran2019sgp}  & time-varying exp.\     & 0.025 & {\color{gray}(0.025)} & 
0.025 & {\color{gray}(0.025)} & 
0.0125 & {\color{gray}(0.013)} \\
    \bottomrule
\end{tabularx}
\end{table}

\subsection{Independence of heterogeneity}\label{apx:exp:heterogeneity}

\begin{figure}[ht]
    \includegraphics[width=\textwidth]{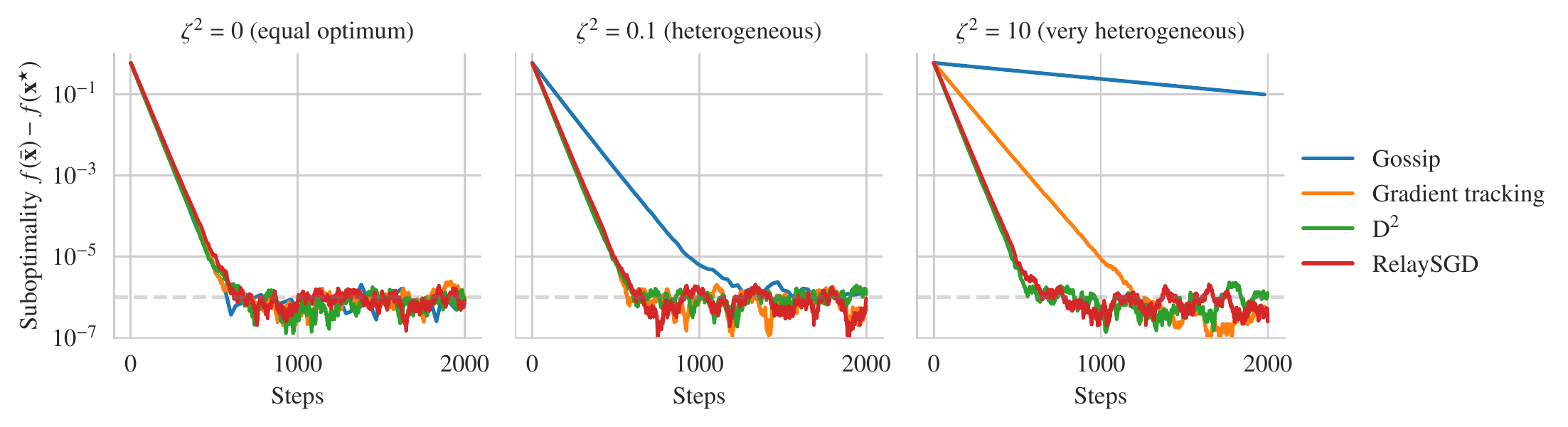}
    \vspace{-4mm}
    \caption{
        \label{fig:effect_of_heterogeneity_fixed_saturation}
        Random quadratics on \emph{ring} networks of size 32 with varying data heterogeneity $\zeta^2$ and all other theoretical quantities fixed. To simulate stochastic noise, we add random normal noise to each gradient update.
        For each method, the learning rate is tuned to reach suboptimality $\leq 10^{-6}$ the fastest.
        \RelaySumModel operates on a chain network instead of a ring. Like \dsquare, it does not require more steps when the worker's objectives are more heterogeneous.
    }
\end{figure}

The benefits of \RelaySumModel over some other methods shows most when workers have heterogeneous training objectives.
\autoref{fig:effect_of_heterogeneity_fixed_saturation} compares several algorithms with varying levels of data heterogeneity on synthetic quadratics on a ring topology with 32 workers.
Like \dsquare, \RelaySumModel converges linearly, and does not require more steps when the data becomes more heterogeneous.
Note that, even though \RelaySumModel operates on a chain network instead of a ring, it is as fast as \dsquare.
On other topologies, such as a star topology, or on trees, \RelaySumModel can even be faster than \dsquare (see Appendix~\ref{apx:exp:star}), while maintaining the same independence of heterogeneity.

\subsection{Star topology}\label{apx:exp:star}

\begin{figure}[h]
    \includegraphics[width=\textwidth]{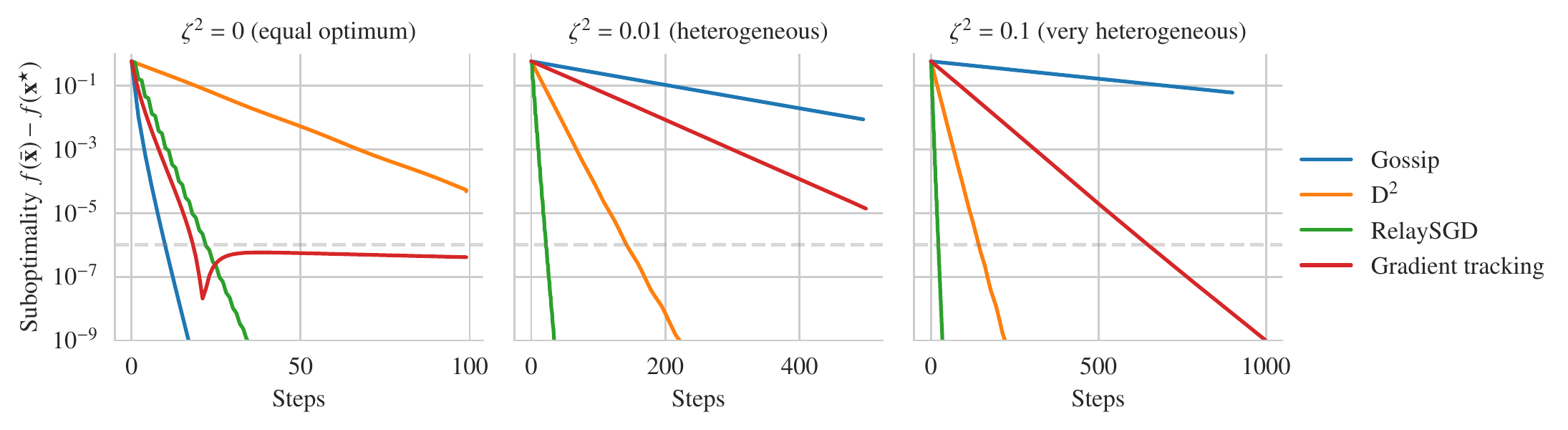}
    \vspace{-4mm}
    \caption{
        \label{fig:effect_of_heterogeneity_fixed_saturation_star}
        Random quadratics on \emph{star} networks of size 32 with varying data heterogeneity $\zeta^2$ and all other theoretical quantities fixed.
        For each method, the learning rate is tuned to reach suboptimality $\leq 10^{-6}$ the fastest.
        Like \dsquare, \RelaySumModel does not require more steps when the worker's objectives are more heterogeneous.
        Note that for $\zeta^2=0$ (left figure), our tuning procedure found a learning rate where \gradtrack does converge to $<\leq 10^{-6}$, but does not converge linearly. It would with a lower learning rate.
    }
\end{figure}

On star-topologies, the set of neighbors of worker 0 is $\{1, 2, \ldots, n\}$ and the set of neighbors for every other worker is just $\{0\}$.
While \dsquare and \RelaySumModel are equally fast in the synthetic experiments on \emph{ring} topologies in \autoref{apx:exp:heterogeneity}, \RelaySumModel is significantly faster on \emph{star} topologies as illustrates by \autoref{fig:effect_of_heterogeneity_fixed_saturation_star}.

\section{\RelaySum for distributed mean estimation}\label{apx:distributed-mean-estimation}

We conceptually separate the optimization algorithm \RelaySumModel from the communication mechanism \RelaySum that uniformly distributes updates across a peer-to-peer network.
We made this choice because we envision other applications of the \RelaySum mechanism outside of optimization for machine learning.
To illustrate this point, this section introduces \RelaySum for Distributed Mean Estimation (Algorithm~\ref{alg:relaysum_dme}).

In distributed mean estimation, workers are connected in a network just as in our optimization setup, but instead of models gradients, they receive samples $\hat \dd^{(t)} \sim \cD$ of the distribution $\cD$ at timestep $t$. 
The workers estimate the mean $\bar \dd$ the mean of $\cD$, and we measure their average squared error to the true mean.

\begin{algorithm}[h]
    \algrenewcommand\algorithmicrequire{\textbf{Input:}}
    \algblockdefx[NAME]{ParallelFor}{ParallelForEnd}[1]{\textbf{for} #1 \textbf{in parallel}}{\textbf{end for}}    \caption{\RelaySum for Distributed Mean Estimation}\label{alg:relaysum_dme}
    \begin{algorithmic}[1] %
        \Require $\forall~i,~\xx_i^{(0)}=\mathbf{0}, \yy_i^{(0)}=\mathbf{0}, s_i^{(0)}=0$; $\forall~i,j, \mm^{(-1)}_{i\to j}=\0$, tree network
        \For{$t=0,1,\ldots$}
        \ParallelFor{node $i$}
        \For{each neighbor $j\in \cN_i$}
        \State Get a sample $\hat \dd_i^{(t)} \sim \cD$.
        \State Send $\mm^{(t)}_{i\to j}=\hat \dd_i^{(t)} + \sum_{k\in\cN_i\backslash j}\mm^{(t-1)}_{k\to i}$.
        \State Send $c^{(t)}_{i\to j}=1 + \sum_{k\in\cN_i\backslash j}c^{(t-1)}_{k\to i}$.
        \State Receive $\mm^{(t)}_{j \to i}$ and  $c^{(t)}_{j \to i}$ from node $j$.
        \EndFor
        \State Update the sum of samples $\yy_i^{(t+1)}= \yy_i^{(t)} +  \hat \dd_i^{(t)} + \sum_{j \in \cN_i} \mm_{j \to i}^{(t)}$.
        \State Update the sum of counts $s_i^{(t+1)}= s_i^{(t)} +  1 + \sum_{j \in \cN_i} c_{j \to i}^{(t)}$.
        \State Output average estimate $\xx_i^{(t)} = \yy_i^{(t)} / s_i^{(t)}$
        \ParallelForEnd
        \EndFor
    \end{algorithmic}
\end{algorithm}

In algorithm~\ref{alg:relaysum_dme}, the output estimates $\xx_i^{(t)}$ of a worker $i$ is a uniform average of all samples that can reach a worker $i$ at that timestep.
This algorithm enjoys variance reduction of $\cO\left(\frac{1}{nT}\right)$, a desirable property that is in general not shared by gossip-averaging-based algorithms on arbitrary graphs.

\begin{figure}
    \centering
    \includegraphics[width=\textwidth]{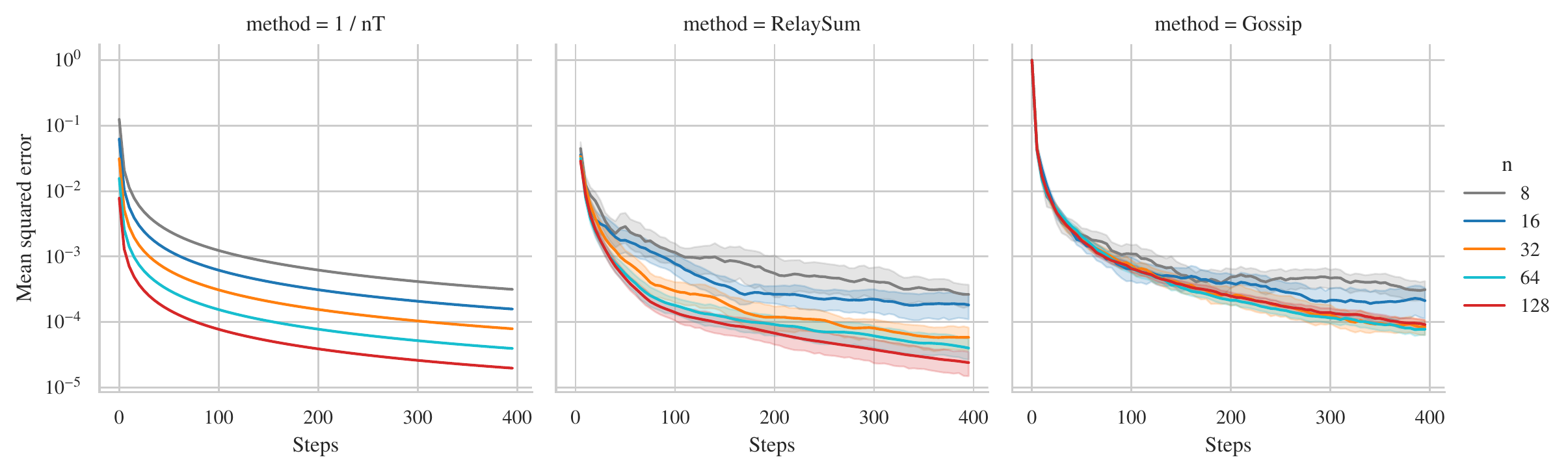}
    \vspace{-7mm}
    \caption{
        \label{fig:relaysum-dme-results}
        \RelaySum for Distributed Mean Estimation compured to a gossip-based baseline, on a ring topology (chain for \RelaySumModel).
        Workers receive samples from a normal distribution $\cN(1, 1)$ with mean 1.
        \RelaySum, using Algorithm~\ref{alg:relaysum_dme} achieves a variance reduction of $\cO\left(\frac{1}{nT}\right)$.
    }
\end{figure}

In \autoref{fig:relaysum-dme-results}, we compare this algorithm to a simple gossip-based baseline.

\section{Alternative optimizer based on \RelaySum}\label{apx:relaysum-grad}

Apart from \RelaySumModel presented in the main paper, there are other ways to build optimization algorithms based on the \RelaySum communication mechanism.
In this section, we describe \RelaySumGrad (Algorithm~\ref{alg:relaysum_grad}), an alternative to \RelaySumModel that does uses the \RelaySum mechanism on \emph{gradient updates} rather than on \emph{models}.

\RelaySumGrad distributes each update uniformly over all workers in a finite number of steps.
This means that worker's models differ by only a finite number of $\cO({\tau_{\max}}max n)$ that are scaled as $\frac{1}{n}$.
With this property, it achieves tighter consensus than typical gossip averaging, and it also works well in deep learning.
Contrary to \RelaySumModel, however, this algorithm is not fully independent of data heterogeneity, due to the delay in the updates.
When the data heterogeneity $\zeta^2 > 0$, \RelaySumGrad does not converge linearly, but its suboptimality saturates at a level that depends on $\zeta^2$.

The sections below study this alternative algorithm in detail, both theoretically and experimentally.
The key differences between \RelaySumModel and \RelaySumGrad are:

\begin{tabularx}{\textwidth}{X c c}
    \toprule
                                                                       & \RelaySumModel & \RelaySumGrad \\
    \cmidrule(lr){2-2} \cmidrule(lr){3-3}
    Provably independent of data heterogeneity $\zeta^2$               & yes            & no            \\
    Distributes updates exactly uniform in finite steps                & no             & yes           \\
    Loses energy of gradient updates (\autoref{apx:alg:learning-rate}) & yes            & no            \\[1mm]
    Works experimentally with momentum / Adam                          & yes            & no            \\
    Robust to lost messages + can support workers joining/leaving      & yes            & no            \\
    \bottomrule
\end{tabularx}

\begin{algorithm}[h]
    \algrenewcommand\algorithmicrequire{\textbf{Input:}}
    \algblockdefx[NAME]{ParallelFor}{ParallelForEnd}[1]{\textbf{for} #1 \textbf{in parallel}}{\textbf{end for}}    \caption{\RelaySumGrad}\label{alg:relaysum_grad}
    \begin{algorithmic}[1] %
        \Require $\forall~i,~\xx_i^{(0)}=\xx^{(0)}$; $\forall~i,j, \mm^{(-1)}_{i\to j}=\0$, learning rate $\gamma$, tree network
        \For{$t=0,1,\ldots$}
        \ParallelFor{node $i$}
        \State $\uu_i^{(t)} = -\gamma \nabla f_i(\xx_i^{(t)}, \xi_i^{(t)})$
        \For{each neighbor $j\in \cN_i$}
        \State Send $\mm^{(t)}_{i\to j}=\uu_i^{(t)} + \sum_{k\in\cN_i\backslash j}\mm^{(t-1)}_{k\to i}$.
        \State Receive $\mm^{(t)}_{j \to i}$ from node $j$.
        \EndFor
        \State $\xx_i^{(t+1)}= \xx_i^{(t)} + \frac{1}{n} \left( \uu_i^{(t)} + \sum_{j \in \cN_i} \mm_{j \to i}^{(t)} \right)$
        \ParallelForEnd
        \EndFor
    \end{algorithmic}
\end{algorithm}

\subsection{Theoretical analysis of \RelaySumGrad}\label{apx:relaysum-grad:theory}
In this section we provide the theoretical analysis for \RelaySumGrad. As the proof and analysis is very similar to \cite{koloskova2021unified}, we only provide the case for the convex objective.
\subsubsection{Proof of \RelaySumGrad for the convex case}
Let $\xx^\star$ be the minimizer of $f$ and define the following iterates
\begin{itemize}
    \item $r_t:=\E\|\bar{\xx}^{(t)} - \xx^\star \|^2$,
    \item $ e_t:= f(\bar{\xx}^{(t)}) - f(\xx^\star)$,
    \item $\Xi_t:=\frac{1}{n}\sum_{i=1}^n\| \xx_i^{(t)} - \bar{\xx}^{(t)} \|^2 $.
\end{itemize}

\begin{proposition}\label{prop:bounded_noise}
    Let function $F_i(\xx, \xi)$, $i\in[n]$ be $L$-smooth (\Cref{assumption:smoothness:stochastic}) with bounded noise at the optimum (\Cref{assumption:zeta2:convex}). Then for any $\xx_i\in\R^d$,
    \begin{align*}
        \E_{\xi_1^t,\ldots,\xi_n^t} \left\| \frac{1}{n} \sum_{i=1}^n(\nabla f_i(\xx_i^{(t)}) - \nabla F_i(\xx_i^{(t)}, \xi_i^{(t)})) \right\|^2
        \leq
        \tfrac{3}{n} (L^2 \Xi_t + 2L e_t + \bar{\sigma}^2)
    \end{align*}
\end{proposition}
\begin{proof}
    In this proof we ignore the superscript $t$ as it does not raise embiguity.
    \begin{align*}
             & \E_{\xi_1,\ldots,\xi_n} \left\| \frac{1}{n} \sum_{i=1}^n(\nabla f_i(\xx_i) - \nabla F_i(\xx_i, \xi_i)) \right\|^2  \leq \frac{1}{n^2}\sum_{i=1}^n\E_{\xi_i}\|\nabla f_i(\xx_i) - \nabla F_i(\xx_i, \xi_i) \|^2          \\
        =    & \frac{1}{n^2}\sum_{i=1}^n\E_{\xi_i} \left\| \nabla f_i(\xx_i) - \nabla F_i(\xx_i, \xi_i)
        \pm \nabla F_i(\bar{\xx}, \xi_i) \pm \nabla f_i(\bar{\xx}) \pm \nabla F_i(\xx^\star, \xi_i) \pm \nabla f_i(\xx^\star) \right\|^2                                                                                               \\
        \leq & \frac{3}{n^2}\sum_{i=1}^n\E_{\xi_i} \left(
        \| \nabla f_i(\xx_i) -\nabla f_i(\bar{\xx}) + \nabla F_i(\bar{\xx}, \xi_i) - \nabla F_i(\xx_i, \xi_i)  \|^2 \right.                                                                                                            \\
             & \qquad + \|\nabla f_i(\bar{\xx}) - \nabla f_i(\xx^\star) + \nabla F_i(\xx^\star, \xi_i) -  \nabla F_i(\bar{\xx}, \xi_i) \|^2                      + \|\nabla f_i(\xx^\star) -  \nabla F_i(\xx^\star, \xx_i) \|^2      ) \\
        \leq &
        \frac{3}{n^2}\sum_{i=1}^n\E_{\xi_i} (
        \|  \nabla F_i(\xx_i, \xi_i) -\nabla F_i(\bar{\xx}, \xi_i)  \|^2
        + \|\nabla F_i(\bar{\xx}, \xi_i)  - \nabla F_i(\xx^\star, \xi_i)\|^2
        + \|\nabla F_i(\xx^\star, \xx_i) - \nabla f_i(\xx^\star) \|^2)                                                                                                                                                                 \\
        \leq &
        \frac{3}{n^2}\sum_{i=1}^n (
        L^2\|  \xx_i-\bar{\xx} \|^2
        + 2L (f_i(\bar{\xx})  - f_i(\xx^\star)) + \sigma_i^2 )
    \end{align*}
\end{proof}

\begin{lemma}\label{lemma:convex:descent}(Descent lemma for convex objective.)
    If $\gamma \le \frac{1}{10L}$, then
    \begin{align*}
        r_{t+1} \le (1-\tfrac{\gamma\mu}{2}) r_t - \gamma e_t + 3\gamma L\Xi_t+\tfrac{3}{n}\gamma^2\bar{\sigma}^2.
    \end{align*}
\end{lemma}
\begin{proof}
    Throughout this proof we use $\E=\E_{\xi_1^t,\ldots,\xi_n^t}$.
    Expand iterate $r_{t+1}=\E\|\bar{\xx}^{(t+1)} - \xx^\star \|^2$
    \begin{align*}
          & \E\|\bar{\xx}^{(t+1)} - \xx^\star \|^2                                                                                                                                                                                                  \\
        = & \E\|\bar{\xx}^{(t)} - \tfrac{\gamma}{n}\tsum_{i=1}^n \nabla F_i(\xx_i^{(t)}, \xi_i^{(t)}) \pm \tfrac{\gamma}{n}\tsum_{i=1}^n \nabla f_i(\xx_i^{(t)})  - \xx^\star \|^2                                                                  \\
        = & \|\bar{\xx}^{(t)} - \xx^\star  - \tfrac{\gamma}{n}\tsum_{i=1}^n \nabla f_i(\xx_i^{(t)}) \|^2  + \E\|\tfrac{\gamma}{n}\tsum_{i=1}^n \nabla F_i(\xx_i^{(t)}, \xi_i^{(t)}) - \tfrac{\gamma}{n}\tsum_{i=1}^n \nabla f_i(\xx_i^{(t)})  \|^2  \\
          & + 2\E\langle\bar{\xx}^{(t)} - \xx^\star  - \tfrac{\gamma}{n}\tsum_{i=1}^n \nabla f_i(\xx_i^{(t)}), \tfrac{\gamma}{n}\tsum_{i=1}^n \nabla F_i(\xx_i^{(t)}, \xi_i^{(t)}) - \tfrac{\gamma}{n}\tsum_{i=1}^n \nabla f_i(\xx_i^{(t)}) \rangle \\
        = & \|\bar{\xx}^{(t)} - \xx^\star  - \tfrac{\gamma}{n}\tsum_{i=1}^n \nabla f_i(\xx_i^{(t)}) \|^2  + \E\|\tfrac{\gamma}{n}\tsum_{i=1}^n \nabla F_i(\xx_i^{(t)}, \xi_i^{(t)}) - \tfrac{\gamma}{n}\tsum_{i=1}^n \nabla f_i(\xx_i^{(t)})  \|^2
    \end{align*}
    The second term is bounded by \Cref{prop:bounded_noise}.
    Consider the first term
    \begin{align*}
            & \|\bar{\xx}^{(t)} - \xx^\star  - \tfrac{\gamma}{n}\tsum_{i=1}^n \nabla f_i(\xx_i^{(t)}) \|^2                                   \\
        \le & \| \bar{\xx}^{(t)} - \xx^\star \|^2 + \gamma^2 \underbrace{ \| \tfrac{1}{n}\tsum_{i=1}^n \nabla f_i(\xx_i^{(t)}) \|^2}_{=:T_1}
        - 2\gamma\underbrace{\langle\bar{\xx}_{t} - \xx^\star, \tfrac{1}{n}\tsum_{i=1}^n \nabla f_i(\xx_i^{(t)})\rangle}_{=:T_2}.
    \end{align*}
    First consider $T_1$,
    \begin{align*}
        T_1 & =\| \tfrac{1}{n}\tsum_{i=1}^n (\nabla f_i(\xx_i^{(t)}) - \nabla f_i(\bar{\xx}^{(t)}) + \nabla f_i(\bar{\xx}^{(t)}) - \nabla f_i(\xx^\star) ) \|^2 \\
            & \le \tfrac{2L^2}{n} \tsum_{i=1}^n\| \xx_i^{(t)} - \bar{\xx}^{(t)} \|^2
        + \tfrac{2}{n} \sum_{i=1}^n \| \nabla f_i(\bar{\xx}^{(t)}) - \nabla f_i(\xx^\star) \|^2                                                                 \\
            & \stackrel{\eqref{eq:smooth:optimum}}{\le}
        \tfrac{2L^2}{n} \tsum_{i=1}^n\| \xx_i^{(t)} - \bar{\xx}^{(t)} \|^2
        +\tfrac{4L}{n} \sum_{i=1}^n (f_i(\bar{\xx}^{(t)}) - f_i(\xx^\star) - \langle\bar{\xx}^{(t)}-\xx^\star, \nabla f_i(\xx^\star) \rangle )                  \\
            & = \tfrac{2L^2}{n} \tsum_{i=1}^n\| \xx_i^{(t)} - \bar{\xx}^{(t)} \|^2
        + 4L (f(\bar{\xx}^{(t)}) -f(\xx^\star))                                                                                                                 \\
            & = 2L^2 \Xi_t + 4L  e_t.
    \end{align*}
    Consider $T_2$,
    \begin{align*}
        T_2 & = \tfrac{1}{n}\tsum_{i=1}^n (\langle\bar{\xx}^{(t)} - \xx_i^{(t)},  \nabla f_i(\xx_i^{(t)})\rangle
        + \langle\xx_i^{(t)} - \xx^\star,  \nabla f_i(\xx_i^{(t)})\rangle)                                                     \\
            & \ge \tfrac{1}{n}\tsum_{i=1}^n
        \left(
        f_i(\bar{\xx}^{(t)}) - f_i(\xx_i^{(t)}) - \tfrac{L}{2} \| \bar{\xx}^{(t)} - \xx_i^{(t)} \|^2
        + \langle\xx_i^{(t)} - \xx^\star,  \nabla f_i(\xx_i^{(t)})\rangle\right)
        \\
            & \ge \tfrac{1}{n}\tsum_{i=1}^n
        \left(
        f_i(\bar{\xx}^{(t)}) - f_i(\xx_i^{(t)}) - \tfrac{L}{2} \| \bar{\xx}^{(t)} - \xx_i^{(t)} \|^2
        + f_i(\xx_i^{(t)}) - f_i(\xx^\star) + \tfrac{\mu}{2} \| \xx_i^{(t)} -\xx^\star \|^2\right)                             \\
            & = f(\bar{\xx}^{(t)}) - f(\xx^\star)+ \tfrac{1}{n}\tsum_{i=1}^n
        \left(\tfrac{\mu}{2} \| \xx_i^{(t)} -\xx^\star \|^2- \tfrac{L}{2} \| \bar{\xx}^{(t)} - \xx_i^{(t)} \|^2\right)         \\
            & \ge f(\bar{\xx}^{(t)}) - f(\xx^\star)+ \tfrac{1}{n}\tsum_{i=1}^n
        \left(\tfrac{\mu}{4} \| \bar{\xx}^{(t)} -\xx^\star \|^2 - \tfrac{\mu+L}{2} \| \bar{\xx}^{(t)} -\xx_i^{(t)} \|^2\right) \\
            & \ge  e_t + \tfrac{\mu}{4} r_t - L \Xi_t
    \end{align*}
    where the first inequality and the second inequality uses the $L$-smoothness and $\mu$-convexity of $f_i$.

    Combine both $T_1$, $T_2$ and \Cref{prop:bounded_noise} we have
    \begin{align*}
        r_{t+1} & \le r_{t} + \gamma^2 (2L^2 \Xi_t + 4L  e_t) - 2\gamma( e_t + \tfrac{\mu}{4} r_t - L \Xi_t)
        + \tfrac{3}{n}\gamma^2(L^2 \Xi_t + 2L e_t + \bar{\sigma}^2)                                                                               \\
                & = (1-\tfrac{\gamma\mu}{2}) r_t - 2\gamma(1-5L\gamma)  e_t + \gamma L (5\gamma L + 2) \Xi_t+ \tfrac{3}{n}\gamma^2\bar{\sigma}^2.
    \end{align*}
    In addition if $\gamma\le \frac{1}{10 L}$, then
    \begin{align*}
        r_{t+1} & \le (1-\tfrac{\gamma\mu}{2}) r_t - \gamma e_t + 3\gamma L\Xi_t +\tfrac{3}{n}\gamma^2\bar{\sigma}^2.
    \end{align*}
\end{proof}

\begin{lemma}\label{lemma:convex:consensus_distance}
    Bound the consensus distance as follows
    \begin{align*}
        \Xi_t\leq 3\gamma^2{\tau_{\max}} \tsum_{t'=[t - {\tau_{\max}}]^+}^{t-1}
        \left( 2L^2 \Xi_{t'} + 4 L\ee_{t'} + (\bar{\sigma}^2+\bar{\zeta}^2) \right).
    \end{align*}
    Furthermore, multiply with a non-negative sequence $\{w_t\}_{t\ge0}$ and average over time gives
    \begin{align*}
        \tfrac{1}{W_T}\tsum_{t=0}^T w_t\Xi_t
         & \le \tfrac{1}{6LW_T} \tsum_{t=0}^T w_t  e_t
        + 6\gamma^2 {\tau_{\max}}^2 (\bar{\sigma}^2+\bar{\zeta}^2)
    \end{align*}
    where $W_T := \sum_{t=0}^T w_t$ and $\gamma\le \frac{1}{10L{\tau_{\max}}}$.
\end{lemma}
\begin{proof}
    Throughout this proof we use $\E=\E_{\xi_1^t,\ldots,\xi_n^t}$.
    Denote $[x]^+:=\max\{x, 0\}$. For all $i\in[n]$,
    \begin{align*}
        \E\norm{\ee_i^t}^2= & \E\norm{\tfrac{\gamma}{n} \tsum_{j = 1}^n\tsum_{t'=[t-{\tau_{\max}}_{ij}]^+ }^{t-1} \nabla F_j (\xx_j^{(t')}, \xi_j^{(t')})\pm \nabla f_j(\xx_j^{(t')} )  }^2   \\
        \leq                & \tfrac{\gamma^2}{n} \tsum_{j = 1}^n \E\norm{\tsum_{t'=[t-{\tau_{\max}}_{ij}]^+ }^{t-1} \nabla F_j (\xx_j^{(t')}, \xi_j^{(t')})\pm \nabla f_j(\xx_j^{(t')} ) }^2 \\
        \le                 & \tfrac{\gamma^2{\tau_{\max}}}{n} \tsum_{j = 1}^n \tsum_{t'=[t-{\tau_{\max}}]^+ }^{t-1}
        \E\norm{ \nabla F_j (\xx_j^{(t')}, \xi_j^{(t')})\pm \nabla f_j(\xx_j^{(t')} ) }^2                                                                                                     \\
        =                   & \tfrac{\gamma^2{\tau_{\max}}}{n} \tsum_{j = 1}^n \tsum_{t'=[t-{\tau_{\max}}]^+ }^{t-1}
        \E\norm{ \nabla F_j (\xx_j^{(t')}, \xi_j^{(t')})-\nabla f_j(\xx_j^{(t')} )}^2                                                                                                         \\
                            & + \underbrace{\tfrac{\gamma^2{\tau_{\max}}}{n} \tsum_{j = 1}^n \tsum_{t'=[t-{\tau_{\max}}]^+ }^{t-1}\norm{\nabla f_j(\xx_j^{(t')} )}^2}_{=:T_3}
    \end{align*}
    We can apply \Cref{prop:bounded_noise} to the first term
    \begin{align*}
        \tfrac{\gamma^2{\tau_{\max}}}{n} \tsum_{j = 1}^n \tsum_{t'=[t-{\tau_{\max}}]^+ }^{t-1}
        \E\norm{ \nabla F_j (\xx_j^{(t')}, \xi_j^{(t')})-\nabla f_j(\xx_j^{(t')} )}^2
        \le 3\gamma^2{\tau_{\max}}\tsum_{t'=[t-{\tau_{\max}}]^+ }^{t-1}
        (L^2 \Xi_{t'} + 2L\ee_{t'} + \bar{\sigma}^2).
    \end{align*}
    The second term $T_3$ can be bounded by adding $0=\pm \nabla f_j(\bar{\xx}^{(t')})\pm\nabla f_j(\xx^\star)$ inside the norm
    \begin{align*}
        T_3 & \leq \tfrac{\gamma^2 {\tau_{\max}}}{n} \tsum_{j = 1}^n \tsum_{t'=[t - {\tau_{\max}}]^+ }^{t-1}  \norm{\nabla f_j (\xx_j^{(t')}) \pm \nabla f_j(\bar{\xx}^{(t')}) \pm \nabla f_j(\xx^\star)}^2                       \\
            & \leq \tfrac{3\gamma^2 {\tau_{\max}}}{n} \tsum_{j = 1}^n \sum_{t'=[t - {\tau_{\max}}]^+}^{t-1}  \left(L^2 \norm{\xx_j^{(t')} - \bar{\xx}^{(t')}}^2 +  \norm{\nabla f_j(\bar{\xx}^{(t')}) - \nabla f_j(\xx^\star) }^2
        + \norm{\nabla f_j(\xx^\star)}^2 \right)                                                                                                                                                                                  \\
            & = 3\gamma^2 {\tau_{\max}}\tsum_{t'=[t - {\tau_{\max}}]^+}^{t-1}  \left(L^2 \Xi_{t'} +  \tfrac{1}{n} \tsum_{j = 1}^n\norm{\nabla f_j(\bar{\xx}^{(t')}) - \nabla f_j(\xx^\star) }^2
        + \bar{\zeta}^2
        \right)                                                                                                                                                                                                                   \\
            & \stackrel{\eqref{eq:smooth:optimum}}{\leq}
        3\gamma^2 {\tau_{\max}}\tsum_{t'=[t - {\tau_{\max}}]^+}^{t-1}  \left(L^2 \Xi_{t'}
        + 2 L (f(\bar{\xx}^{(t')}) - f(\xx^\star)) + \bar{\zeta}^2
        \right)
    \end{align*}
    Therefore
    \begin{align*}
        \E\norm{\ee_i^t}^2 \leq &
        3\gamma^2 {\tau_{\max}}\tsum_{t'=[t - {\tau_{\max}}]^+}^{t-1}
        (2L^2 \Xi_{t'} + 4L\ee_{t'}+(\bar{\sigma}^2+\bar{\zeta}^2)).
    \end{align*}
    Average over $i$ on both sides and note the right hand side does not depend on index $i$,
    \begin{align*}
        \Xi_t = \tfrac{1}{n}\tsum_{i=1}^n \|\ee_i^t\|^2 \leq 3\gamma^2 {\tau_{\max}} \sum_{t'=[t - {\tau_{\max}}]^+}^{t-1}
        \left( 2L^2 \Xi_{t'} + 4L\ee_{t'}+\bar{\sigma}^2 \right).
    \end{align*}
    Multiply both sides by $w_t$ and sum over $t$ gives
    \begin{align*}
        \tfrac{1}{W_T}\tsum_{t=0}^T w_t\Xi_t
         & \le \tfrac{3\gamma^2{\tau_{\max}}^2}{W_T} \tsum_{t=0}^T w_t
        \left( 2L^2 \Xi_{t} + 4 L  e_t + \bar{\sigma}^2\right)                 \\
         & =\tfrac{6\gamma^2L^2{\tau_{\max}}^2}{W_T} \tsum_{t=0}^T w_t \Xi_{t}
        + \tfrac{12\gamma^2L{\tau_{\max}}^2}{W_T} \sum_{t=0}^T w_t  e_t
        + 3\gamma^2 {\tau_{\max}} (\bar{\sigma}^2+\bar{\zeta}^2)
    \end{align*}
    where $W_T := \sum_{t=0}^T w_t$. Rearrage the terms and let $\gamma\le\frac{1}{10L{\tau_{\max}}}$ give
    \begin{align*}
        \tfrac{1}{W_T}\tsum_{t=0}^T w_t\Xi_t
         & \le\frac{1}{1-6\gamma^2L^2{\tau_{\max}}^2} \left(
        \tfrac{12\gamma^2L{\tau_{\max}}^2}{W_T} \tsum_{t=0}^T w_t  e_t
        + \tfrac{3\gamma^2{\tau_{\max}}^2}{n}( \bar{\sigma}^2+\bar{\zeta}^2)
        \right)                                              \\
         & \le \tfrac{1}{6LW_T} \tsum_{t=0}^T w_t  e_t
        + 6\gamma^2 {\tau_{\max}}^2 (\bar{\sigma}^2+\bar{\zeta}^2)
    \end{align*}
\end{proof}

\begin{theorem}
    For convex objective, we have
    \begin{align*}
        \frac{1}{T+1}\sum_{t=0}^T \left(f(\bar{\xx}^{(t)}) - f(\xx^\star)\right)
        \le 4\left(\frac{3\bar{\sigma}^2 r_0}{n(T+1)}\right)^{\frac{1}{2}}
        +4\left(\frac{6{\tau_{\max}} \sqrt{L (\bar{\sigma}^2 + \bar{\zeta}^2)} r_0}{T+1}\right)^{\frac{2}{3}}
        + \frac{10L(\tau_{\max}+1)r_0}{T+1}.
    \end{align*}
    where $r_0=\| \xx^0 - \xx^\star \|^2$.
\end{theorem}
\begin{remark}
    For target accuracy $\epsilon>0$, then $\frac{1}{T+1}\sum_{t=0}^T \left(f(\bar{\xx}^{(t)}) - f(\xx^\star)\right)<\epsilon$ after
    \begin{align*}
        \cO \left(
        \frac{\bar{\sigma}^2r_0}{n\epsilon^2}
        + \frac{\tau_{\max}\sqrt{L(\bar{\sigma}^2+\bar{\zeta}^2)}r_0}{\epsilon^{3/2}}
        + \frac{10L(\tau_{\max}+1)r_0}{\epsilon}
        \right)
    \end{align*}
    iterations. This result is similar to \cite[Theorem 2]{koloskova2021unified} except that here we replace spectral gap $p$ with the inverse of maximum delay $\frac{1}{\tau_{\max}}$.
\end{remark}
\begin{proof}
    Consider \Cref{lemma:convex:descent} and multiply both sides with $\frac{w_t}{\gamma}$ and average over time
    \begin{align*}
        \tfrac{1}{W_T} \tsum_{t=0}^T w_t  e_t
         & \le  \tfrac{1}{W_T}\tsum_{t=0}^T (
        \tfrac{w_t}{\gamma} r_t - \tfrac{w_t}{\gamma} r_{t+1})+ \tfrac{3L}{W_T} \tsum_{t=0}^T w_t \Xi_t
        + \tfrac{3\gamma}{nW_T}\tsum_{t=0}^T w_t \bar{\sigma}^2
        \\
         & \le \tfrac{1}{W_T} \tsum_{t=0}^T (
        \tfrac{w_t}{\gamma} r_t - \tfrac{w_t}{\gamma} r_{t+1})
        + \tfrac{1}{2W_T} \tsum_{t=0}^T w_t  e_t + 18 \gamma^2 {\tau_{\max}}^2 L (\bar{\sigma}^2+\bar{\zeta}^2)
        + \tfrac{3\gamma\bar{\sigma}^2}{n}
    \end{align*}
    where the second inequality comes from \Cref{lemma:convex:consensus_distance}. Then
    \begin{align*}
        \tfrac{1}{2W_T} \tsum_{t=0}^T w_t  e_t
         & \le \tfrac{1}{W_T} \tsum_{t=0}^T (
        \tfrac{w_t}{\gamma} r_t - \tfrac{w_t}{\gamma} r_{t+1}+ \tfrac{3\bar{\sigma}^2}{n}\gamma + 18 {\tau_{\max}}^2 L (\bar{\sigma}^2+\bar{\zeta}^2)\gamma^2).
    \end{align*}
    We can further consider
    \begin{align*}
        \tfrac{3L}{W_T}\tsum_{t=0}^T w_t \Xi_t = &
        \tfrac{1}{2W_T} \tsum_{t=0}^T w_t  e_t +18 {\tau_{\max}}^2 L (\bar{\sigma}^2+\bar{\zeta}^2)\gamma^2 \\
        \le                                      & \tfrac{1}{W_T} \tsum_{t=0}^T (
        \tfrac{w_t}{\gamma} r_t - \tfrac{w_t}{\gamma} r_{t+1} +
        \tfrac{3\bar{\sigma}^2}{n}\gamma + 36 {\tau_{\max}}^2 L (\bar{\sigma}^2+\bar{\zeta}^2)\gamma^2)=:\Psi_T.
    \end{align*}
    Taking $\{w_t=1\}_{t\ge0}$, then
    \begin{align*}
        \Psi_T
        \le \tfrac{r_0}{\gamma(T+1)}+ \tfrac{3\bar{\sigma}^2}{n}\gamma + 36 {\tau_{\max}}^2 L (\bar{\sigma}^2+\bar{\zeta}^2)\gamma^2.
    \end{align*}
    Apply \Cref{lemma:final_recursion} we have
    \begin{align*}
        \Psi_T
        \le
        2\left(\frac{3\bar{\sigma}^2 r_0}{n(T+1)}\right)^{\frac{1}{2}}
        +2\left(\frac{6{\tau_{\max}} \sqrt{L (\bar{\sigma}^2 + \bar{\zeta}^2)} r_0}{T+1}\right)^{\frac{2}{3}}
        + \frac{dr_0}{T+1}.
    \end{align*}
    where $d=\max\{10L, 10L{\tau_{\max}}\}\le 10L(\tau_{\max}+1)$ and at the same time
    \begin{align*}
        \frac{1}{2(T+1)} \sum_{t=0}^T  e_t
        \le & 2\left(\frac{3\bar{\sigma}^2 r_0}{n(T+1)}\right)^{\frac{1}{2}}
        +2\left(\frac{6{\tau_{\max}} \sqrt{L (\bar{\sigma}^2 + \bar{\zeta}^2)}  r_0}{T+1}\right)^{\frac{2}{3}}
        + \frac{dr_0}{T+1}                                                   \\
        \frac{3L}{T+1}\sum_{t=0}^T \Xi_t
        \le & 2\left(\frac{3\bar{\sigma}^2 r_0}{n(T+1)}\right)^{\frac{1}{2}}
        +2\left(\frac{6{\tau_{\max}} \sqrt{L (\bar{\sigma}^2 + \bar{\zeta}^2)}r_0}{T+1}\right)^{\frac{2}{3}}
        + \frac{dr_0}{T+1}
    \end{align*}
\end{proof}
\subsection{Empirical analysis of \RelaySumGrad}\label{apx:relaysum-grad:results}

In \autoref{tab:cifar-model-vs-grad}, we compare \RelaySumGrad to \RelaySumModel on deep-learning based image classification on \cifar with \vgg.
Without momentum, and with low levels of heterogeneity, \RelaySumGrad sometimes outperforms \RelaySumModel.

\begin{figure}[ht]
    \includegraphics[width=\textwidth]{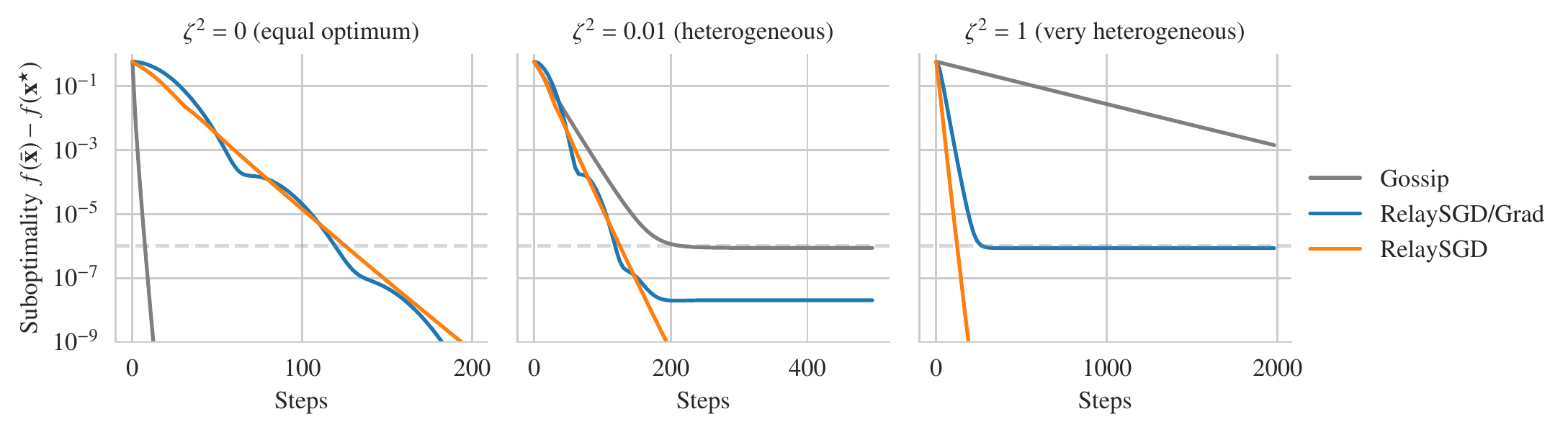}
    \vspace{-7mm}
    \caption{
        Comparing \RelaySumGrad against \RelaySumModel on random quadratics with varying levels of heterogeneity $\zeta^2$, without stochastic noise, on a ring/chain of 32 nodes.
        Learning rates are tuned to reach suboptimality $\leq 10^{-6}$ as quickly as possible.
        In contrast to \RelaySumModel, \RelaySumGrad with a fixed learning rate does not converge linearly.
        Compared to \dpsgd (Gossip), \RelaySumGrad is still less sensitive to data heterogeneity.
        \label{fig:model-vs-grad-curves}
    }
\end{figure}

\autoref{fig:model-vs-grad-curves} illustrates a key difference between \RelaySumGrad and \RelaySumModel. While \RelaySumModel behaves independently of heterogeneity, and converges linearly with a fixed step size, \RelaySumGrad reaches a plateau based on the learning rate and level of heterogeneity.

\begin{table}[ht]
    \caption{
        Comparing \RelaySumGrad with \RelaySumModel on \cifar \cite{krizhevsky2009learning} with the \vgg architecture.
        We vary the data heterogeneity~$\alpha$~\citep{lin2021quasiglobal} between 16 workers.
        For low-heterogeneity cases and without momentum, \RelaySumGrad sometimes performs better than \RelaySumModel.
        \label{tab:cifar-model-vs-grad}
    }
    \vspace{-1mm}
    \tablefontsize
\begin{tabularx}{\textwidth}{l X l l l}
    \toprule
    Algorithm & Topology & $\alpha=1.00$ & $\alpha=0.1$ & $\alpha=.01$ \\
    && (most homogeneous) & & (most heterogeneous) \\
    \cmidrule(lr){1-2} \cmidrule(lr){3-5}
All-reduce {\color{gray}(baseline)} & fully connected & 87.0\% 
\tikz{
    \draw[white,line width=5pt] (0,0) -- (1.2,0);
    \draw[gray,line width=.3pt,->] (0,0) -- (1.2,0);
    \draw[line width=.6pt] (0.748364080082286, 0pt) -- (0.7920004224235362, 0pt);
    \draw[line width=.6pt] (0.7920004224235362, -2pt) -- (0.7920004224235362, 2pt);
    \draw[line width=.6pt] (0.748364080082286, -2pt) -- (0.748364080082286, 2pt);
    \draw[line width=.6pt] (0.7505459189414959, -2pt) -- (0.7505459189414959, 2pt);
} &
87.0\% 
\tikz{
    \draw[white,line width=5pt] (0,0) -- (1.2,0);
    \draw[gray,line width=.3pt,->] (0,0) -- (1.2,0);
    \draw[line width=.6pt] (0.748364080082286, 0pt) -- (0.7920004224235362, 0pt);
    \draw[line width=.6pt] (0.7920004224235362, -2pt) -- (0.7920004224235362, 2pt);
    \draw[line width=.6pt] (0.748364080082286, -2pt) -- (0.748364080082286, 2pt);
    \draw[line width=.6pt] (0.7505459189414959, -2pt) -- (0.7505459189414959, 2pt);
} &
87.0\% 
\tikz{
    \draw[white,line width=5pt] (0,0) -- (1.2,0);
    \draw[gray,line width=.3pt,->] (0,0) -- (1.2,0);
    \draw[line width=.6pt] (0.748364080082286, 0pt) -- (0.7920004224235362, 0pt);
    \draw[line width=.6pt] (0.7920004224235362, -2pt) -- (0.7920004224235362, 2pt);
    \draw[line width=.6pt] (0.748364080082286, -2pt) -- (0.748364080082286, 2pt);
    \draw[line width=.6pt] (0.7505459189414959, -2pt) -- (0.7505459189414959, 2pt);
} \\
$\quad+$momentum &  & 90.2\% 
\tikz{
    \draw[white,line width=5pt] (0,0) -- (1.2,0);
    \draw[gray,line width=.3pt,->] (0,0) -- (1.2,0);
    \draw[line width=.6pt] (1.0707277520136405, 0pt) -- (1.1334550787102082, 0pt);
    \draw[line width=.6pt] (1.1334550787102082, -2pt) -- (1.1334550787102082, 2pt);
    \draw[line width=.6pt] (1.0707277520136405, -2pt) -- (1.0707277520136405, 2pt);
    \draw[line width=.6pt] (1.124182316389951, -2pt) -- (1.124182316389951, 2pt);
} &
90.2\% 
\tikz{
    \draw[white,line width=5pt] (0,0) -- (1.2,0);
    \draw[gray,line width=.3pt,->] (0,0) -- (1.2,0);
    \draw[line width=.6pt] (1.0707277520136405, 0pt) -- (1.1334550787102082, 0pt);
    \draw[line width=.6pt] (1.1334550787102082, -2pt) -- (1.1334550787102082, 2pt);
    \draw[line width=.6pt] (1.0707277520136405, -2pt) -- (1.0707277520136405, 2pt);
    \draw[line width=.6pt] (1.124182316389951, -2pt) -- (1.124182316389951, 2pt);
} &
90.2\% 
\tikz{
    \draw[white,line width=5pt] (0,0) -- (1.2,0);
    \draw[gray,line width=.3pt,->] (0,0) -- (1.2,0);
    \draw[line width=.6pt] (1.0707277520136405, 0pt) -- (1.1334550787102082, 0pt);
    \draw[line width=.6pt] (1.1334550787102082, -2pt) -- (1.1334550787102082, 2pt);
    \draw[line width=.6pt] (1.0707277520136405, -2pt) -- (1.0707277520136405, 2pt);
    \draw[line width=.6pt] (1.124182316389951, -2pt) -- (1.124182316389951, 2pt);
} \\[1mm]
\RelaySumModel  & chain & 87.3\% 
\tikz{
    \draw[white,line width=5pt] (0,0) -- (1.2,0);
    \draw[gray,line width=.3pt,->] (0,0) -- (1.2,0);
    \draw[line width=.6pt] (0.7696367773142714, 0pt) -- (0.8241822529922823, 0pt);
    \draw[line width=.6pt] (0.7696367773142714, -2pt) -- (0.7696367773142714, 2pt);
    \draw[line width=.6pt] (0.8241822529922823, -2pt) -- (0.8241822529922823, 2pt);
    \draw[line width=.6pt] (0.802909523248672, -2pt) -- (0.802909523248672, 2pt);
} & 
87.2\% 
\tikz{
    \draw[white,line width=5pt] (0,0) -- (1.2,0);
    \draw[gray,line width=.3pt,->] (0,0) -- (1.2,0);
    \draw[line width=.6pt] (0.7505458376624349, 0pt) -- (0.8116368136622666, 0pt);
    \draw[line width=.6pt] (0.7505458376624349, -2pt) -- (0.7505458376624349, 2pt);
    \draw[line width=.6pt] (0.7783641002394932, -2pt) -- (0.7783641002394932, 2pt);
    \draw[line width=.6pt] (0.8116368136622666, -2pt) -- (0.8116368136622666, 2pt);
} & 
86.5\% 
\tikz{
    \draw[white,line width=5pt] (0,0) -- (1.2,0);
    \draw[gray,line width=.3pt,->] (0,0) -- (1.2,0);
    \draw[line width=.6pt] (0.6867277459664789, 0pt) -- (0.7178186611695717, 0pt);
    \draw[line width=.6pt] (0.7178186611695717, -2pt) -- (0.7178186611695717, 2pt);
    \draw[line width=.6pt] (0.6867277459664789, -2pt) -- (0.6867277459664789, 2pt);
    \draw[line width=.6pt] (0.7140004716136235, -2pt) -- (0.7140004716136235, 2pt);
} \\
$\quad+$local momentum  &  & 89.5\% 
\tikz{
    \draw[white,line width=5pt] (0,0) -- (1.2,0);
    \draw[gray,line width=.3pt,->] (0,0) -- (1.2,0);
    \draw[line width=.6pt] (1.006909530271185, 0pt) -- (1.060909590937875, 0pt);
    \draw[line width=.6pt] (1.0519095686349, -2pt) -- (1.0519095686349, 2pt);
    \draw[line width=.6pt] (1.060909590937875, -2pt) -- (1.060909590937875, 2pt);
    \draw[line width=.6pt] (1.006909530271185, -2pt) -- (1.006909530271185, 2pt);
} & 
89.2\% 
\tikz{
    \draw[white,line width=5pt] (0,0) -- (1.2,0);
    \draw[gray,line width=.3pt,->] (0,0) -- (1.2,0);
    \draw[line width=.6pt] (0.9572732367298801, 0pt) -- (1.042364106936885, 0pt);
    \draw[line width=.6pt] (1.0216368490999386, -2pt) -- (1.0216368490999386, 2pt);
    \draw[line width=.6pt] (1.042364106936885, -2pt) -- (1.042364106936885, 2pt);
    \draw[line width=.6pt] (0.9572732367298801, -2pt) -- (0.9572732367298801, 2pt);
} & 
88.4\% 
\tikz{
    \draw[white,line width=5pt] (0,0) -- (1.2,0);
    \draw[gray,line width=.3pt,->] (0,0) -- (1.2,0);
    \draw[line width=.6pt] (0.9021822728893965, 0pt) -- (0.9523641277443294, 0pt);
    \draw[line width=.6pt] (0.9523641277443294, -2pt) -- (0.9523641277443294, 2pt);
    \draw[line width=.6pt] (0.9021822728893965, -2pt) -- (0.9021822728893965, 2pt);
    \draw[line width=.6pt] (0.9032732329585319, -2pt) -- (0.9032732329585319, 2pt);
} \\[1mm]
\RelaySumGrad  & chain & 88.8\% 
\tikz{
    \draw[white,line width=5pt] (0,0) -- (1.2,0);
    \draw[gray,line width=.3pt,->] (0,0) -- (1.2,0);
    \draw[line width=.6pt] (0.9420004459944636, 0pt) -- (0.9845459542491215, 0pt);
    \draw[line width=.6pt] (0.9529095590114591, -2pt) -- (0.9529095590114591, 2pt);
    \draw[line width=.6pt] (0.9420004459944636, -2pt) -- (0.9420004459944636, 2pt);
    \draw[line width=.6pt] (0.9845459542491215, -2pt) -- (0.9845459542491215, 2pt);
} & 
88.5\% 
\tikz{
    \draw[white,line width=5pt] (0,0) -- (1.2,0);
    \draw[gray,line width=.3pt,->] (0,0) -- (1.2,0);
    \draw[line width=.6pt] (0.9005459140647525, 0pt) -- (0.9480004744096229, 0pt);
    \draw[line width=.6pt] (0.9005459140647525, -2pt) -- (0.9005459140647525, 2pt);
    \draw[line width=.6pt] (0.9480004744096229, -2pt) -- (0.9480004744096229, 2pt);
    \draw[line width=.6pt] (0.9403641196814445, -2pt) -- (0.9403641196814445, 2pt);
} & 
83.5\% 
\tikz{
    \draw[white,line width=5pt] (0,0) -- (1.2,0);
    \draw[gray,line width=.3pt,->] (0,0) -- (1.2,0);
    \draw[line width=.6pt] (0.0, 0pt) -- (0.7380004796114839, 0pt);
    \draw[line width=.6pt] (0.7380004796114839, -2pt) -- (0.7380004796114839, 2pt);
    \draw[line width=.6pt] (0.6414550082250071, -2pt) -- (0.6414550082250071, 2pt);
} \\
$\quad+$local momentum  &  & 86.9\% 
\tikz{
    \draw[white,line width=5pt] (0,0) -- (1.2,0);
    \draw[gray,line width=.3pt,->] (0,0) -- (1.2,0);
    \draw[line width=.6pt] (0.7456368749791926, 0pt) -- (0.7652732215144421, 0pt);
    \draw[line width=.6pt] (0.7652732215144421, -2pt) -- (0.7652732215144421, 2pt);
    \draw[line width=.6pt] (0.7456368749791926, -2pt) -- (0.7456368749791926, 2pt);
    \draw[line width=.6pt] (0.75600042668256, -2pt) -- (0.75600042668256, 2pt);
} & 
87.8\% 
\tikz{
    \draw[white,line width=5pt] (0,0) -- (1.2,0);
    \draw[gray,line width=.3pt,->] (0,0) -- (1.2,0);
    \draw[line width=.6pt] (0.8334549909288239, 0pt) -- (0.8618186522613883, 0pt);
    \draw[line width=.6pt] (0.8334549909288239, -2pt) -- (0.8334549909288239, 2pt);
    \draw[line width=.6pt] (0.8432731926441188, -2pt) -- (0.8432731926441188, 2pt);
    \draw[line width=.6pt] (0.8618186522613883, -2pt) -- (0.8618186522613883, 2pt);
} & 
68.6\% 
\tikz{
    \draw[white,line width=5pt] (0,0) -- (1.2,0);
    \draw[gray,line width=.3pt,->] (0,0) -- (1.2,0);
    \draw[line width=.6pt] (0.0, 0pt) -- (0.33272774382071063, 0pt);
    \draw[line width=.6pt] (0.33272774382071063, -2pt) -- (0.33272774382071063, 2pt);
} \\[1mm]
    \bottomrule
\end{tabularx}
\end{table}

\end{document}